\documentclass[mnsc]{informs4_arxiv}

\usepackage{informs4_arxiv}

\usepackage{eqndefns-left} 
\RequirePackage{tgtermes}
\RequirePackage{newtxtext}
\RequirePackage{newtxmath}
\RequirePackage{bm}
\RequirePackage{endnotes}
\DoubleSpacedXI
\usepackage{algorithm}
\usepackage{algpseudocode}
\usepackage{tikz}

\usepackage{natbib}
 \bibpunct[, ]{(}{)}{,}{a}{}{,}%
 \def\bibfont{\small}%
\EquationsNumberedThrough    

\TheoremsNumberedThrough     
\ECRepeatTheorems  %

\usepackage{lscape}
\usepackage{booktabs}

\usepackage{array}
\usepackage{tabularray}
\usepackage{colortbl}

\newcolumntype{P}[1]{>{\centering\arraybackslash}m{#1}}
\usepackage{multirow,calc,amsmath}
\newlength\mylenA
\newlength\mylenB
\newlength\mylenC
\newlength\mylenD
\usepackage{tikz}
\usepackage{etoolbox}
\usepackage{adjustbox}[valign=t]
\makeatother
\usepackage{fancyvrb}
\DefineVerbatimEnvironment{shadedverbatim}{Verbatim}{
  breaklines=true,
  fontsize=\small,
  bgcolor=gray!12
}
\usepackage{color, colortbl}
\usepackage{graphicx}
\usepackage{caption}
\usepackage{subcaption}
\usepackage{dsfont}
\usepackage{mathtools}
\usepackage{physics}
\usepackage{xcolor}
\usepackage{hyperref}
\usepackage{makecell}
\usepackage[most]{tcolorbox}

\tcbset {
  base/.style={
    arc=0mm, 
    bottomtitle=0.5mm,
    boxrule=0mm,
    colbacktitle=black!10!white, 
    coltitle=black, 
    fonttitle=\bfseries, 
    left=2.5mm,
    leftrule=1mm,
    right=3.5mm,
    title={#1},
    toptitle=0.75mm, 
  }
}

\definecolor{brandblue}{rgb}{0.34, 0.7, 1}
\newtcolorbox{mainbox}[1]{
  colframe=lightgray, 
  base={#1}
}

\newtcolorbox{subbox}[1]{
  colframe=black!30!white,
  base={#1}
}

\newcommand{\bx}{\mathbf{x}}
\newcommand{\bu}{\mathbf{u}}
\newcommand{\by}{\mathbf{y}}

\newcommand{\bX}{\mathbf{X}}

\newcommand{\bI}{\mathbf{I}}

\newcommand{\bW}{\mathbf{W}}
\newcommand{\bxi}{\boldsymbol{\xi}}

\newcommand{\bS}{\boldsymbol{\Sigma}}
\newcommand{\bG}{\boldsymbol{\Gamma}}

\newcommand{\mG}{\mathcal{G}}
\newcommand{\mM}{\mathcal{M}}

\usepackage{bm}
\usepackage{soul}
\usepackage{float}

\usepackage{algorithm}
\usepackage{algorithmicx}
\usepackage{algpseudocode}

\usepackage{placeins}

\MANUSCRIPTNO{MS-0001-1922.65}



\makeatletter
\providecommand{\@openbib@code}{}%
\makeatother

\begin{document}



\RUNTITLE{The Risk of Extrapolation of Post hoc Explanations}

\TITLE{From Model Explanation to Data Misinterpretation: A Cautionary Analysis of Post Hoc Explainers in Business Research}

\ARTICLEAUTHORS{%
\AUTHOR{Tong Wang}
\AFF{School of Management, Yale University, New Haven, CT 0620, \EMAIL{tong.wang.tw687@yale.edu}} 
\AUTHOR{Ronilo Ragodos}
\AFF{Peter T. Paul College of Business and Economics, 
 University of New Hampshire, Durham, NH 03824, \EMAIL{Ronilo.Ragodos@unh.edu}}
\AUTHOR{Lu Feng}
\AFF{School of Economics and Management, 
 Tsinghua University , Beijing, China, \EMAIL{fenglu@sem.tsinghua.edu.cn}}
\AUTHOR{Yu Jeffrey Hu}
\AFF{Mitch Daniels School of Business 
 Purdue University, West Lafayette, IN 47907, \EMAIL{yuhu@purdue.edu}}
} 


\ABSTRACT{%
Post hoc explainers such as SHAP and LIME are used widely in business research to interpret complex machine learning models. Although they were designed to explain model predictions, there has been an increasing trend in which the explanations they generate are treated as evidence about underlying data relationships. Based on a systematic review of 181 studies, including 56 published in leading journals, we document that this explanation interpretation is widespread and examine its validity.
We also evaluate how well post hoc explanations reliably recover the direction and relative importance of features in true data-generating process. We introduce two metrics—direction alignment and strength alignment—and assess SHAP and LIME using simulated data with known ground truth. Although explanations often appear reasonable on average, they exhibit substantial heterogeneity in their alignment and can fail to be aligned even when their predictive accuracy is high. High predictive performance is therefore necessary but insufficient for reliable explanation.
We further show that feature correlation and the Rashomon effect (where many equally accurate models rely on different feature attributions) are key drivers of misalignment. Agreement in explanations across such models provides a practical diagnostic of reliability. Overall, our findings caution against using post hoc explainers for hypothesis validation and instead position them as exploratory tools that generate, rather than confirm, substantive insights.
}%



\maketitle

\section{Introduction}


Over recent decades, the growth in data volume and complexity has made machine learning (ML) a mainstay of business analytics, particularly for prediction tasks \citep{kleinberg2015prediction,mullainathan17applied}. In these settings, researchers typically solve a supervised learning problem - predicting a target ($Y$) using a set of explanatory variables or features ($X$). However, in business applications, prediction alone is often insufficient. Researchers and practitioners are also interested in gaining insights about their data, in particular, \textit{how explanatory variables ($X$) are associated with the target variable ($Y$)}. Unfortunately, most ML models with high accuracy operate as black boxes\footnote{It is worth acknowledging that recent years have seen the emergence of inherently interpretable models that can attain comparable performance to black box models, sans the reliance on post hoc explanations. However, these models, which lie beyond the purview of this paper's discussion, will not be further delved into here.} that produce predictions without revealing how features relate to the targets.

\begin{figure}[t]
        \centering
        \includegraphics[trim={4.75cm 7.5cm 1.125cm 4.25cm},clip, width = 0.9\textwidth]{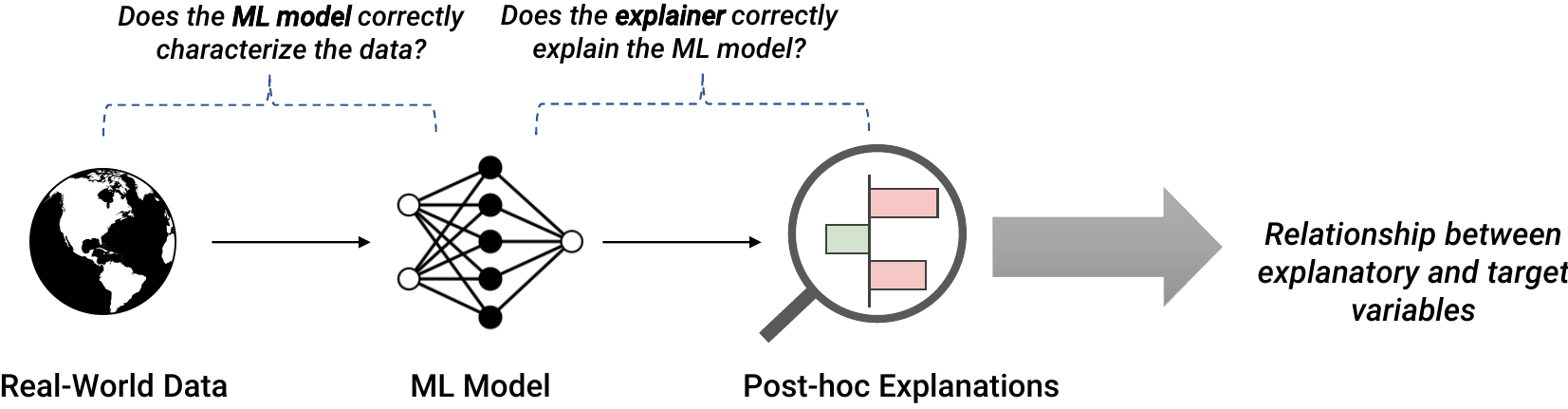}
        \caption{Explanation Pipeline: the pipeline of using post hoc explainers to explain a machine learning model. Additionally inferring information about the actual marginal effects of the features constitutes mistaken usage of the pipeline.}
        \label{fig:pipeline}
    \end{figure}

In response to this opacity, post hoc explainers have naturally garnered significant attention from business researchers. These tools provide various ways to interpret black box models, such as feature importance \citep{ribeiro16lime,lundberg17shap}, partial dependence plots \citep{friedman01pdp}, counterfactual examples \citep{mothilal20dice}, etc. Many recent studies in business research follow a two-stage pipeline that we will refer to as the \emph{Explanation Pipeline}. The Explanation Pipeline, illustrated in Figure \ref{fig:pipeline}, consists of a first stage where a black box ML model is employed for a prediction task, and a second stage where the ML model is explained by a post hoc explainer. This pipeline can illuminate model behavior by, for example, revealing which features appear most influential for the model’s predictions.

Among post hoc explainers, two are most widely used in empirical research: Shapley Additive Explanations (SHAP) \citep{lundberg17shap} and Local Interpretable Model-Agnostic Explanations (LIME) \citep{ribeiro16lime}. SHAP is theoretically grounded in cooperative game theory and assigns each feature a Shapley value which represents its contribution to the model’s prediction relative to a baseline. Intuitively, SHAP asks how much a feature contributes to the prediction when averaged over all possible subsets of other features, thereby providing a principled attribution of importance. LIME approximates the behavior of a complex model locally around a specific instance with a sparse linear model. It perturbs the input features in the neighborhood of the instance and fits a simple, interpretable surrogate model—typically a linear regression—using proximity-weighted samples. The coefficients of this local surrogate model are then interpreted as measures of feature influence for that particular prediction. Both SHAP and LIME provide information about how features influence a model’s predictions, in terms of both the direction of influence (whether a feature increases or decreases the predicted outcome) and the strength of influence (the magnitude of that effect). In applied research, these explanations are often examined across many observations and aggregated to reflect broader patterns in model behavior.

Although post hoc explainers are designed to explain predictive models, a growing body of research interprets explanations they produce as evidence about underlying data-generating processes. In particular, these studies generalize explanations that characterize relationships between features and model predictions ($X \mapsto \hat{Y}$)—which are inherently model-dependent—to infer relationships between features and the true outcome ($X \mapsto Y$) in the data. This perspective is increasingly visible across recent business research. For example, \citet{park2024extracting} interpret SHAP explanations as indicators of which smartphone features customers prefer; \citet{jiang24churn} suggest that customers with high values of features associated with large SHAP values are more likely to churn; and \citet{sanchez2023role} argue that SHAP explanations reveal drivers of hotel guest satisfaction and use these explanations to motivate managerial actions. Some recent work has even advocated combining machine learning models with SHAP as an ``alternative to OLS" \citep{berger2023supply}.
To assess how prevalent this practice is, we conducted a manual review of 181 papers that made substantive SHAP or LIME. We find that approximately 42.5\% of these papers interpret post hoc explanations as evidence of data-level relationships, indicating that this practice is widespread in business research.

These interpretations raise important conceptual questions about the scope of validity of post hoc explainers. Specifically, \textbf{to what extent can explanations derived from machine learning models ($X \mapsto \hat{Y}$) reflect ($X \mapsto Y$) in the underlying data-generating process?} To address this question, we conduct extensive experiments to assess whether post hoc explanations reliably recover the direction and relative strength of true $\mathbf{X}\mapsto Y$ relationships.

Our study proceeds in four steps. First, we identify problematic yet prevalent ways post hoc explainers are interpreted in prior research. Across a wide range of applications, researchers implicitly rely on two broad types of explanatory claims, which we term \textbf{direction} and \textbf{strength} interpretations. To formalize these interpretations, we give mathematical definitions of direction alignment and strength alignment that are applicable to any additive post hoc explainer. Direction alignment evaluates whether the direction of outcome change implied by an explainer under feature perturbations matches the true direction of change implied by the data-generating process. Strength alignment evaluates whether the relative importance ranking induced by an explainer matches the true importance ranking in the data-generating process. 

Second, we conduct evaluations using simulated datasets in which the ground-truth $\mathbf{X} \mapsto Y$ relationships are known and precisely controlled. We systematically vary multiple data-generation factors to construct a rich set of ground truths spanning a wide range of complexities.
We find that, although explanations often appear accurate on average, their alignment with $\mathbf{X} \mapsto Y$ exhibits substantial heterogeneity with pronounced long-tailed deviations, even when predictive performance is high. This dispersion in outcomes undermines the practical utility of post hoc explanations, as researchers typically lack diagnostics to assess \textit{when or to what extent} an explanation faithfully reflects the true $X \mapsto Y$ relationship.

Third, we investigate the drivers of explanation misalignment—i.e., when explanations of $X\mapsto\hat{Y}$ fail to recover the direction and strength of the underlying $\mathbf{X}\mapsto Y$ relationships. We focuses on:  (i) data-level factors such as correlation structure among features, the interaction of features, etc, (ii) the predictive performance of the trained model, and (iii) the Rashomon Effect \citep{semenova22rashomon, paes23inevitable}. Crucially, we find that high predictive accuracy is \textbf{necessary but insufficient} for post hoc explanations to align with $\mathbf{X}\mapsto Y$. Models with nearly identical predictive performance can yield markedly different explanations, reflecting the presence of large Rashomon sets. Among data characteristics, feature correlation emerges as the dominant driver of misalignment, whereas other forms of data complexity have limited practical impact.

Finally, we identify diagnostic signals that help assess this misalignment. 
We operationalize it through two measurements that we define called the \emph{prediction agreement} and \emph{explanation agreement} within Rashomon sets, and examine how these measures relate to direction and strength alignment of post hoc explanations. We find that explanation-based agreement is strongly correlated with alignment (especially for recovering strength) whereas prediction-based agreement is  less informative. This result clarifies that it is not disagreement in predictions, but disagreement in feature attributions among observationally equivalent models, that signals heightened risk of misinterpreting $\mathbf{X}\mapsto Y$ from post hoc explanations. In this way, Rashomon agreement provides a practical, dataset-level diagnostic for assessing when explanations of $X\mapsto\hat{Y}$ are unlikely to recover the underlying $X \mapsto Y$ structure.

Overall, our paper identifies and cautions against a growing trend of the inappropriate use of post hoc explanations to draw inferences about data, leading to data misinterpretation. We argue that post hoc explanations should not be used to \textbf{validate} hypotheses about data; rather, their proper role in the research process is to \textbf{propose} hypotheses that can subsequently be tested through more rigorous methods. Making the distinction between exploration and validation is critical to avoid overstating the reliability or generalizability of insights derived from post hoc explainers and to ensure that business research continues to rest on sound empirical and theoretical foundations.

  The paper is organized as follows. Section~\ref{sec:related_work} reviews related work on SHAP, LIME, and their known limitations. Section~\ref{sec:use_in_b_research} surveys common problematic uses of post hoc explanations and illustrates these misinterpretations using a demonstration dataset.
Section~\ref{sec:main_alignment} presents empirical analyses showing how much post hoc explanations align with the true $\mathbf{X}\mapsto Y$, and Section \ref{sec:factors} investigates the factors that drive the misalignment. Section~\ref{sec:robustness} reports a series of robustness checks.
Finally, we examine whether agreement among approximately equally accurate models can be used to diagnose explanation reliability in Section~\ref{sec:rashomon_agreement}. The paper concludes in Section~\ref{sec:discussion} with a summary of our findings and a discussion of the implications for the appropriate use of post hoc explanations in business research.

\section{Related Work}
    \label{sec:related_work}
    This section introduces SHAP and LIME, reviews critiques of SHAP and LIME in the computer science literature, and then provides a discussion of the use and reception of SHAP and LIME in the business literature.
    To serve as a point of contrast with SHAP and LIME, we also review a standard set of econometrics tools used in business literature to extract data-level insights. Finally, we explain how the \emph{Rashomon Effect} puts the worth of using post hoc explainers to make inferences about $X \mapsto Y$ relationships into great doubt.

    \subsection{Post Hoc Explainers - SHAP and LIME}
SHAP \citep{lundberg17shap} and LIME \citep{ribeiro16lime} are the most widely used post hoc explainers. We use them as representative methods because they dominate applied and empirical research across domains.

SHAP (SHapley Additive exPlanations) is a unified framework for interpreting machine-learning predictions through feature attribution.\footnote{It is ``unified" in the sense that its framework generalizes LIME's.} It builds on the concepts of game theory, particularly the Shapley value, to assign an importance value to each feature for a particular prediction. The values represent the contribution of each feature to the difference between the actual prediction and the average prediction over the dataset. A large part of SHAP's appeal is due to theoretical properties of its explanation model, the \emph{consistency} and \emph{local accuracy} properties in particular. Consistency guarantees that if a model changes in a way that makes a feature more important, the SHAP value of that feature will not decrease. Local accuracy ensures that the contributions of all features to a prediction sum up to the actual predicted value minus the average prediction, providing a detailed and accurate decomposition of the prediction. 
    

Although not used as widely as SHAP in business research, LIME (Local Interpretable Model-agnostic Explanations) is the earliest model-agnostic post hoc explainer, introduced by \citet{ribeiro16lime}. LIME provides local, model-agnostic explanations by approximating a complex predictive model with a simple, interpretable surrogate model—typically a sparse linear regression—fitted in a neighborhood around a focal observation. The coefficients of this locally fitted surrogate capture how variations in each feature within this neighborhood are associated with changes in the model’s prediction. As such, LIME coefficients are commonly interpreted as \textit{approximate local marginal effects} of the underlying model, defined relative to the locality induced by the perturbation and weighting scheme. 

\subsection{Critiques in Computer Science Literature}
\label{sec:cs_critique}

Concerns about post hoc explanation methods such as SHAP and LIME originated in the machine learning literature, where researchers have examined their limitations along dimensions including prediction accuracy, stability, robustness, and human interpretability. While our work shares the broad objective of rigorously evaluating post hoc explainers, it differs fundamentally in \emph{what} is being evaluated and \emph{for what purpose}.

A central distinction is that most critiques in the computer science literature assess post hoc explainers with respect to the \emph{predictive model} being explained. From this perspective, the model is treated as the primary object of explanation, and the goal is to assess how well an explainer reflects properties of the mapping $X \mapsto \hat{Y}$ learned by the model. In contrast, the perspective adopted in this paper is explicitly \emph{data-level}. We ask whether explainers can recover a minimum amount of information about the \emph{data-generating process} (DGP) $X \mapsto Y$ that is necessary to support common forms of inference made in empirical research—specifically, inference about the \emph{direction} and \emph{relative strength} of feature effects. Accordingly, our focus is not on whether explainers provide comprehensive or human-ideal representations of a model, but on whether they satisfy basic validity conditions implicitly assumed when researchers generalize SHAP or LIME explanations to data-level.

Within the model-level literature, different dimensions of post hoc explanations are evaluated. One stream assesses explanations' ability to reproduce a model’s predictions or decision boundary, effectively treating the explainer as a surrogate function whose outputs should closely match those of the model itself \citep{han22which}. Other work probes explanation correctness indirectly through stress tests, such as sensitivity to irrelevant features or robustness to perturbations \citep{blunsom2019can,nguyen2020quantitative,deyoung20eraser,luss2021leveraging,carmichael21framework,li23xaibenchmark}. A related but more extreme line of research considers adversarial settings in which explanations are deliberately manipulated without altering model predictions \citep{slack19fooling}. 

A second, complementary strand of work emphasizes \emph{human-centered interpretability}. Common desiderata include \emph{stability}, \emph{completeness}, \emph{simplicity}, and related concepts\citep{doshi17rigorous,markus2021role,agarwal22openxai}. Stability, a requirement that similar inputs--output pairs to receive similar explanations, is typically evaluated locally and pairwise, making it a stronger condition than global notions such as directional consistency \citep{alvarez18robustness,yeh2019fidelity,visani21stability,kelodjou2024shaping}. Completeness requires explanations to contain sufficient information to reconstruct the model’s input--output mapping \citep{nguyen2020quantitative,bassan2023towards}, while broadness concerns the scope over which an explanation applies \citep{nguyen2020quantitative,zhou21evaluating}. Simplicity-based criteria deliberately trade off completeness and broadness to enhance human comprehensibility \citep{nguyen2020quantitative,markus2021role,makridis2023towards}. The most demanding proposals further require explanations to align with psychological models of human understanding or to be interactive and personalized \citep{jacovi2023diagnosing,kaur2022sensible,ehsan2023charting,sokol2020one,dazeley21conversational}.

Our contribution is orthogonal to both strands. We ask whether attributive explainers satisfy two \emph{minimal validity conditions} required for data-level inference: (i) whether the sign of feature attributions reliably reflects the direction of a feature’s impact on $Y$, and (ii) whether the relative magnitudes of attributions meaningfully reflect the relative importance of features in the data. These requirements are substantially weaker than those commonly studied in the computer science literature. Nonetheless, we show that they are frequently violated even under favorable conditions, e.g. when models are well-specified and explainers are used exactly as intended.

\subsection{SHAP and LIME Use in the Business Literature}
SHAP and LIME have become increasingly adopted in the business literature as practical tools for interpreting complex predictive models. 
Reflecting the applied orientation of the field, existing critiques in the business literature tend to focus on the downstream consequences of using post hoc explanations rather than on their theoretical properties. Specifically, one branch of the literature studies the (negative) ways in which the use of explainable AI can impact use behavior in general \citep{bauer23infoproc,von2025knowing} and in specific contexts like  healthcare \citep{hassan25unlocking} and the identification of false information \citep{lee24false}. Legal regulations requiring model explanations has led to research on the resulting challenges for regulatory compliance \citep{mohammadi2025regulating} in domains such as credit scoring \citep{goodman17gdpr,bucker2022transparency,chen2024interpretable,debock24xaisurvey,ballegeer25stability}. These studies emphasize the practical and institutional implications of post hoc explainers in real-world decision-making contexts.

A second line of work uses identified limitations of SHAP and LIME as motivation for developing alternative explanation approaches. For instance, \citet{fernandez20explaining} contrast feature-importance methods with counterfactual explanations and characterize settings in which counterfactuals more directly capture how features influence decisions. \citet{kim2023rolex} proposes a new explanation method that  offers improvements/generalizations to each of the three stages of LIME explanation (sampling, local model fitting, explanation generation).


However, these papers still focus on scenarios involving the intended use of post hoc explainers: explaining the behavior of the model to which they are applied. In contrast, our work studies scenarios in which model-level explanations are used to used as a basis for data-level conclusions. Specifically, we ask whether model explanations \textit{can be used to reliably infer fundamental properties of the underlying data-generating mechanism $\mathbf{X}\mapsto Y$}.

\subsection{Classical Approaches to Investigating $\mathbf{X}\mapsto Y$ in Econometrics}

Econometric research is primarily concerned with inferring relationships between covariates and outcomes at the level of the data-generating process $\mathbf{X}\mapsto Y$, typically under explicit causal and identification assumptions \citep{haavelmo1944probability,wold1954causality,pearl2009causality,heckman15casual}. Rather than explaining the behavior of a fitted predictive model, these methods aim to estimate how changes in $\mathbf{X}$ affect $Y$ directly from observed data.

Classical econometric approaches rely on regression-based, quasi-experimental, and structural modeling techniques—such as instrumental variables, control strategies, and panel designs—to support interpretable claims about $\mathbf{X}\mapsto Y$ in observational settings \citep{angrist01iv,winship1999estimation,chamberlain1984panel}. Crucially, these inferences are drawn without passing through an intermediate machine-learning model. In contrast to post hoc explanation pipelines—where conclusions about $\mathbf{X}\mapsto Y$ are mediated by both the learned model and the explainer—econometric analyses seek to avoid this additional layer of abstraction, thereby reducing the risk that a model's inductive biases distort inference about the underlying data-generating process.

\subsection{Rashomon Sets and the Rashomon Effect}
A main contributing factor to the misalignment between post hoc explanations and $\mathbf{X}\mapsto Y$ is the Rashomon Effect.
Originally articulated in the context of statistical learning by Breiman, the \emph{Rashomon effect} refers to the existence of a large set of models that achieve comparably strong predictive performance, yet rely on substantially different internal representations or feature usage to generate predictions \citep{breiman2001statistical}. Recent work formalizes this idea through the notion of a \emph{Rashomon set}: a collection of all models whose predictive performance lies within a small tolerance of the optimal model \citep{semenova22rashomon,rudin2024amazingthingscomehaving}. Research stemming from the idea of a Rashomon effect emphasizes that predictive accuracy alone is insufficient to pin down a unique explanation of the data-generating process, as many structurally distinct models may be observationally equivalent.

It is important to recognize that the presence of a Rashomon set in a prediction problem is not a pathological edge case. Intuitively, when a data-generating process is noisy, many distinct functional relationships can fit the observed outcomes almost equally well, especially when only finite samples are available. Noise flattens the empirical risk landscape, reducing the extent to which competing models that rely on different features or structures can be distinguished. Formalizing this intuition, \citet{paes23inevitable} show that the Rashomon effect is inevitable when learning from finite data, implying that ambiguity across well-performing models (in terms of how explanatory variables relate to the target) is a fundamental property of empirical modeling rather than a failure of estimation. Moreover, as highlighted by \citet{rudin2024amazingthingscomehaving}, the resulting Rashomon sets can be large and diverse even in seemingly well-behaved prediction problems, with meaningful consequences for interpretability and scientific inference.

Crucially for our setting, the same data that permit multiple near-optimal models also permit multiple, potentially conflicting explanations. Because each model in the Rashomon set encodes a different plausible mapping from $X$ to $Y$, a post hoc explanation of any single model reflects only one of many admissible interpretations. As a result, even when a model achieves high predictive accuracy, its explanations need not align with the underlying data-generating relationship. This multiplicity of admissible explanations is a central challenge we address in this paper.

Consistent with this perspective, feature importance, or more generally, model explanations, can vary widely across models within a Rashomon set. Even when models exhibit near-identical predictive accuracy, they may disagree on individual predictions and on how input features contribute to those predictions, making post hoc explanations particularly fragile on unseen or unlabeled data \citep{hsu2022rashomon}.

    \section{Current Uses and Interpretations of SHAP and LIME in Business Research}
        \label{sec:use_in_b_research}
        To assess how post hoc explanation methods are currently applied and interpreted in business research. We first describe the typical \emph{Explanation Pipeline} employed in prior studies, which involves training a predictive model and subsequently applying an explainer to interpret it. Then, we analyze the types of claims researchers draw from explanations.
 Finally, we conduct a comprehensive review of the prior studies that made substantive use of SHAP or LIME to show the prevalence of the identified interpretations. 

\subsection{The Explanation Pipeline in Business Research}
    \label{sec:pipeline_demo}

    
An Explanation Pipeline consists of two stages. In the first stage—the prediction stage—a machine learning model $\mathcal{M}$ is trained on data $\mathcal{D}$ to predict an outcome of interest, and in the second stage—the explanation stage—an explainer model $\mathcal{E}$ (e.g., SHAP or LIME) is applied to $\mathcal{M}$ to generate post hoc explanations.
    
To demonstrate how the Explanation Pipeline is typically applied, we simulate an app-downloading dataset \footnote{Real datasets are less suitable for the present study because evaluating the faithfulness of post hoc explanations requires access to the true underlying relationships.}. The dataset includes two demographic features (``Income" and ``Age") and eight features describing users’ app-store behavior such as ``Time Spent on Phone," ``Data Plan Unlimited," ``Peer Influence," etc.
 We generate the labels using a linear ground-truth model that simulates individuals’ decisions about whether to download a certain mobile app. 
We then create the dataset $\mathcal{D}$ with 5000 data points by sampling from the distributions of features to simulate user profiles ($\mathbf{X}$) and using $\mathcal{G}$ to create the labels ($Y$). See Appendix \ref{apx_sec:app_details} for more details of the data generation. 

Following the Explanation Pipeline, we first train an XGBoost classifier $\mathcal{M}$ using 5-fold cross-validation and a grid search for hyperparameter tuning. The resulting model achieves an accuracy of approximately 72\% on the held-out test set. We then apply SHAP and LIME to $\mathcal{M}$ to generate explanations.

\subsection{Types of Interpretations Drawn from SHAP and LIME}
\label{sec:interpretations}
In this subsection, we illustrate how SHAP and LIME explanations are interpreted.
\subsubsection{SHAP Interpretations}\label{sec:shap_interpretation}
\begin{figure}
\centering
\begin{subfigure}{.6\textwidth}
  \centering
  \includegraphics[width=.9\linewidth]{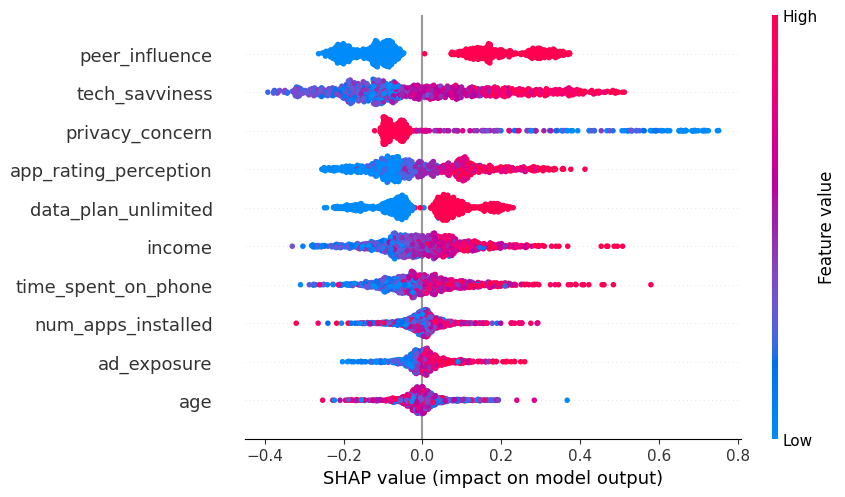}
  \caption{SHAP Beeswarm}
  \label{fig:shap_beeswarm}
\end{subfigure}%
\begin{subfigure}{.45\textwidth}
  \centering
  \includegraphics[width=.85\linewidth]{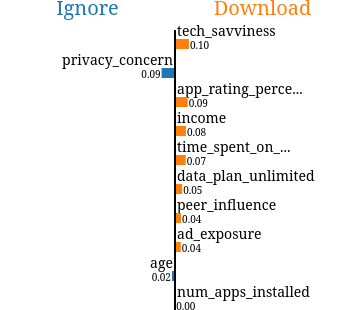}
  \caption{LIME Local}
  \label{fig:sub2}
\end{subfigure}
\caption{SHAP and LIME Demo}
\label{fig:sec_3_shap_lime_demo}
\end{figure}

SHAP provides feature attributions that represent the marginal contributions of each feature, i.e., how much a feature adds to the model’s prediction when considered across all possible combinations of features. When aggregated across observations, SHAP attributions are often visualized using a beeswarm plot (Figure \ref{fig:shap_beeswarm}), which displays the distribution of SHAP  values for each feature. Each dot represents a single observation’s SHAP value for a given feature, with color indicating the corresponding input value (e.g., red for high, blue for low). The horizontal position of a dot shows both the direction and magnitude of the feature’s contribution; points to the right indicate positive contributions to the predicted outcome (higher $\hat{Y}$), whereas points to the left indicate negative contributions (lower $\hat{Y}$).

From this visualization, researchers typically draw two types of interpretations: \emph{direction interpretation} and \emph{strength interpretation}.

\paragraph{Direction Interpretation — Larger Shapley values correspond to larger model predictions ($\hat{Y}$)}
Under this interpretation, direction refers to how feature values are ordered with respect to the model’s predictions: feature values that correspond to larger Shapley values are interpreted as leading the model to produce larger predicted outcomes $\hat{Y}$, and vice versa. For example, if higher values of ``Peer Influence" are associated with larger Shapley values, researchers interpret this as indicating that greater peer influence leads to larger predicted likelihoods of downloading the app. Conversely, if higher ``Privacy Concern" corresponds to smaller Shapley values, this is interpreted as indicating lower model predictions.

\paragraph{Strength Interpretation — Mean absolute Shapley values indicate a feature’s importance to the model}
Under this interpretation, a feature’s strength is quantified by its aggregated contribution, measured as the mean absolute Shapley value across observations. This aggregation reflects how strongly the model relies on that feature, on average, when generating predictions. Features with larger mean absolute Shapley values are therefore interpreted as being more influential in determining $\hat{Y}$. For example, ``Peer Influence" has the largest mean absolute Shapley value, indicating that it contributes the most, on average, to the model’s predictions across the dataset. 

\subsubsection{LIME Interpretations}
LIME provides local, model-agnostic explanations by approximating a complex predictive model with a sparse linear regression fitted in a neighborhood around a focal observation. 

Similarly, researchers typically draw two types of interpretations analogous to those used for SHAP:

\paragraph{Direction Interpretation — The sign of a coefficient indicates whether increasing the feature raises or lowers the model’s prediction locally.}
A positive coefficient implies that, in the local region around the focal observation, an increase in that feature moves the surrogate model’s prediction upward (i.e., increases the predicted likelihood of the outcome). For example, ``Peer Influence" has a positive LIME coefficient for an individual user, then higher peer influence increases the locally predicted likelihood of app downloads for that specific user. ``Privacy Concerns" has a negative coefficient, then within that local neighborhood, greater privacy concerns reduce the predicted likelihood of downloading the app.

\paragraph{Strength Interpretation — The magnitude of a coefficient reflects the strength of the feature’s local influence on the model.}
The absolute magnitude of a LIME coefficient reflects how influential the feature is in shaping the surrogate model’s prediction for that specific observation. Larger absolute coefficients indicate stronger local effects. Thus, for a given observation, the most influential feature is the one with the largest absolute LIME coefficient, followed by features with progressively smaller magnitudes. 

As we will show in the next section, direction and strength interpretations summarize the primary ways in which SHAP and LIME outputs are used to infer model behavior in applied business research.

\subsection{The Controversial Generalization to Interpreting $X \mapsto Y$ in the Literature}
The interpretations described in the previous subsection are formally statements about how a trained model maps inputs to predictions ($X \mapsto \hat{Y}$). The controversial practice we investigate arises when these model-level interpretations are generalized into claims about the underlying relationship between features and outcomes in the data-generating process ($X \mapsto Y$). Under this generalization, quantities defined to describe model behavior are implicitly treated as evidence about how features influence outcomes in the data.

In the applied literature, this shift in interpretation is typically not made explicit. Instead, language that is formally appropriate for describing model behavior is framed as describing how features are associated with the outcome in the data. As a result, direction interpretations are often read as indicating whether a feature corresponds to larger or smaller values of the outcome in the data, while strength interpretations are read as indicating the relative prominence or importance of features with respect to the observed outcome.

Boxes below provide prototypical examples of this linguistic shift. In each case, explanations that describe how a model’s prediction varies with features are articulated as statements about corresponding variation in the outcome $Y$, rather than the predicted outcome $\hat{Y} $.
\begin{mainbox}{Examples of controversial claims about the direction of influence} 
\begin{itemize}
    \item ``\textit{Positive SHAP values indicate that the variable has a facilitative effect on $\langle Y\rangle$, while negative values indicate a suppressive effect.}''
    \item ``\textit{Individuals whose predictions received larger SHAP values for this feature are more likely to have $\langle Y=1\rangle$.}''
    \item ``\textit{Choosing features with positive mean SHAP values is recommended because they have positive effects on $\langle Y\rangle$.}'' 
\end{itemize}
\end{mainbox}

\begin{mainbox}{Examples of controversial claims about the strength of influence}
\begin{itemize}
    \item ``\textit{This study provides insights for planning the design of $\langle X\rangle$ by assessing the relative importance and rankings of the $\langle X_j\rangle$ in predicting $\langle Y\rangle$.}''
    \item ``\textit{Shapley values enable us to identify and rank relevant $\langle X_j\rangle$ for the investigated $\langle Y\rangle$.}''
    \item ``\textit{Feature rankings and visualizations of impact patterns provide policymakers with evidence for strategies to improve $\langle Y\rangle$.}"
\end{itemize}
\end{mainbox}

\paragraph{Prevalence of Data-Level Inference}

To assess the prevalence of the interpretive practices described above, we conducted a systematic review of the literature.

We compiled a sample of 181 papers published in journals included in the \textit{UTD 24}, the \textit{FT50}, and \textit{INFORMS} lists, as well as papers indexed in Web of Science and SSRN (to account for the long publication timelines of academic journals). All selected papers made substantive use of SHAP or LIME. Our objective was to quantify (i) the prevalence of directional and strength-based interpretations of feature attributions and (ii) the prevalence of data-level inference based on post hoc explanations.

To mitigate potential human bias in coding, we complemented manual annotation with an LLM-assisted review of each paper. Final classifications were determined through an adjudication process that reconciled disagreements between human coders and LLM outputs. Detailed documentation of the review protocol, coding scheme, and adjudication procedure is provided in Online Appendix \S 1. 

Across the full sample,   94\% used SHAP and 13\% used LIME, with 7\% using both methods. 75.4\% of papers made directional interpretations of feature attributions, and 97.8\% made strength-based interpretations. These two dimensions constitute the core components of our coding framework and serve as the conceptual basis for identifying subsequent data-level inference.

Overall, 42.5\% of the sampled papers engaged in what we classify as problematic interpretations of post hoc explanations. The prevalence of such interpretations is substantially lower in leading journals: 14.3\% among \textit{UTD 24} journals, 14.0\% among \textit{FT50} journals, and 17.4\% among \textit{INFORMS} journals. Across the union of these three journal lists, the overall rate is 16.1\%.

\section{Do Post hoc Explanations Align with Data?}
\label{sec:main_alignment}

In this section, we assess whether post hoc explanations on $X\mapsto\hat{Y}$ align with true $X\mapsto Y$ relationship in the data. We first define metrics that capture directional alignment and strength alignment between explanations and ground truth. We then discuss how we simulate datasets with different properties such that the $X\mapsto Y$ relationships can be precisely controlled and systematically varied, enabling a reliable assessment of alignment and the relationships between explanation alignment and data properties. Finally, we evaluate SHAP and LIME explanations using these metrics.

\subsection{Evaluation Metrics for Direction and Strength Alignment}\label{sec:metrics}

We propose a  framework for measuring \textit{direction alignment} and \textit{strength alignment} that unifies SHAP and LIME explanations. The framework is defined at a level of generality that applies to any post hoc explanation method that assigns feature-level attribution values, any predictive model $\mathcal{M}$, and any ground-truth data-generating process $\mathcal{G}$. Method-specific evaluations, such as those for SHAP and LIME used in our experiments, arise as instances of our definition schema.

\subsubsection{Direction Alignment}
Direction alignment evaluates, \textbf{for each feature}, whether the \textit{direction of outcome change implied by an explainer} matches the \textit{true direction of outcome change in the data}. Intuitively, if an explainer implies that increasing feature $j$ increases (or decreases) the model prediction $\hat{Y}$, direction alignment checks whether the same perturbation produces a corresponding directional change in the true outcome $Y$.

The evaluation is performed at the instance level and then aggregated across instances and perturbations to obtain an overall measure of directional consistency.

Let $e_{\mathcal{E}}(\mathbf{x}^{(i)}, j)$ denote the explanation assigned by method $\mathcal{E}$ to feature $j$ for instance $i$. To compare a DGP with what an explainer can be understood as implying about that DGP, we consider perturbations of a single feature while holding all other features fixed.

For an instance $\mathbf{x}^{(i)}$, feature $j$, and perturbation magnitude $\delta$, we define the perturbed instance
\begin{equation}
    \mathbf{x}^{(i)}_{j,\delta} 
    = 
    (x^{(i)}_1,\ldots,x^{(i)}_j+\delta,\ldots,x^{(i)}_d).
\end{equation}
and consider a finite set of allowable perturbations to choose $\delta$ from:
\[
    \Delta = \{m \cdot d \mid m \in \{-M,\ldots,-1,1,\ldots,M\}\},
\]
where $d$ is a step size controlling perturbation magnitude and $M$ determines the extent of the evaluation range. 
The choice of $\Delta$ is independent of the explanation method and defines the evaluation protocol used to compare explainer-implied and true directional responses.

To formalize direction alignment, we decompose the evaluation into three components: 
(i) the direction of change implied by the explainer when a feature is perturbed, 
(ii) the true direction of change induced by the same perturbation under the data-generating process, 
and (iii) an agreement indicator that compares the two and aggregates the comparison across instances and perturbations.

\noindent\textbf{Explainer-indicated direction of change} The explainer-indicated direction of change for feature $j$ is defined as 
\begin{equation}
    \Delta e^{(i)}_{j}(\delta)
    =
    e_{\mathcal{E}}(\mathbf{x}^{(i)}_{j,\delta}, j)
    -
    e_{\mathcal{E}}(\mathbf{x}^{(i)}, j).
\end{equation}
This quantity captures whether the explainer believes the contribution of feature $j$ increases or decreases when $x_j$ is perturbed. This definition applies to any explanation method that produces feature-level attributions and is evaluated over the shared perturbation set $\Delta$.

\noindent\textbf{True direction of change}
The true direction of change induced by perturbing feature $j$ is defined as
\begin{equation}
    \Delta Y^{(i)}_{j}(\delta)
    =
    \mathcal{G}(\mathbf{x}^{(i)}_{j,\delta})
    -
    \mathcal{G}(\mathbf{x}^{(i)}),
\end{equation}
which represents the actual effect of modifying feature $j$ on the outcome generated by the ground-truth process $\mathcal{G}$. Evaluating this quantity requires access to $\mathcal{G}$, allowing us to compute the true outcome at the perturbed input.

\noindent\textbf{Agreement indicator and aggregation}
An explainer is directionally aligned for instance $i$, feature $j$, and perturbation $\delta$ if the explainer-indicated and true changes point in the same direction, i.e.,
\begin{equation}\label{eqn:sign}
    \operatorname{sign}\!\left(\Delta e^{(i)}_{j}(\delta)\right)
    =
    \operatorname{sign}\!\left(\Delta Y^{(i)}_{j}(\delta)\right).
\end{equation}

Requiring strict sign agreement for all perturbations, however, is undesirable when the true effect is extremely small, as the ground truth then provides no reliable directional signal. To avoid penalizing explainers for such cases, we relax the criterion by treating small-magnitude true changes as directionally uninformative.

We define the agreement indicator
\begin{equation}\label{eqn:dir_indicator}
\small
    I^{\text{dir}}_{ij}(\delta)
    \triangleq
    \begin{cases*}
            \mathbb{I}\!\left(
        \Delta e^{(i)}_{j}(\delta)\,\Delta Y^{(i)}_{j}(\delta) > 0
    \right) &if $|\Delta Y_j^{(i)}(\delta)| > \epsilon_1$ \\
    \mathbb{I}\!\left(
       |\Delta e^{(i)}_{j}(\delta)| < \epsilon_2
    \right) & otherwise.
    \end{cases*}    
\end{equation}
which enforces sign agreement when the ground truth exhibits a meaningful directional effect, and requires only a small explainer response when the true effect is negligible.

Aggregating across instances $i$ and perturbations $\delta \in \Delta$, the \emph{direction alignment score} for feature $j$ is defined as
\begin{equation}\label{eqn:direction}
    \rho_{\text{dir},j}
    =
    \frac{1}{|S|\cdot|\Delta|}
    \sum_{i=1}^{|S|}
    \sum_{\delta\in\Delta}
    I^{\text{dir}}_{ij}(\delta),
\end{equation}
where $S$ denotes the evaluation set. The score $\rho_{\text{dir},j} \in [0,1]$ measures how reliably the explainer captures the correct \emph{direction of change} of $Y$ along feature $j$ across the dataset.

\subsubsection{Instantiations: SHAP and LIME}

The above definitions are method-agnostic. We now describe how they are instantiated for SHAP and LIME in our experiments.

\paragraph{SHAP}
For SHAP, $e_{\mathcal{E}}(\mathbf{x}^{(i)}, j)$ corresponds to the Shapley value of feature $j$ for instance $\mathbf{x}^{(i)}$. For each $\delta \in \Delta$, we evaluate the SHAP explanation 
$e_{\mathcal{E}}(\mathbf{x}^{(i)}_{j,\delta}, j)$ and the corresponding true outcome 
$\mathcal{G}(\mathbf{x}^{(i)}_{j,\delta})$, yielding 
$\Delta e^{(i)}_j(\delta)$ and $\Delta Y^{(i)}_j(\delta)$ as defined above.
This instantiation directly tests whether the SHAP interpretation (that larger Shapley values correspond to larger model predictions) extends beyond the model's learned relationship $\mathbf{X} \mapsto \hat{Y}$ to the true data-generating relationship $\mathbf{X} \mapsto Y$.

\paragraph{LIME}
For LIME, the explainer represents the local influence of feature $j$ through the coefficient $w^{(i)}_j$ of a locally fitted surrogate model around instance $\mathbf{x}^{(i)}$. We therefore consider small perturbations $\delta$ so that the evaluation remains within the local neighborhood of $\mathbf{x}^{(i)}$. In this case,
\[
    \Delta e^{(i)}_{j}(\delta) = w^{(i)}_j \cdot \delta,
\]
where $w^{(i)}_j$ is the LIME coefficient for feature $j$ at instance $i$. In our experiments, we set $\delta = 0.1\,\texttt{Std}$.

\subsubsection{Strength Alignment}

Strength alignment evaluates whether an explainer correctly captures the relative and undirected importance of features in the data-generating process. Unlike direction alignment, which is assessed at the feature level, strength alignment is a global metric that compares feature-importance rankings aggregated across the dataset.

Let $I_{\mathcal{E}}(j)$ denote the global importance of feature $j$ as implied by explainer $\mathcal{E}$, and let $I_{\mathcal{G}}(j)$ denote the ground-truth importance of feature $j$ under the data-generating process $\mathcal{G}$. We rank features under each scheme to obtain rankings $R_{\mathcal{E}}(j)$ and $R_{\mathcal{G}}(j)$, respectively. The \emph{strength alignment} score is defined as the Spearman Rank Correlation \citep{zar1972significance} between these two rankings:
\[
\rho_{\text{strength}}(R_{\mathcal{E}}, R_{\mathcal{G}})
=
1 - 
\frac{
6\sum_{j=1}^d (R_{\mathcal{E}}(j) - R_{\mathcal{G}}(j))^2
}{
d(d^2 - 1)
},
\]
which takes values in $[-1,1]$. A value close to $1$ indicates strong agreement between the explainer and the ground truth, while values near $0$ or negative indicate weak or systematically incorrect alignment. Because this metric compares importance rankings rather than directional effects, it evaluates exclusively the explainer's ability to recover the \emph{relative} contributions of features in the underlying model.

\paragraph{SHAP}
For SHAP, global feature importance is typically defined as the \textit{mean absolute SHAP value} of each feature across the dataset. To maintain consistency with the unified framework, we define the ground-truth importance $I_{\mathcal{G}}(j)$ using the mean absolute Shapley values computed from the ground-truth model $\mathcal{G}$. Therefore, a strength alignment score for SHAP compares whether the explainer ranks features correctly (based on feature rankings from the true data-generation process). 

\paragraph{LIME}
For LIME, 
The ground-truth importance $I_{\mathcal{G}}(j)$ is defined analogously by applying LIME to the ground-truth model $\mathcal{G}$ and aggregating absolute coefficients across instances, which are compared against LIME applied to the model $\mathcal{M}$. Substituting these quantities into the general formulation above yields the strength alignment score for LIME.

\subsection{Data Simulation}
\label{sec:data}

To evaluate how well explanation-based measures align with the true relationship between $\mathbf{X}$ and $Y$, we use simulated data in all experiments.

Each dataset is generated by sampling feature values for 5{,}000 instances and applying a ground-truth data-generating process $\mathcal{G}$ to produce the outcome variable. The process $\mathcal{G}$ combines untransformed feature terms, nonlinear (squared) terms, and pairwise interaction terms in an additive index, with randomly sampled coefficients.

We systematically vary four scenario-defining factors—\texttt{Number of features}, \texttt{Correlation strength}, \texttt{Nonlinear terms}, and \texttt{Interaction terms}—while holding sample size and noise variance fixed. Each factor takes three levels, yielding a total of \textbf{81 datasets}, as summarized in Table~\ref{tab:data_factors}. The complete data-generation procedure is provided in Algorithm~\ref{alg:synth-data} in Appendix~\ref{apx_sec:81_details}.


For each dataset, we follow a standard explanation pipeline by first training a predictive model $\mathcal{M}$ on 80\% of the data and evaluating it on the remaining 20\%. The best model is obtained via a systematic search over multiple model families, including gradient-boosted decision trees (e.g., XGBoost, CatBoost, LightGBM, Gradient Boosting) and other commonly used learners (e.g., Random Forests, multilayer perceptrons, and support vector machines). Within each model class, hyperparameters are tuned using cross-validation, and the model achieving the highest validation performance is selected. 

\subsection{Evaluation Results — \textit{Can Post hoc Explainers Recover Feature Direction and Strength from Data}?}

\begin{figure}[ht!]
        \centering
  \begin{subfigure}[t]{.35\textwidth}
    \centering
    \includegraphics[width=\linewidth]{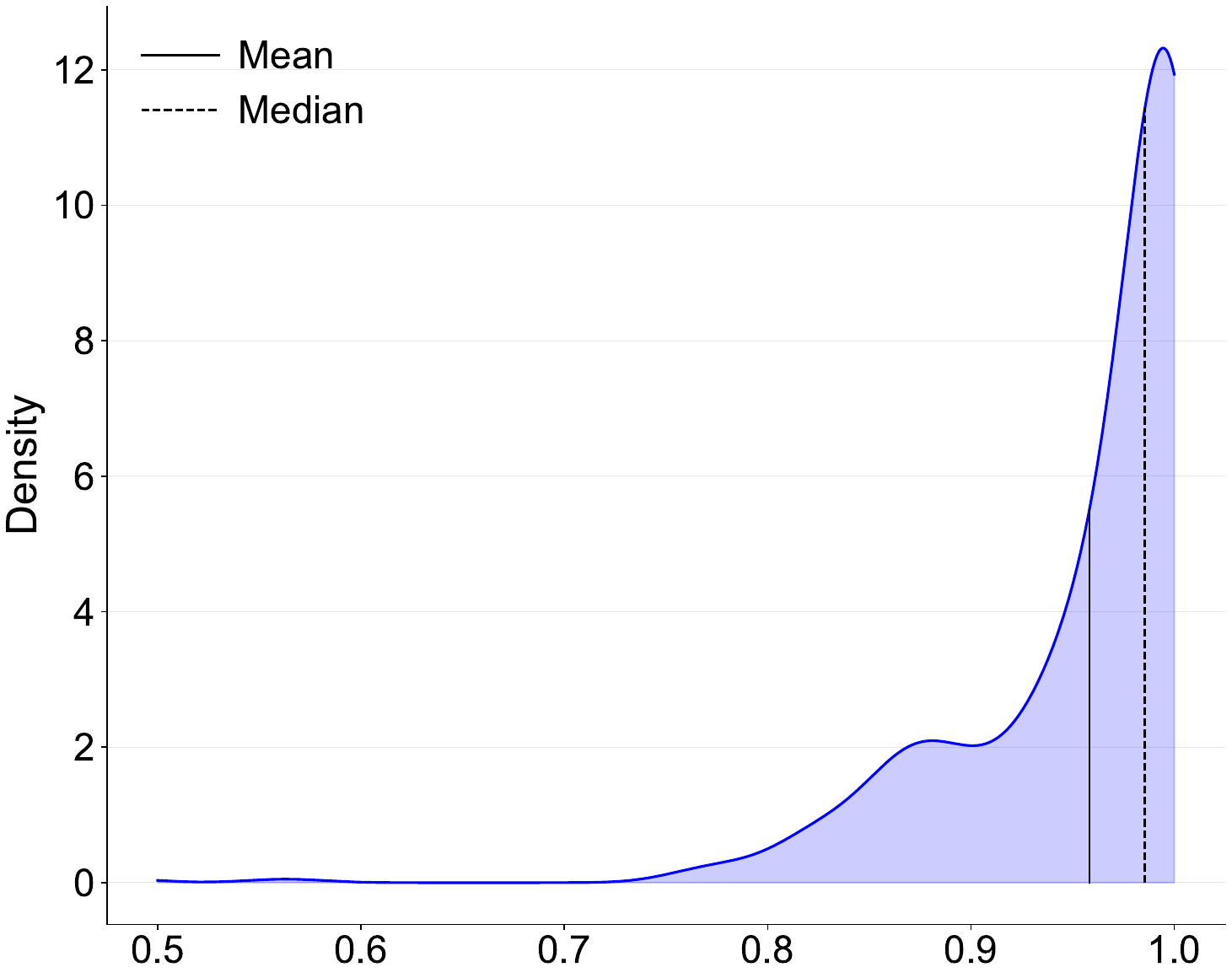}
    \caption{SHAP direction alignment}
    \label{fig:shap_sec4_direction}
  \end{subfigure}
  \begin{subfigure}[t]{.35\textwidth}
    \centering
    \includegraphics[width=\linewidth]{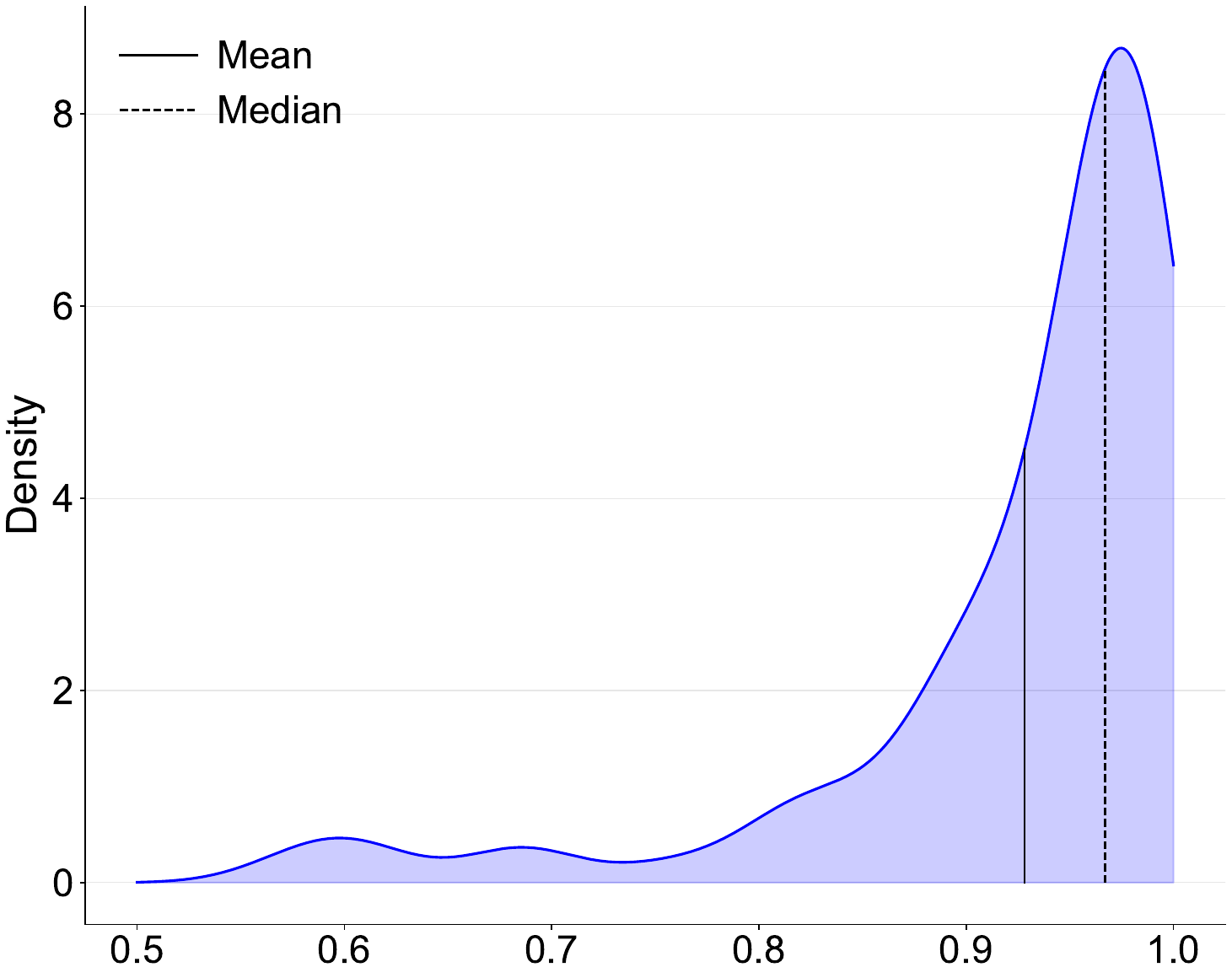}
    \caption{SHAP strength alignment}
    \label{fig:shap_sec4_strength}
  \end{subfigure}
  \smallskip

  \begin{subfigure}[t]{.35\textwidth}
    \centering
    \includegraphics[width=\linewidth]{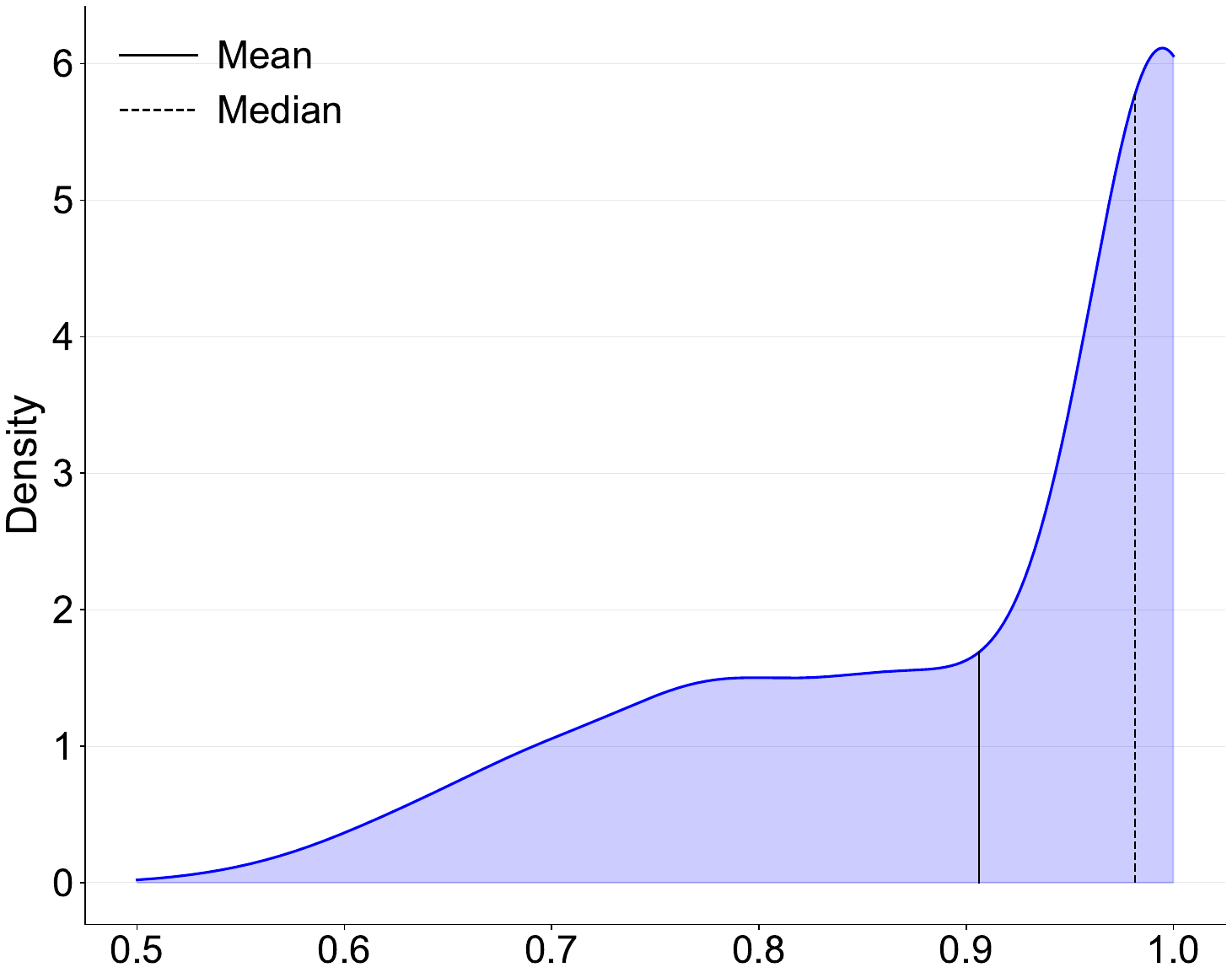}
    \caption{LIME direction alighment}
    \label{fig:lime_sec4_direction}
  \end{subfigure}
  \begin{subfigure}[t]{.35\textwidth}
    \centering
    \includegraphics[width=\linewidth]{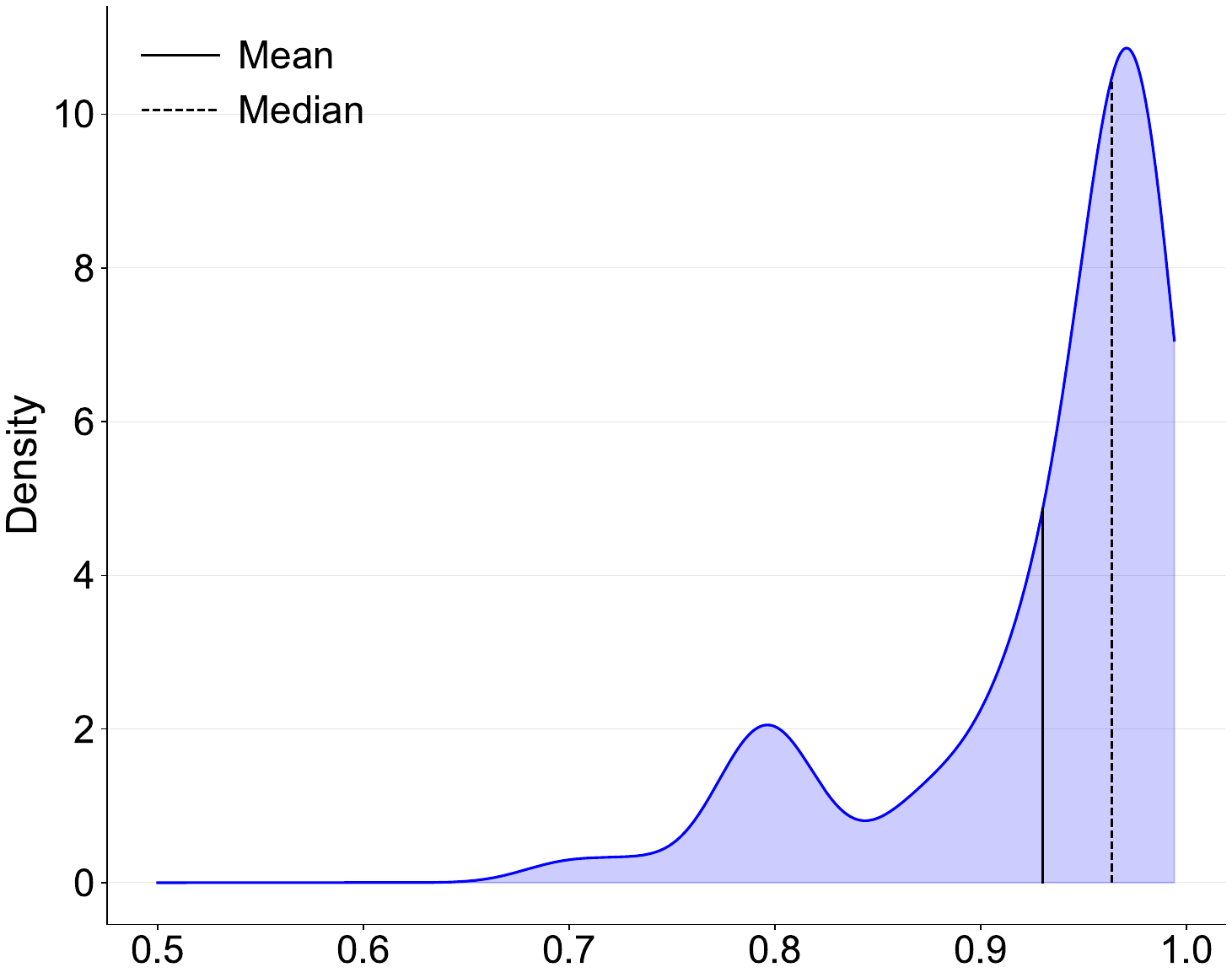}
    \caption{LIME strength alignment}
    \label{fig:lime_sec4_strength}
  \end{subfigure}
  \caption{Evaluation of direction and strength metrics for explanations of the best models for all 81 datasets by SHAP and LIME }\label{fig:sec4_best81}
\end{figure}

We evaluate SHAP and LIME using the direction and strength alignment metrics defined in Section~\ref{sec:metrics}. For each dataset–model pair, we summarize performance by averaging alignment scores over the five features with the largest mean explanation values, and then examine the distribution of these  alignment scores across all datasets. The probability density functions (PDFs) of direction and strength alignment are shown in Figure~\ref{fig:sec4_best81}. The results yield two main findings.

\noindent\textbf{Finding 1: High Average Performance.}
Across datasets, post hoc explainers exhibit high average alignment with the ground-truth data-generating process in both direction and strength. In particular, SHAP consistently outperforms LIME, indicating a stronger ability to recover the direction of feature effects on average.

\noindent\textbf{Finding 2: Long Left Tails Undermine Dataset-Level Reliability.}
For both direction and strength alignment, the full distributions exhibit pronounced long left tails, indicating substantial heterogeneity in explainer performance across datasets. Although average alignment is high, a nontrivial subset of dataset–model pairs exhibits markedly poorer alignment.

The nature of these failures differs across explainers and metrics. For SHAP, strength alignment displays a substantially longer and heavier left tail than direction alignment, with values dropping to approximately 0.5 in the worst cases, indicating severe mis-ranking of feature importance for certain datasets. In contrast, for LIME, direction alignment exhibits the more pronounced left tail, with values also falling to approximately 0.5, corresponding to near-random agreement under a binary directional criterion. Strength alignment for LIME, while variable, is comparatively less extreme.

These long left tails imply that high average alignment masks meaningful risks at the dataset level. In particular, explainer outputs can be unreliable for specific dataset–model pairs, even when overall performance appears strong. Because users typically lack reliable ex ante indicators of when an explainer is operating in these low-alignment regimes, the presence of heavy left tails raises concerns about the robustness of post hoc explanations for drawing conclusions about the underlying $X \mapsto Y$ relationship.

\section{What Factors Influence Alignment?}\label{sec:factors}
The preceding section establishes that alignment between post hoc explanations and the true data-generating process varies substantially across dataset–model pairs, even when average performance appears strong. We now turn to understanding the sources of this variation. In particular, we examine which factors systematically influence whether explanations recover the correct direction and strength of feature effects.

Our analysis focuses on three classes of factors. First, we study the predictive performance of the learned model $\mathcal{M}$, which serves as a practical indicator of how closely $\mathcal{M}$ approximates the underlying data-generating process $\mathcal{G}$. Second, we examine the role of model multiplicity through the Rashomon effect, which highlights how multiple, equally accurate models can encode {fundamentally different internal representations} of the same task. Finally, we investigate how intrinsic properties of the data—such as feature correlation, nonlinearity, and interaction structure—shape alignment by constraining or expanding the space of plausible models.
These analyses clarify when alignment failures arise from limitations of the predictive model, when they stem from  ambiguity among high-performing models, and when they are driven by structural properties of the data itself.

\subsection{Predictive performance of $\mathcal{M}$ }
        \label{sec:role_of_predictive_power}
Post hoc explainers such as SHAP and LIME explain the predictive model $\mathcal{M}$, not the underlying data-generating process $\mathcal{G}$ directly. For explanations to reflect the true $\mathbf{X} \mapsto Y$ relationship, the model $\mathcal{M}$ must approximate $\mathcal{G}$ sufficiently well. Predictive accuracy is therefore a necessary condition for alignment: when predictive accuracy is low, $\mathcal{M}$ fails to capture key aspects of $\mathcal{G}$, making alignment with $\mathcal{G}$ unattainable for any post hoc explainer.

To examine how alignment varies with model quality, we construct multiple predictive models for each dataset spanning a wide range of test accuracies. These models are generated by varying both hyperparameter configurations and the size of the training data. We categorize test accuracy into four ranges: $[0.70,0.75)$, $[0.75,0.80)$, $[0.80,0.85)$, and $[0.85,0.90)$. For each accuracy range, we continue generating models until we obtain three models whose test accuracies fall within the corresponding interval. This procedure yields a balanced set of models across accuracy levels for each dataset. We then evaluate direction and strength alignment for each model and report the mean alignment and standard error within each accuracy range. This design allows us to isolate the relationship between predictive performance and explanation alignment while holding the underlying data-generating process fixed.

 \begin{figure}[h]
\centering
\includegraphics[width=0.7\linewidth]{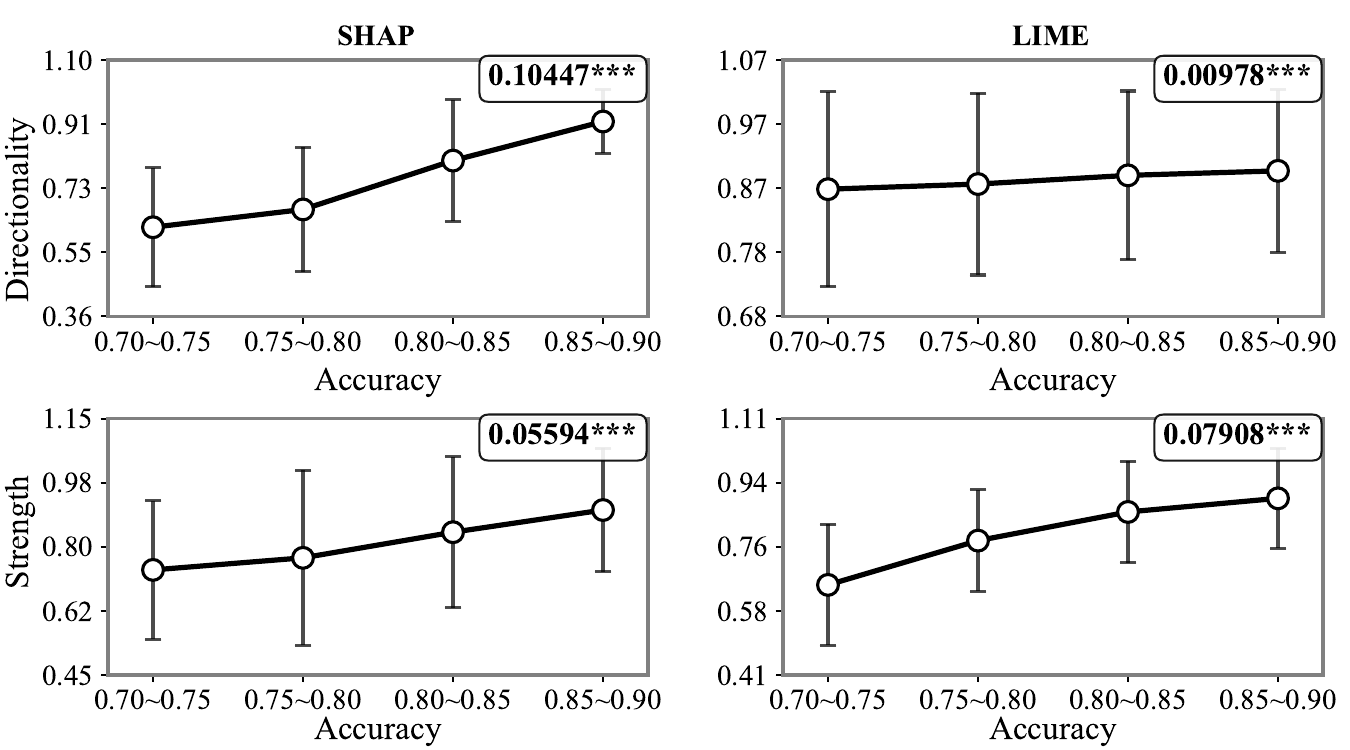}
\caption{The impact of model accuracy on explanation alignment.}\label{fig:sec5_accuracy}
\end{figure}

Figure~\ref{fig:sec5_accuracy} shows that model accuracy has a statistically significant impact on explanation alignment, particularly for SHAP’s direction alignment. The number reported in the top-right corner of each plot is the coefficient from a univariate OLS regression, where each accuracy bucket is coded as an ordered category (1, 2, 3, and 4). Accordingly, the coefficient represents the average change in the alignment metric associated with a one-bucket increase in accuracy, corresponding to an accuracy improvement of 0.05. For example, a 0.05 increase in test accuracy is associated with an average increase of 0.104 in SHAP’s direction alignment.

\subsection{Rashomon Effect}\label{sec:rashomon_effect}

While model accuracy has significant impact, it is \textit{necessary but insufficient} to  guarantee alignment between the learned model $\mathcal{M}$ and the true data-generating process $\mathcal{G}$. This possibly unintuitive limitation arises from the presence of a \emph{Rashomon set}: a collection of models that achieve similarly strong predictive performance on the same task. When a Rashomon set is large, many distinct models map $X \mapsto \hat{Y}$ with comparable accuracy while relying on substantially different internal decision paths or feature relationships. In such settings, the accuracy of $\mathcal{M}$ does not uniquely identify the underlying structure of $\mathcal{G}$. Consequently, even a highly accurate model may yield post hoc explanations that reflect only one of many plausible decision-making mechanisms consistent with the observed data, rather than the true $X \mapsto Y$ relationship. This multiplicity poses a fundamental challenge for inference. When many viable $X \mapsto \hat{Y}$ mappings exist, explanations derived from any single predictive model cannot reliably recover the true $\mathbf{X} \mapsto Y$. As a result, post hoc explanations may vary substantially across models that are  equivalent in terms of predictive performance.

\begin{figure}
\centering
  \begin{subfigure}[t]{.47\textwidth}
    \centering
    \includegraphics[width=\linewidth]{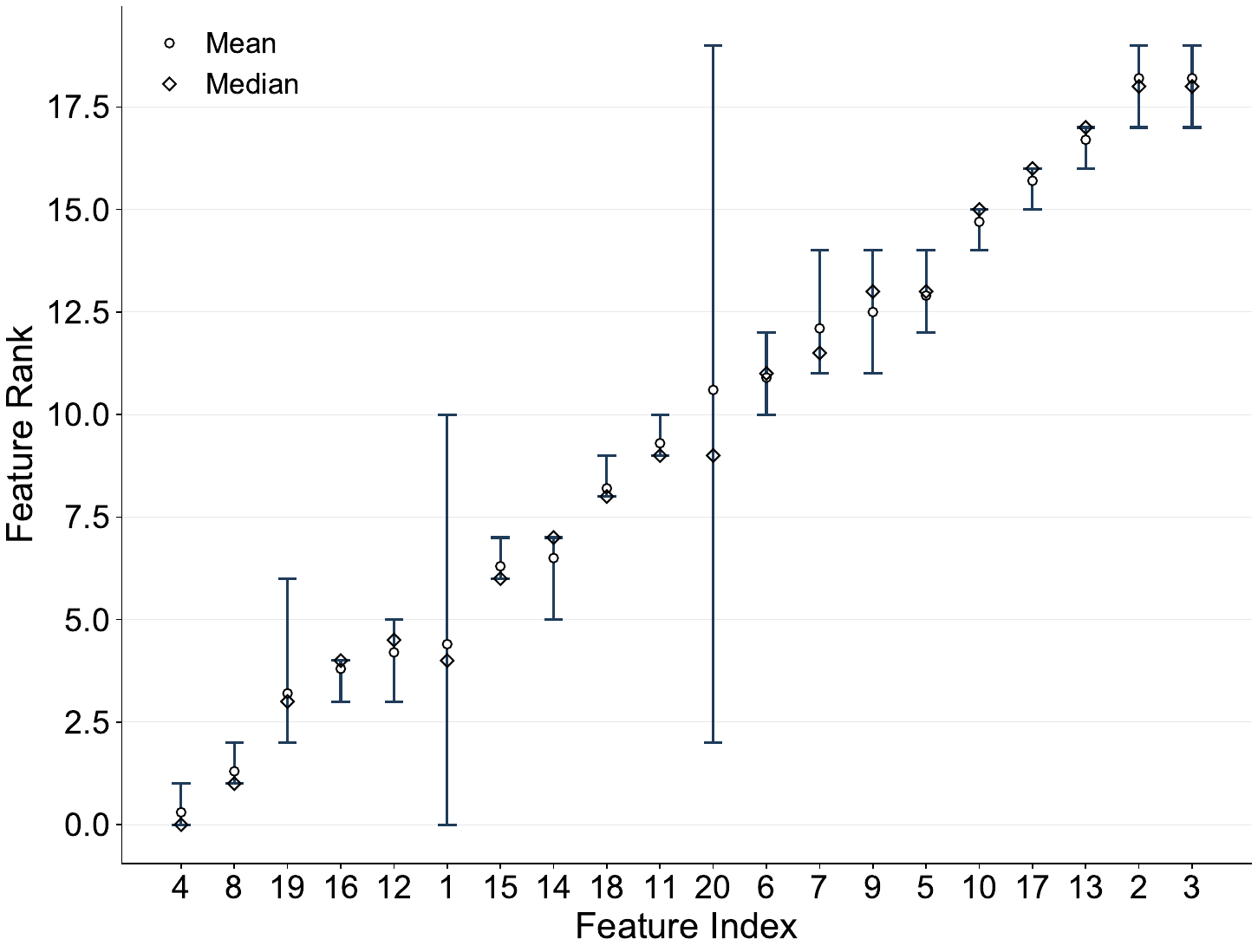}
    \caption{The maximum difference of ranks by SHAP for models in a Rashomon Set in one dataset}\label{fig:dataset_ranking}
  \end{subfigure}
\hfill
  \begin{subfigure}[t]{0.47\textwidth}
    \centering
    \includegraphics[width=\linewidth]{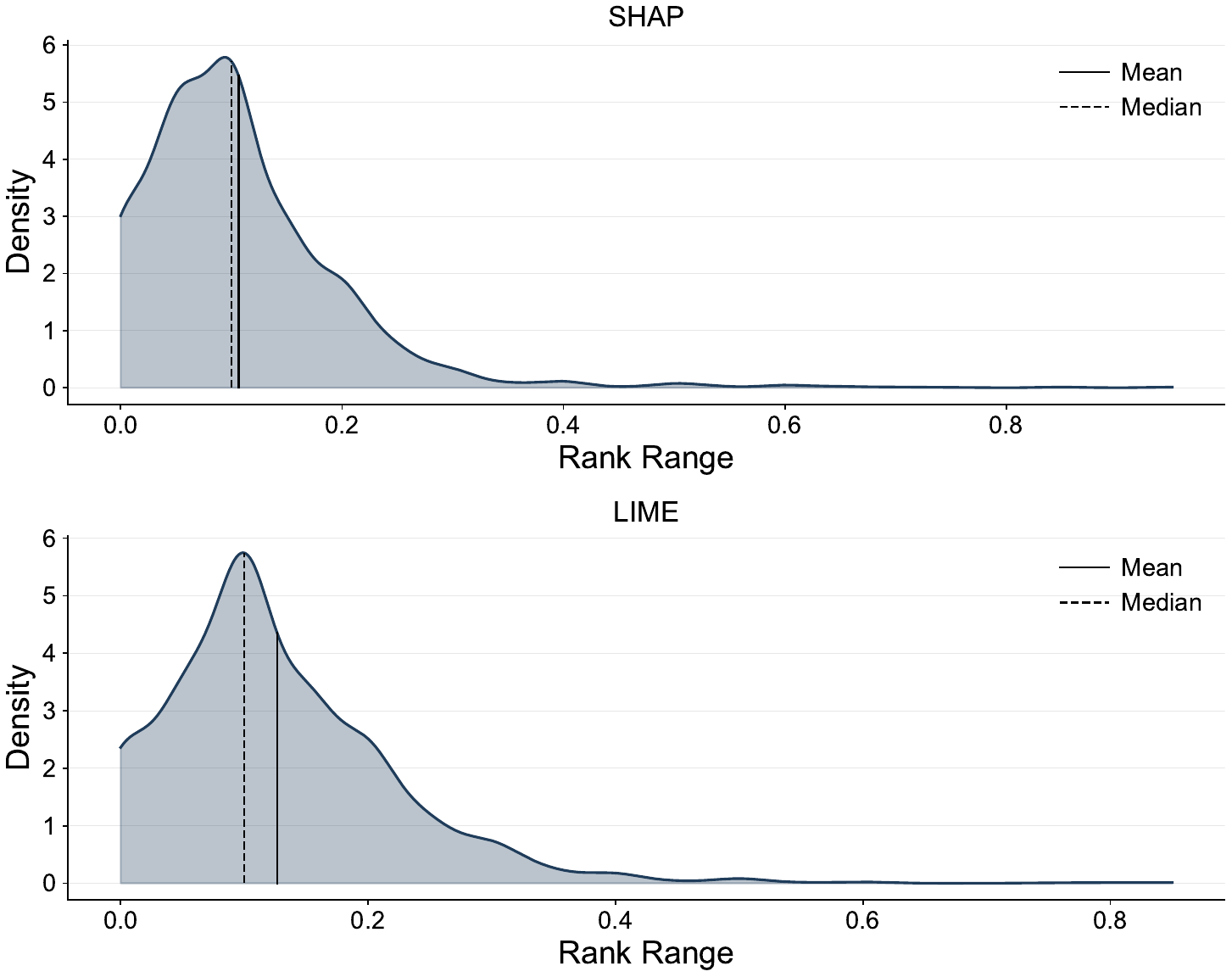}
    \caption{The distribution of differences of feature ranks}
  \end{subfigure}
  \caption{The evidence of Rashomon Effect on Post hoc explanations}\label{fig:range_dist}
\end{figure}

To empirically illustrate explanation multiplicity induced by the Rashomon effect, we obtained 10 best-performing models for each of our 81 datasets. For each dataset, the selected models achieve test accuracies within 3\% of one another. We then applied SHAP to each model and ranked features according to their mean absolute Shapley values. This procedure yields 10 feature-importance rankings per dataset, one for each predictive model. 

We plot in Figure \ref{fig:dataset_ranking} the range of feature rankings for a representative dataset. Each line spans from the minimum to the maximum rank assigned to a feature across the 10 models, with features ordered by their median rank. If the models’ internal decision-making processes were highly consistent, we would expect minimal variation across these rankings. However, Figure~\ref{fig:dataset_ranking} shows that while many features exhibit modest variability, others show substantial disagreement. For instance, in the demo dataset we picked from the 81 datasets, Feature~0 is ranked as both the most important feature (rank 1) and relatively unimportant (rank 11 out of 20) by different models, despite all models achieving nearly identical predictive accuracy.

To provide a holistic view of this phenomenon, we compute the ranking variation for all features across all datasets and visualize the resulting distribution. To ensure comparability across datasets with different numbers of features, we normalize the ranking range by the total number of features, yielding values between 0 and 1. Figure~\ref{fig:range_dist} shows that, on average, feature rankings vary by approximately 10\%, with a pronounced right tail. This indicates that for a nontrivial subset of features, ranking disagreement can be substantial, further underscoring the instability of post hoc explanations within large Rashomon sets.
As a consequence, high accuracy alone cannot guarantee alignment between explanations and the true data-generating mechanism. The Rashomon effect ensures that this misalignment persists even in best-case predictive scenarios, as multiple distinct decision rules remain equally compatible with the observed data.

Nevertheless, \emph{high predictive accuracy is necessary while not sufficient for alignment}. While high accuracy does not imply that $\mathcal{M}$ aligns with $\mathcal{G}$, low accuracy provides strong evidence that $\mathcal{M}$ fails to represent $\mathcal{G}$, and therefore that explanations of $\mathcal{M}$ cannot align with the truth. This asymmetry motivates our investigation into how varying levels of model accuracy influence alignment.

\subsection{Data Properties}

In addition to predictive performance and model multiplicity, we examine how intrinsic properties of the data influence explanation alignment. Data properties are directly observable characteristics that empirically predict when and how severely explanation misalignment occurs.

\begin{figure}[h]
\centering
  \begin{subfigure}[t]{.95\textwidth}
    \centering
    \includegraphics[width=\linewidth]{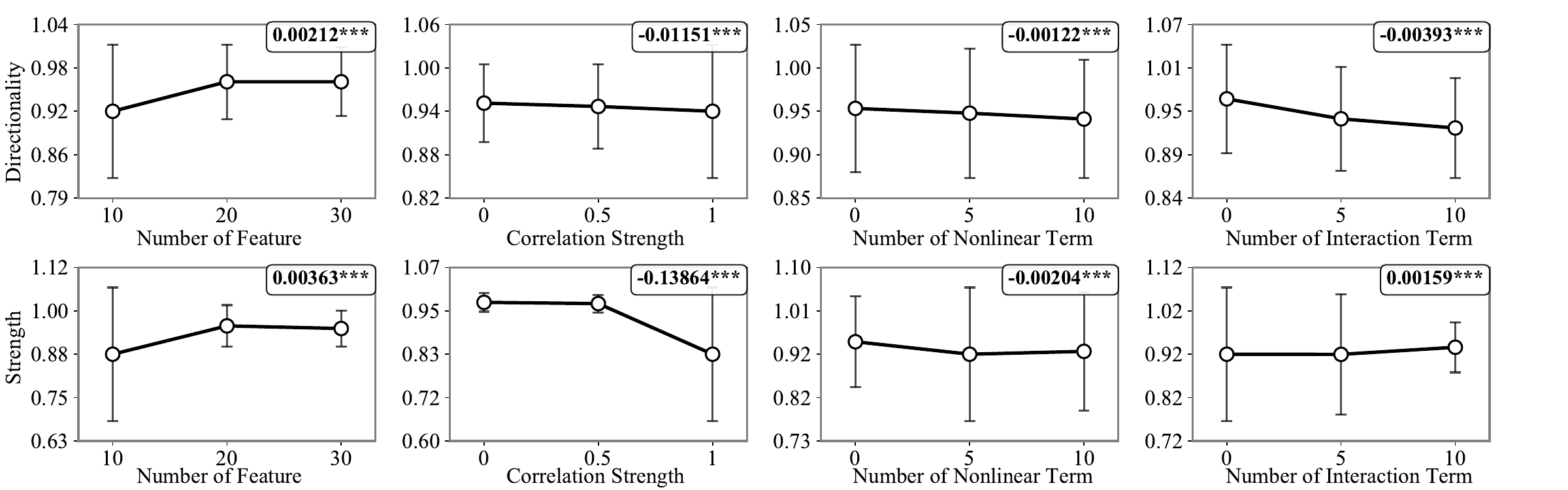}
    \caption{Impact of data characteristics on SHAP alignment}
  \end{subfigure}

  \begin{subfigure}[t]{0.95\textwidth}
    \centering
    \includegraphics[width=\linewidth]{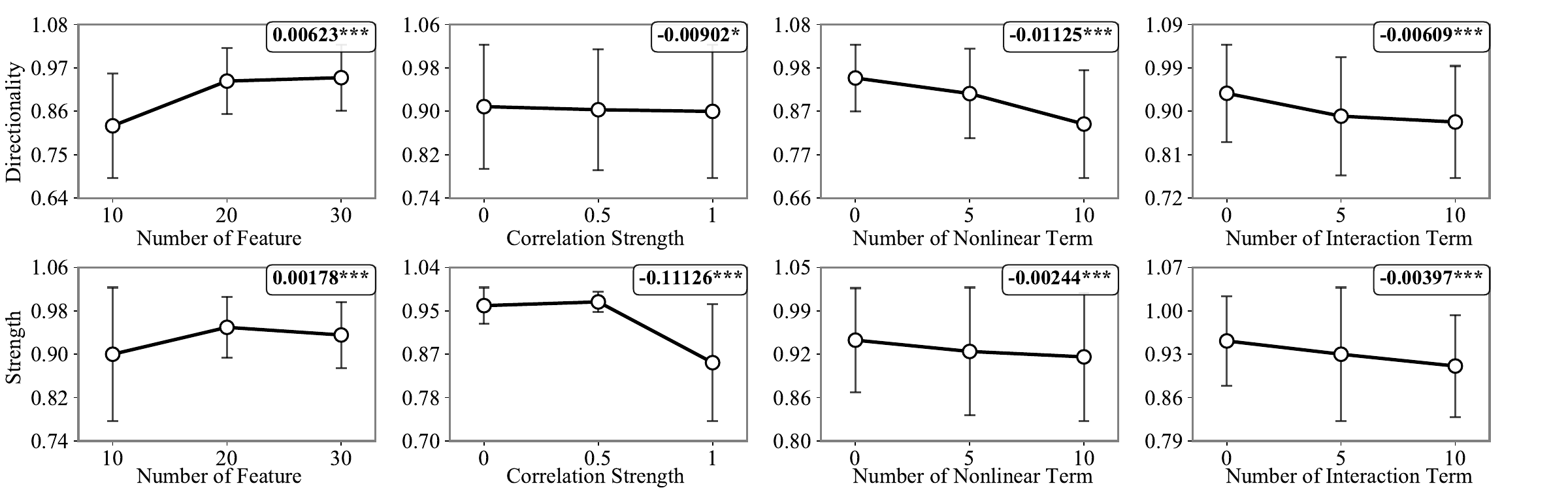}
    \caption{Impact of data characteristics on LIME alignment}
  \end{subfigure}
  \caption{Impact of data characteristics on post hoc explanation alignment. Numbers on the top right are the coefficients of the univariate OLS analyses.}
  \label{fig:alignment_data}
\end{figure}

To visualize the effect, we plot the direction and strength alignment for datasets with different factors, and the coefficients in the OLS regression. Figure~\ref{fig:alignment_data} shows that feature correlation strength, the number of nonlinear terms, and the number of interaction terms are statistically associated with reduced alignment. 

When features are highly correlated, multiple variables can act as substitutes or proxies for one another, enabling models with substantially different internal attributions to achieve comparable predictive accuracy. Empirically, this is associated with greater disagreement in post hoc explanations. A plausible explanation is that feature substitutability increases the number of near-optimal models consistent with the data, making the learned model’s internal attribution more sensitive to arbitrary modeling choices.

A similar mechanism applies to the number of nonlinear terms and the number of interaction terms. As the data-generating process becomes more complex, a larger set of functional forms can fit the observed data equally well, increasing model flexibility without a commensurate increase in identifiability. As a result, learned models that achieve comparable predictive performance may rely on qualitatively different internal representations, leading to greater variability in post hoc explanations.

Overall, these results suggest that explanation misalignment is not driven solely by explainer choice or model performance, but is systematically related to intrinsic properties of the data. In particular, feature correlation, nonlinearities, and interactions amplify ambiguity in how predictive relationships are encoded, making reliable recovery of the underlying $X \mapsto Y$ relationship more difficult even when predictive accuracy is high.

One counterintuitive observation is that the number of features is associated with a small positive effect on explanation alignment. However, the magnitude of this effect is negligible (below 1\%) and substantially smaller than the effect of feature correlation strength. We therefore do not interpret this result as evidence that higher dimensionality improves explanation faithfulness. 

\section{Robustness Checks}\label{sec:robustness}

The preceding results show that data misalignment in post hoc explanations arises systematically rather than as an artifact of a particular modeling choice. We now examine whether this conclusion depends on specific implementation or evaluation decisions, or whether misalignment persists even under conditions designed to be favorable to alignment. To do so, we conduct three robustness checks that probe aggregation across models, sensitivity to explainer hyperparameters, and sensitivity to evaluation metrics.

Each check is motivated by a natural stabilizing intuition: that averaging across equally accurate models may recover a more faithful signal; that careful explainer tuning may reduce approximation noise; and that alternative metric specifications may yield different conclusions. Across all settings, we find that these intuitions fail to overturn our main result. Even under favorable configurations, data misalignment remains persistent. Full experimental details are reported in Appendix~\ref{apx_sec:robustness_results}.

\subsection*{Robustness Check 1: Averaging over a Rashomon Set}

The Rashomon effect implies that many distinct models can achieve near-identical predictive performance while relying on different internal decision logic. This raises the possibility that while any single model’s explanation may be misaligned, averaging explanations across multiple equally accurate models could yield a more faithful representation of the data-generating process.

To test this idea, we construct, for each dataset, a sample of a Rashomon set \footnote{Note that a Rashomon set could contain tens of thousands of models. Here we construct only a small set to illustrate the idea.} consisting of the ten highest-performing models drawn from diverse model families. All selected models lie within a 3\% absolute accuracy margin of the best-performing model. For each test instance, we average instance–feature attributions across models and evaluate alignment using the direction and strength metrics defined in Section~\ref{sec:main_alignment}. This test assesses whether consensus across plausible models mitigates misalignment.

\subsection*{Robustness Check 2: Sensitivity to Explainer Hyperparameters}
\label{sec:lime_params}

Post hoc explainers introduce additional layers of approximation of $\mG$ (beyond that of $\mM$) through sampling schemes, surrogate models, and algorithmic hyperparameters. We examine whether misalignment is driven by such explainer-induced noise.
For SHAP, we vary the number of bootstrap samples. For LIME, we vary kernel bandwidths and neighborhood sizes, motivated by theorems in Appendix \ref{apx_sec:lime_thms}. Because LIME relies on randomized perturbations, we also average explanations across multiple random seeds while holding all other parameters fixed. This analysis isolates whether stabilizing explainer-level variability improves alignment independently of model behavior.

\subsection*{Robustness Check 3: Sensitivity to Metric Parameters}

Our direction alignment metric depends on a perturbation magnitude $\delta$ that determines the scope of feature perturbations. Smaller perturbations emphasize local behavior, while larger perturbations probe broader directional effects. We vary $\delta$ while holding all other components of the evaluation fixed and recompute alignment scores. This test assesses whether our conclusions hinge on a particular perturbation scale or persist across reasonable metric specifications.

\subsection*{Results}
\label{sec:summary_and_strategies}

Plots summarizing the results of our robustness checks are reported in Appendix~\ref{apx_sec:robustness_results}; here we highlight the main findings. Consistent with our primary analyses, we continue to observe persistent misalignment between post hoc explanations and the ground-truth data-generating process. In particular, alignment distributions exhibit pronounced left tails, indicating that severe failures in recovering both direction and strength occur with nontrivial frequency.

First, varying the perturbation step size in $\Delta$ does not materially alter the shape of the alignment distributions, suggesting that our conclusions are robust to reasonable choices of this metric parameter.
Second, certain aggregation strategies yield modest improvements in alignment, though none eliminate misalignment entirely. Averaging explanations across models within a Rashomon set improves direction recovery for SHAP, consistent with recent findings \citep{donnelly2023rashomon}. Similarly, averaging LIME explanations across multiple runs with different random seeds improves both direction and strength alignment.
Third, improved parameter choices within the explainer can lead to small gains in alignment but again do not resolve misalignment. In particular, the choice of kernel bandwidth has a statistically significant effect on alignment outcomes.

Overall, these results reinforce our central conclusion: while aggregation and careful parameter tuning can slightly improve alignment, post hoc explanations remain unreliable for inferring the true $\mathbf{X} \mapsto Y$ relationship given the long left tail in the distribution. Therefore, even under favorable conditions, explanation outputs should be interpreted with caution.

\section{Diagnosing Explanation Reliability via Rashomon Sets}\label{sec:rashomon_agreement}

The results in the previous sections demonstrate that post hoc explanations are not uniformly reliable in recovering true effects in data: while they perform well on average, they can fail in a non-negligible fraction of datasets. This raises a natural and practically important question: \textit{when should a post hoc explanation be trusted, and when should it be treated with caution}? In this section, we explore whether disagreement among equally high-performing models in a Rashomon set can serve as a signal of the risk associated with trusting a given post hoc explanation.

\subsection{Our Hypothesis: Rashomon Agreement Signals Explanation Alignment}
As discussed previously, the models comprising a Rashomon set achieve highly similar predictive performance on the observed data, making them effectively indistinguishable under standard evaluation metrics. Despite this similarity in accuracy, these models may rely on substantially different internal representations and decision pathways to map $X$ to $\hat{Y}$.  Figure \ref{fig:range_dist} illustrates such multiplicity reflected as  the disagreement in post hoc explanations for models in a Rashomon Set.

We refer to the degree of consistency among models in a Rashomon set as \textit{Rashomon agreement}. We hypothesize that, if models in a Rashomon set  exhibit low agreement in the sense that they disagree in their predictions or explanations, this disagreement provides evidence that \textbf{the space of hypotheses compatible with the data is relatively large}. In such settings, any single model represents only one of many plausible explanations for how $X \mapsto Y$, and the likelihood that its post hoc explanation aligns closely with the true data-generating process $\mathcal{G}$ is correspondingly reduced. Conversely, high Rashomon agreement among models with similar predictive performance suggests a more constrained hypothesis space, increasing confidence that the explanations produced by any one model reflect stable, data-supported structure rather than artifacts of model choice.

\subsection{Measuring Rashomon Agreement}

To operationalize Rashomon agreement, we must confront the fact that agreement among models in a Rashomon set cannot be observed directly: the models may differ in architecture, parameterization, or internal representations, making a direct comparison of their learned functions infeasible. Instead, we assess agreement through observable manifestations of the underlying $\mathbf{X} \mapsto Y$  mapping. In particular, agreement can be revealed through (i) the models’ predictions, which reflect their externally observable behavior, and (ii) their post hoc explanations, which provide a window into how each model attributes importance to features when producing those predictions. We therefore define two complementary agreement metrics—one based on predictions and one based on explanations—to capture these distinct but related notions of consistency within a Rashomon set. Agreement is computed pairwise between models in the Rashomon set and then averaged over all pairs to obtain a dataset-level agreement score.

\noindent\textbf{Prediction Agreement}
We begin with agreement metrics based on model predictions, since predictions are the primary externally observable outputs of a model and the basis on which models in a Rashomon set are deemed equally accurate. If two models implement similar $\mathbf{X}$ to $\hat{Y}$ mappings, this similarity should be reflected in how often they produce the same predicted labels on the same instances. Given two models, we define the prediction agreement metric as \textit{the fraction of instances on which two models agree}. Larger values correspond to greater agreement. 

\noindent\textbf{Explanation Agreement}
While prediction agreement captures similarity in observable behavior, it does not reveal whether models rely on similar feature attributions to generate those predictions. To assess agreement in the inferred structure of the $\mathbf{X}\mapsto Y$ mapping, we next define agreement metrics based on post hoc explanations.

Because explanation values are real-valued, and can vary substantially in scale across models and explainers, we focus on rank-based measures that are invariant to monotonic transformations. Specifically, for each model we rank features based on the mean absolute explanation values across instances, and define explanation agreement between two models as the Spearman correlation between these vectors. 


\paragraph{Results}

If these agreement metrics are informative, we would expect higher agreement among models in a Rashomon set to correspond to stronger alignment between post hoc explanations and the data-generating process. In other words, Rashomon agreement should be positively correlated with explanation alignment for a given model trained on a dataset.

To evaluate this hypothesis, we examine the correlation between each Rashomon agreement metric and the explanation alignment metrics, along both the direction and strength dimensions. The results are shown in Table \ref{tab:rashomon_agreement_corr}.

\begin{table}[ht]
\centering
\small
\caption{Correlation between Rashomon agreement and explanation alignment}
\label{tab:rashomon_agreement_corr}
\begin{tabular}{l l c c}
\toprule
\textbf{Explainer} & \textbf{Agreement Type} & \textbf{Direction Alignment} & \textbf{Strength Alignment} \\
\midrule
\multirow{2}{*}{SHAP} 
    & Explanation agreement & 0.326 & \textbf{0.792} \\
    & Prediction agreement  & 0.344 & 0.088  \\
\midrule
\multirow{2}{*}{LIME} 
    & Explanation agreement & 0.217 & \textbf{0.695} \\
    & Prediction agreement  & 0.131 & 0.023  \\
\bottomrule
\end{tabular}
\end{table}
\normalsize

Table \ref{tab:rashomon_agreement_corr} reveals two key findings. First, both Rashomon agreement metrics are positively correlated with explanation alignment, indicating that higher agreement among equally accurate models is indeed associated with closer alignment between post hoc explanations and the data-generating structure. This relationship is particularly strong for explanation-based agreement. For SHAP, explanation agreement correlates with alignment by up to approximately 0.792, while for LIME the corresponding correlations reach roughly 0.695. These high correlations suggest that Rashomon agreement provides a strong signal of explanation reliability in recovering true effects in the data.

Moreover, agreement based on explanations is consistently more informative than agreement based on predictions. This difference is expected. Prediction-based agreement captures similarity in model outputs, which can remain high even when models rely on different internal representations or feature attributions to arrive at the same predictions. In contrast, explanation-based agreement directly measures consistency in how models attribute importance to features. As a result, it is a more sensitive indicator of whether explanations reflect stable, data-supported structure rather than model-specific artifacts.

Table~\ref{tab:rashomon_agreement_corr} also highlights an important asymmetry between SHAP and LIME: alignment for SHAP is more strongly correlated with Rashomon agreement than alignment for LIME.  This suggests that SHAP alignment is more predictable from agreement-based signals among equally accurate models, whereas LIME’s alignment is noisier and harder to anticipate, even when models exhibit substantial agreement.

\subsection{Dataset-Level Limits of Post hoc Explanations}

To understand the \textit{dataset-level origins} of the disagreement, we examine how intrinsic data properties—such as dimensionality, feature correlation, nonlinearity, and feature interactions—affect both prediction and explanation agreement. We compute the average agreement for datasets with varying characteristics and fit ordinary least squares (OLS) regressions to quantify the contribution of each factor.

\begin{figure}[H]
    \centering
    \includegraphics[width=\linewidth]{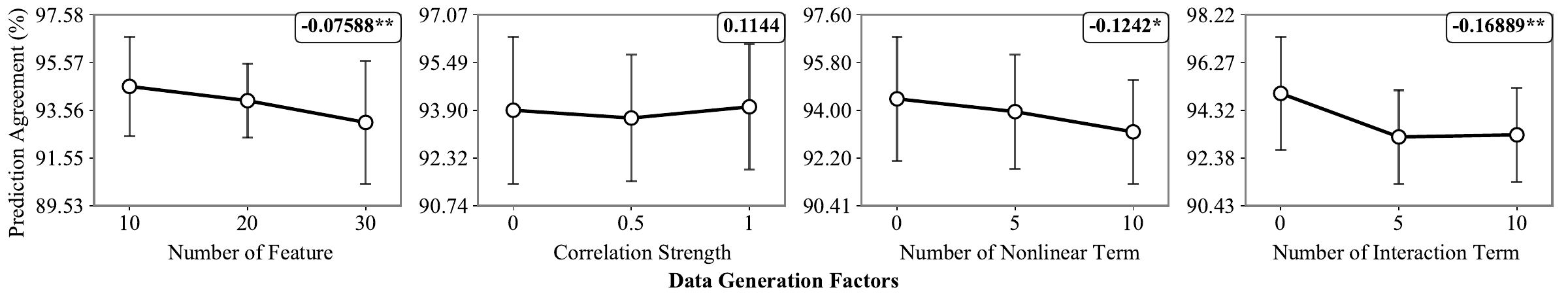}
    \caption{The impact of data properties on prediction agreement.}
    \label{fig:prediction_agreement_data}
\end{figure}

The results for prediction agreement, shown in Fig.~\ref{fig:prediction_agreement_data}, indicate that several data characteristics have a strong and statistically significant impact. The number of interaction terms is the most influential factor: increasing the number of interactions by one is associated with a 16.9\% decrease in prediction agreement. The number of nonlinear terms also has a substantial effect, with an estimated decrease of 12.4\% per additional nonlinear term. These findings suggest that certain forms of data complexity systematically weaken the extent to which accuracy constrains model behavior.

We observe a qualitatively different pattern for explanation agreement. While explanation agreement is strongly correlated with explanation alignment, its relationship with dataset characteristics is weaker and less systematic. Among the factors considered, only feature correlation strength exhibits a statistically significant effect, and this effect is negative but small in magnitude. The remaining factors—dimensionality, nonlinearity, and interaction terms—have no consistent or significant impact. This contrast highlights that explanation agreement is not determined solely by dataset complexity, but also reflects how a particular explainer captures model behavior.

\begin{figure}[h]
    \centering
    \includegraphics[width=\linewidth]{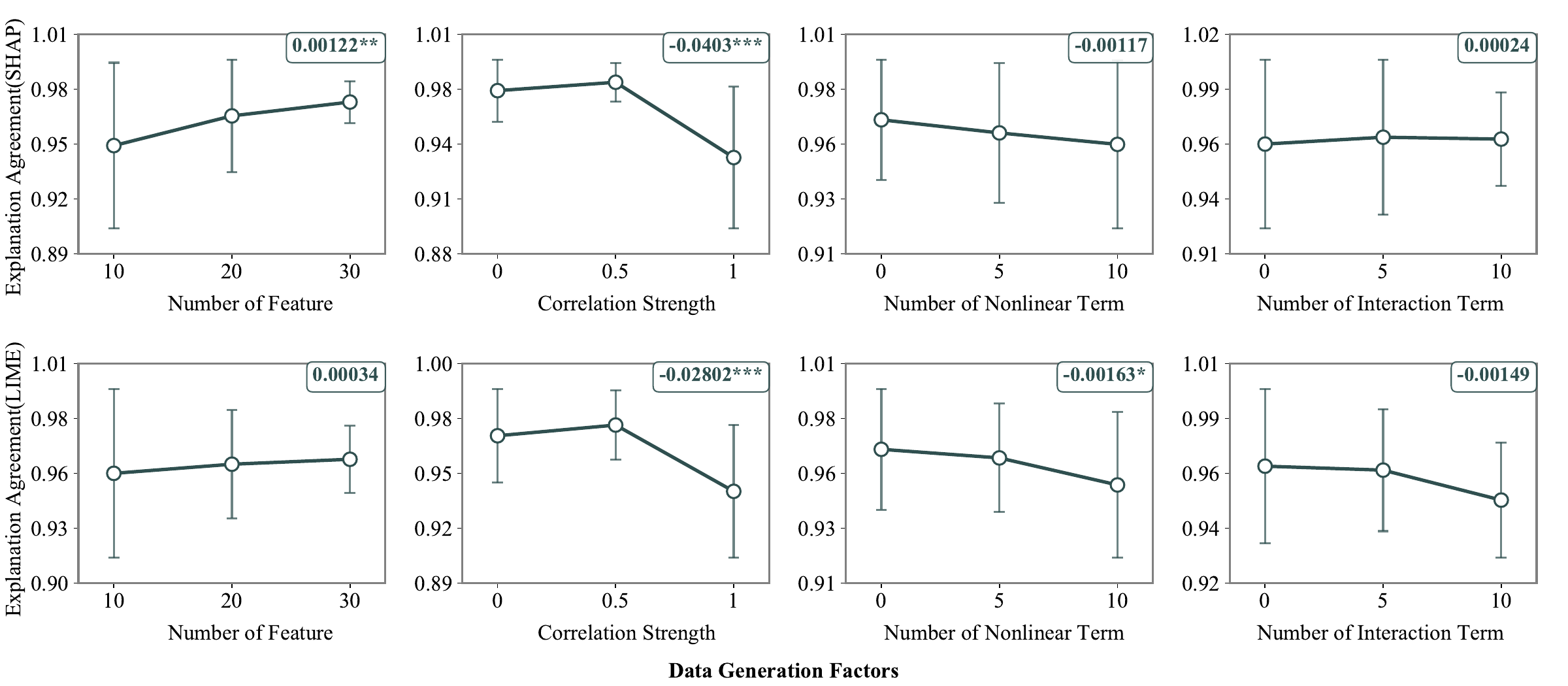}
    \caption{The impact of data properties on explanation agreement}
    \label{fig:explanation_agreement_data}
\end{figure}

\subsection{Discussion}


The results in this section support a central diagnostic message: predictive accuracy is necessary but insufficient for post hoc explanations to reliably recover the true direction and strength of the $\mathbf{X} \mapsto Y$ relationship. When multiple models achieve comparable accuracy yet rely on different internal representations, post hoc explanations are inherently less reliable indicators of the data-generating structure. Rashomon agreement provides a practical way to detect precisely settings where accuracy alone fails to meaningfully constrain the underlying $\mathbf{X}\mapsto Y$ mapping.

From a dataset-level perspective, disagreement among models arises naturally when the observed data admit multiple predictive hypotheses. As data become more complex—through nonlinear relationships, feature interactions, or correlated inputs—there are more distinct ways for models to combine features to achieve comparable predictive performance. In such settings, accuracy does not uniquely pin down how predictive information is represented. Models may rely on different subsets of features, encode interactions differently, or resolve trade-offs among correlated variables in alternative ways, all while achieving nearly identical performance on standard metrics.

Crucially, this ambiguity is not merely theoretical: it is directly observable through low Rashomon agreement. When agreement is low, the space of hypotheses consistent with the data is large, and any single model’s post hoc explanation reflects only one of many plausible explanations. In these cases, misalignment between explanations and the true data-generating process is not a failure of the explainer per se, but a consequence of insufficient \textit{identifiability} in the data. As a result, explanations produced in low-agreement regimes should be interpreted with caution, regardless of a model’s predictive accuracy.

Conversely, high Rashomon agreement indicates that accuracy tightly constrains model behavior. When equally accurate models converge in their predictions and—especially—their feature attributions, the hypothesis space consistent with the data is more restricted. In these regimes, post hoc explanations are substantially more stable and more likely to align with true underlying effects. This distinction clarifies why explanation-based agreement is a particularly informative diagnostic: it directly probes consistency in the mechanisms that explainers are designed to recover, rather than surface-level agreement in outputs.

Overall, this section reframes explanation reliability as a dataset-level property rather than a property of a specific model or explainer. Rashomon agreement does not guarantee that an explanation is correct, but low agreement provides a clear and actionable warning sign: when many equally accurate models tell different stories, none of them should be taken at face value. This diagnostic perspective complements existing evaluations of post hoc explanations and provides practitioners with a principled way to identify when explanations are most likely to \emph{mislead rather than inform}.

\section{Conclusion}
    \label{sec:discussion}
    As post hoc explanations are gaining popularity in business research, we found an alarming trend of using them to make claims about the direction and strength of variables at the data level rather than at the model level.  However, the validity of this use of post hoc explanations has not been studied in the literature. 
In this paper, we systematically evaluate this practice.
     In the following sections, we will summarize our findings and give some parting advice for the use of post hoc explainers for obtaining data insights in business research. 

    \subsection{Main Findings}
        We contribute several important findings to the use of post hoc explanations in business research.

            First, through an extensive literature review, we found that the use of post hoc explanations to infer information about the data is quite prevalent. 
            Such use deviates from the intended purpose of post hoc explainers; they are designed to explain the machine learning model. Identifying this misuse of post hoc explanations is an important contribution of this paper. 

Second, we show that SHAP and LIME explanations do not reliably recover the direction or relative strength of true feature effects, even when predictive models achieve high accuracy. 

Third, we identify the fundamental sources of explanation misalignment and show that these failures are structural rather than incidental. At a conceptual level, misalignment is driven by the Rashomon effect: the existence of many distinct models with near-identical predictive performance but different internal representations and feature attributions. At the data level, misalignment arises from the complexity of the data-generating process, including feature correlation, nonlinearities, and interaction effects, which together admit multiple substitutable predictive explanations. These forces imply that even in best-case predictive settings, post hoc explanations reflect only one of many plausible hypotheses consistent with the data, rather than a uniquely supported account of the data-generating process.

Fourth, we show that disagreement among equally accurate models provides a powerful diagnostic of explanation unreliability, while common robustness strategies offer only limited mitigation. In particular, agreement in post hoc explanations across models within a Rashomon set is strongly correlated with alignment to the ground truth, especially for feature-importance rankings. Conversely, low agreement reliably signals settings in which explanations are most likely to mislead. Although averaging explanations across models or explainer runs can modestly improve alignment, such strategies do not eliminate long-tailed failures and cannot guarantee reliable recovery of true feature effects.

\subsection{General Concerns About Using Post Hoc Explainers to Get Data Insights}

    Although our paper focuses on the two most widely used post hoc explainers,our concerns about extracting data level $\mathbf{X} \mapsto Y$ insights through post hoc explainers are considerably more general. 
    The Rashomon effect \citep{semenova22rashomon,donnelly2023rashomon,rudin2024amazingthingscomehaving} in machine learning models shows that in any given learning problem there can be an abundance of models with similarly high predictive performance. Each of these models represents a distinct hypothesis about the relationship between the inputs X and the outputs y. As a result, the existence of many near-optimal hypotheses implies that even a perfect explanation of a single model is unlikely to provide a faithful or comprehensive summary of the true relationships encoded in the data.

    Explainable machine learning (XAI) is a relatively young field, having begun in the late 2010s as a subfield of machine learning and only recently being explored as a means of explaining ground truths $\mathcal{G}$ rather than models $\mathcal{M}$. A principal challenge consists in the fact that XAI has not yet figured out best practices for extracting information about $\mathcal{G}$ in a way that mitigates the Rashomon effect. After all, the first post hoc explainer that aims to directly mitigate the Rashomon effect (RID \cite{donnelly2023rashomon}) was only published in 2023.
    
    \subsection{Using Explainers as Hypothesis Generators}

While this paper cautions against using post hoc explanations to make definitive claims about the underlying data-generating process, this does not imply that post hoc explainers are without value.  Instead of serving as tools to \textbf{validate} hypotheses or as evidence about $X \mapsto Y$ relationships, post hoc explanations are best viewed as tools to \textbf{propose} hypotheses that can guide subsequent empirical investigation.

When used in this exploratory capacity, post hoc explanations can help researchers reveal potentially important variables, uncover candidate mechanisms, or suggest heterogeneity patterns that merit closer scrutiny. However, because explanations are not guaranteed to be data-aligned—and may reflect one of many plausible models consistent with the data—any hypothesis they generate must be treated as provisional. For example, examining the degree of agreement in explanations across equally accurate models within a Rashomon set can help assess whether an explanation reflects a stable pattern in the data or one that is sensitive to modeling choices. Low agreement serves as a warning that hypotheses generated from explanations should be interpreted with particular caution.

Credible inference therefore requires follow-up analyses using methods explicitly designed for identification, such as regression-based approaches, causal inference techniques, or experimental designs. Several recent studies illustrate this more appropriate workflow. For example, \cite{zhang2023can} use SHAP to identify potentially important features for restaurant survival, but then rely on causal forests to estimate treatment effects and assess whether those features exert a causal influence. \cite{hanauer22boosting} treat traditional regression as a trusted baseline against which SHAP explanations are compared, rather than as conclusions drawn directly from the explanations themselves. \cite{feng2025ai} uses SHAP to rank important predictors  of celebrity visual
potential and then immediately compare the empirical results with theories to validate the hypotheses. \cite{senoner2022using} uses nonlinear modeling with Shapley additive explanations to infer how a
set of production parameters and the process quality of a manufacturing system are related, and then validate the hypotheses with real-world application. In these cases, post hoc explainers play an exploratory role, while inferential claims are delegated to methods with clearer identification guarantees.

This distinction between hypothesis generation and hypothesis validation is critical. When post hoc explanations are used to guide inquiry rather than to justify conclusions, they can complement existing empirical tools without overstating their inferential content.

\FloatBarrier
{
\tiny
\bibliographystyle{informs2014}
\bibliography{refs}}

\newpage

\newpage
    \renewcommand{\thesection}{OA \arabic{section}}
    \renewcommand{\thesubsection}{\thesection.\arabic{subsection}}
    \setcounter{section}{0}
\section*{Online Appendix}
The following online appendices provide additional information about our experimental designs, additional figures and experimental results, and some theoretical justification for the robustness checks we proposed involving LIME's hyperparameters. 

\section{Paper Review on the Uses of SHAP and LIME}
We conducted two waves of literature search and review.

In the first wave, we used Web of Science to identify publications containing either “SHAP” or “LIME” in the body of the paper within business-related subject areas, including management, economics, supply chain \& logistics, and operations research \& management science. This search yielded over one thousand papers published between 2017 and January 2026. Because publications in business journals typically involve long review cycles, we complemented this search with SSRN to capture relevant working papers. We then manually screened each paper to identify studies that substantively used SHAP or LIME for model explanation, rather than merely citing or mentioning these methods. We stopped once we had identified 150 such papers.

In the second wave, to ensure comprehensive coverage of papers published in leading journals, we focused on journals listed in the \textit{UTD 24}, the \textit{FT50}, and journals published by \textit{INFORMS}. Using the same search keywords (“SHAP” or “LIME”), we manually examined all returned papers and retained those that made substantive use of SHAP or LIME. This screening process yielded 56 papers.

After removing overlaps between the two waves, our final sample consists of 181 unique papers that make substantive use of SHAP and/or LIME. Among these papers, 94\% used SHAP and 13\% used LIME, with 7\% using both methods.

We next classified papers based on how post hoc explanations were interpreted and used. Specifically, we assessed whether authors explicitly or implicitly framed explanations as describing properties of the trained model, or instead extrapolated them to make claims about relationships in the underlying data-generating process. Papers exhibiting such extrapolation were coded as making data-level inferences. Based on this criterion, we categorized papers into three groups: (i) controversial papers that clearly extrapolated post hoc explanations to data-level claims; (ii) ambiguous papers whose wording could plausibly be interpreted as making such extrapolations; and (iii) proper papers that used explanations strictly for model interpretation, consistent with the intended scope of these methods.

To reduce potential labeling bias, we additionally employed GPT-5.2 to review each paper and generate multiple annotations. The full prompt is provided below. We applied strict criteria to classify a paper as having “misused” SHAP or LIME. A paper was given the “misuse” label only if all of the following conditions were satisfied: (1) the paper did not explicitly state that the interpretation was limited to the model level, nor caution readers against causal or data-level interpretations; (2) the LLM identified quotations suggesting an inference of $\mathbf{X} \mapsto Y$ relationships in the underlying data; (3) the LLM identified quotations that derived managerial insights or recommendations directly from the explanations; and (4) the assigned severity level was 4. These criteria are intentionally stringent, making the misuse classification conservative.

Finally, we compared human labels with LLM-generated labels, identified disagreements, and manually re-examined the corresponding papers to determine the final classification.

\begin{tcolorbox}[
  colback=gray!15,   
  boxrule=0pt,       
  frame hidden,      
  breakable,
  left=6pt,right=6pt,top=6pt,bottom=6pt
]
\begin{Verbatim}[fontsize = \footnotesize]
You are assisting with a systematic literature review in business and management research.

Research Question
"Do authors misuse SHAP or LIME by generalizing model-level explanations (X → Ŷ) into claims 
about real-world relationships (X → Y)?"

Core Definition of Misuse
A misuse occurs when feature attributions from: - SHAP (including Shapley-value-based methods and 
variants), or -LIME (including variants) are interpreted as evidence about relationships between
X and the real-world outcome Y, rather than about the model’s prediction Ŷ.

Key Conceptual Distinction

Correct Usage (Model-Level): Statements about how features influence the model prediction (Ŷ). 
Example: "Feature X increases the predicted probability."

Misuse (Data-Level Interpretation): Statements about how features influence the real-world 
outcome Y. Example: "Feature X increases customer churn."

Important Clarifications 
1. This review is NOT about causality. 
2. Even associative language counts as misuse if it refers to real-world Y. Example: "X is 
associated with higher Y." 
3. If the paper explicitly states that explanations apply only to the model or predictions, do
NOT classify as misuse. 
4. Only evaluate language appearing in sections interpreting SHAP/LIME results (e.g., Results, 
Discussion, Implications).

Output Requirement Return a SINGLE JSON object following this EXACT schema:

{ "title": string|null, 
"journal": string|null, 
"year": number|null,
"explainer_used": "SHAP"|"LIME"|"Both"|"None"|"Other", 
"outcome_Y": string|null, 
"use_interpretation_pipeline": "Yes"|"No"|"Unclear",
"explicit_model_level_only": { "answer": "Yes"|"No", "evidence_quotes":[string] }, 
"infer_relationships_X_to_real_Y": { "answer": "Yes"|"No","evidence_quotes": \[string] },
"policy_managerial_recommendations_from_explanations": { "answer":"Yes"|"No", "evidence_quotes": 
[string] }, 
"overall_severity_of_misuse": 1|2|3,
"notes": string|null }

Field Definitions & Coding Rules

explainer_used: - "SHAP" includes Shapley-based methods and variants -"LIME" includes LIME and 
variants - "Both" if both are used - "Other" if other post-hoc explainers are used - "None" if no
explainer is used

use_interpretation_pipeline: Answer "Yes" only if ALL three occur: - The paper trains a 
predictive model - The paper applies SHAP/LIME (or variant) - The paper interprets feature 
attributions in text

explicit_model_level_only: Answer "Yes" ONLY if authors explicitly state explanations apply only
to the model or predictions, or caution against interpreting SHAP/LIME as reflecting real-world 
relationships.

infer_relationships_X_to_real_Y: Answer "Yes" if SHAP/LIME results are described using language 
that refers to the real-world outcome Y (e.g., increases Y, affects Y, drives Y, leads to higher Y).
If authors explicitly limit interpretation to predictions, do NOT code as misuse.

policy_managerial_recommendations_from_explanations: Answer "Yes" if managerial or policy 
recommendations are explicitly derived from SHAP/LIME explanations.

Evidence Rules - Evidence quotes must be verbatim - Maximum 40 words per quote - Do NOT 
paraphrase - If unavailable, use []

Misuse Severity Scale 
1 = No Y-level language 
2 = Minor or isolated loose wording 
3 = Clear, repeated data-level interpretation (X → Y)

Additional Rules - If title/journal/year are not available, set them to
null and explain in notes. - If evidence is ambiguous, explain reasoning
in notes. - Do NOT return anything outside the JSON object.
\end{Verbatim}
\end{tcolorbox}
\section{Data Generation and Model Construction Details for the Explanation Pipeline Demo}

In this first appendix we we document how the synthetic app downloading dataset and the accompanying XGBoost model trained on it, both used in Section \ref{sec:pipeline_demo}, were created.

\subsection{Details on Synthetic App Downloading Dataset}
    \label{apx_sec:app_details}

    The data generation process is based on the linear model shown below as equation \ref{eqn:gt}. This equation was written to express the idea that the probability that people will download a certain app increases with their income, phone usage, usage limits, their exposure to peer pressure and advertisements, positive opinion of the app, and tech savviness. It also expresses that the probability decreases with user age and the extent of their personal privacy concerns. 
    
\begin{equation}
\label{eqn:gt}
\begin{aligned}
\mathcal{G}(X) ={}& 0.0 - .03\cdot\text{Age}
+ 0.5\cdot\text{Income} \\
& {}+ 0.4\cdot\text{Time Spent on Phone}\\
&+ 0.15\cdot\text{Num Apps Installed} \\
& {}+ 0.8\cdot\text{Peer Influence}
+ 0.25\cdot\text{Ad Exposure} \\
& {}- 2.0\cdot\text{Privacy Concern}\\
&+ 0.6\cdot\text{App Rating Perception} \\
& {}+ 1.2\cdot\text{Tech Savviness}\\
&+ 0.7\cdot\text{Unlimited Data Plan}
+ \epsilon.
\end{aligned}
\end{equation}
    \noindent where $\epsilon \sim \mathcal{N}(0,1)$ captures all the variation in the response variable that is not explained by the explanatory variables. The meanings of the feature names in Equation \ref{eqn:gt} as well as the statistical distributions used to simulate them when synthesizing user profiles appear below, in Table \ref{tab:app_features}.

    There are several important items about our procedure for synthesizing user profiles to note. First, after initially generating users' features according the distributions in Table \ref{tab:app_features}, the generated value ranges of ``Tech Savviness" and ``App Rating Perception" are clipped between 0 and 1. Then, ``Privacy Concern" was defined to be correlated with ``Tech Savviness" using the following definition.
    \begin{equation*}
        \text{Privacy Concern} = .7 \cdot \text{Tech Savviness} + \sqrt{1-.7^2}
    \end{equation*}
    We then clip the values of ``Privacy Concern" to be between 0 and 1. ``Privacy Concern" and ``Tech Savviness" then have a Pearson correlation of about .63. The most important is item to note is that the feature values that are input to $\mG$ are standardized into z-scores so that the so that the relative importances of the features can be compared directly using the coefficients of $\mG$. 
    
    The binary target is synthesized for users as follows. Using the logistic sigmoid function $\sigma(\cdot)$, we begin by assigning $Y = 1$ (downloaded) for cases with $\sigma(\mathcal{G}(\mathbf{x}))\geq 0.5$, and $Y = 0$ (ignored) otherwise. 

    \begin{table}[ht]
\centering
\small
\begin{tabular}{l|l|l}
\toprule
\small
\textsc{Feature} & \textsc{Meaning} & \textsc{Distribution} \\ \toprule
\small
Age        &        User Age        & $\mathcal{N}(35,10)$\\ 
Income        & User Income        & $\mathcal{N}(60000,20000)$ \\
Time Spent on Phone        & Average phone hours per day        & $\mathcal{N}(3,1)$ \\ 
Num Apps Installed         & Number of apps installed        & Uniform(0,100)\\ 
Peer Influence        & Measure of peer pressure to install        & Bernoulli(.4) \\ 
Ad Exposure        & Number of times app ad seen & Uniform(0,10) \\ 
Privacy Concern        & Measure of concern about app permissions        & see above \\ 
App Rating Perception    & Measure of impression of app store rating & Uniform(1,5) \\ 
Tech Savviness     & Measure of technological literacy  & $\mathcal{N}(.6,.2)$ \\ 
Unlimited Data Plan & Whether user has unlimited cellular data & Bernoulli(.5) \\ \bottomrule
\end{tabular}%
\caption{Description of features for app downloading data.}
\label{tab:app_features}
\end{table}

\subsection{Details on the App Download Prediction Model}

The model whose explanations were used to produce Figure \ref{fig:sec_3_shap_lime_demo} in the main body of the paper was an XGBoost model from the Python xgboost package. It was obtained using 5 fold nested cross validation. The parameter search loop covered maximum tree depths in [3,4,5], learning rates in [.001,.01,.1], gamma in [0,.25,.5,.75,1], number of estimator in [100,500,1000], colsample\_bytree in [.5,1]. The subsample parameter was fixed at .9, and the number of early stopping rounds was fixed at 10. 

It is also worth mentioning that a root cause of this model's test set accuracy only being 72\% is the variance of the noise term $\epsilon$ in $\mG$. We have confirmed that it is the case that changing the distribution of $\epsilon$ from $\epsilon \sim \mathcal{N}(0,1)$ to, e.g., $\epsilon \sim \mathcal{N}(0,.1)$ makes it so that xgboost models trained as detailed in the previous paragraph can achieve at least 98\% test set accuracy.

\section{Details of the Data Generation Procedure}
\label{apx_sec:81_details}

This section of the appendix describes how we generated the 81 datasets used in Section \ref{sec:main_alignment} of the main paper, as well as how the models we trained on those data were configured.

\subsection{Constructing the datasets}

We generated 81 binary classification datasets by varying four scenario-defining factors while holding sample size and noise fixed. The factor design is summarized in Table~\ref{tab:data_factors} of the main text. 
\begin{table}[htbp]
\centering
\small
\caption{Data Generation Factors and Configurations}
\label{tab:data_factors}
\begin{tabular}{lllc}
\toprule
\textbf{Factor} & \textbf{Variable} & \textbf{Type} & \textbf{Values} \\
\midrule
Sample size & \texttt{Num\_instance} & Fixed & 5{,}000 \\
Noise std. dev. & \texttt{Noise\_std} & Fixed & 0.1 \\
Number of features & \texttt{Num\_feature} & Varied & 10, 20, 30 \\
Correlation strength & \texttt{Corr\_strength} & Varied & 0, 0.5, 1 \\
Nonlinear terms & \texttt{Nonlinear\_num} & Varied & 0, 5, 10 \\
Interaction terms & \texttt{Interaction\_num} & Varied & 0, 5, 10 \\
\bottomrule
\end{tabular}
\end{table}\normalsize

Below, we summarize the step-by-step procedure used to generate the datasets.

\begin{algorithm}[htb]
\caption{Generating Synthetic Dataset}
\label{alg:synth-data}
\begin{algorithmic}[1]
\Require 
    \Statex $N \gets$ number of instances per class (total $= 2N$)
    \Statex $m \gets$ number of original features
    \Statex $s \gets$ noise standard deviation
    \Statex $\rho \gets$ correlation strength
    \Statex $p \gets$ number of nonlinear (squared) features
    \Statex $q \gets$ number of interaction terms
\Ensure 
    \Statex $\mathbf{X}_\text{final}$, $\mathbf{Y}$

\Statex \textbf{1. Generate Original Features:}
    \For{$i=1$ to $m-1$}
        \State $x_i \gets$ sample $2N$ values from $\mathcal{N}(0,1)$
    \EndFor

\Statex \textbf{2. Introduce Correlated Feature:}
    \State Randomly pick $k$, sample $\epsilon \sim \mathcal{N}(0,1)$
    \State $x_m \gets \rho \cdot x_k + \sqrt{1-\rho^2}\cdot\epsilon$
    \State $\mathbf{X} \gets [x_1,\ldots,x_m]$

\Statex \textbf{3. Nonlinear (Squared) Terms:}
    \State Select $p$ features, add squared terms

\Statex \textbf{4. Interaction Terms:}
    \State Select $q$ pairs, add interaction terms

\Statex \textbf{5. Coefficients:}
    \State Sample coefficients from $\text{Uniform}(-1,1)$
\algstore{myalg}
\end{algorithmic}
\end{algorithm}

\begin{algorithm}[htb]
\ContinuedFloat
\caption*{Generating Synthetic Dataset (Continued)}
\begin{algorithmic}[1]
\algrestore{myalg}

\Statex \textbf{6. Compute Probability Vector:}
    \State $z \gets$ linear + quadratic + interaction contributions
    \State $\mathbf{P} \gets \text{sigmoid}(z + \mathcal{N}(0,s))$

\Statex \textbf{7. Generate Labels:}
    \State $\mathbf{Y}[i] = 1$ if $p_i > 0.5$, else $0$

\Statex \textbf{8. Balance Dataset:}
    \State Sample $N$ positives and $N$ negatives
    \State Form $\mathbf{X}_\text{final}, \mathbf{Y}$
\State \Return $\mathbf{X}_\text{final}, \mathbf{Y}$
\end{algorithmic}
\end{algorithm}

\subsection{Constructing the Predictive Model $\mathcal{M}$}
\noindent
Beyond the synthetic data generation process, we also formalize the construction of the predictive black-box models $\mathcal{M}$. 
These models are trained on each generated dataset under varying training sizes and hyperparameters, 
ensuring a range of accuracies across pre-defined bins. 
The following algorithm provides a detailed step-by-step procedure for constructing $\mathcal{M}$.

\begin{algorithm}[htb]
\caption{Construct All Models $\mathcal{M}$}
\label{alg:construct-models}
\begin{algorithmic}[1]
\Require 
    \Statex $D = \{D_1, \ldots, D_{81}\}$ 
    \Statex $H = \{\text{max\_depth}, \text{n\_estimators}\}$
    \Statex $\text{valid\_sizes}$ 
    \Statex $A\_\text{bins} = \{[0.70,0.75), [0.75,0.80), [0.80,0.85), [0.85,0.90)\}$
    \Statex $\text{Models\_per\_bin}=3$
\Ensure $\mathcal{M} = \{\mathcal{M}_{i}^{j}\}$

\State $\mathcal{M} \gets \emptyset$
\For{each $D_i \in D$}
    \State $\mathcal{M}_i \gets \emptyset$
    \For{each bin $a \in A\_\text{bins}$}
        \State $\text{Count} \gets 0$
        \While{$\text{Count} < \text{Models\_per\_bin}$}
            \State Sample $s \in \text{valid\_sizes}$, $h \in H$
            \State Split $D_i$ into (train, test) of size $s$
            \State Train XGBoost $M_\text{temp}$ with $h$
            \State Evaluate $\text{acc}$ on test
\algstore{modelM}
\end{algorithmic}
\end{algorithm}

\begin{algorithm}[htb]
\ContinuedFloat
\caption*{Construct All Models $\mathcal{M}$ (Continued)}
\begin{algorithmic}[1]
\algrestore{modelM}

            \If{$\text{acc} \in a$}
                \State Add $M_\text{temp}$ to $\mathcal{M}_i$
                \State $\text{Count} \gets \text{Count}+1$
            \EndIf
        \EndWhile
    \EndFor
    \State Add $\mathcal{M}_i$ to $\mathcal{M}$
\EndFor
\State \Return $\mathcal{M}$
\end{algorithmic}
\end{algorithm}



\section{Robustness Check Details and Results}
\label{apx_sec:robustness_results}
This section provides information about the motivation and design of the three robustness checks introduced in Section \ref{sec:robustness} of the main body of the paper. The plots illustrating the experimental results of the three robustness checks are also provided here. 

 The subplots in the figures in this section each show shaded PDF curves for one mode of data alignment for one explainer type. Each subplot will have two PDF curves: one representing the alignment of explanations produced baseline usage of SHAP and LIME and  one representing alignment of explanations produced by intervening on the factor relevant for a robustness check. 
Two-sample t-tests for the difference of means are used to indicate whether SHAP and/or LIME are robust to the intervention. Tests whose p-values are significant can be identified by checking for one the signs "(*)," "(**)," "(***)," or "(****)." The number of asterisks between the parentheses is suggestive of \emph{how} significant a p-value is. 

\subsection{Averaging Explanations within A Rashomon Set}

A useful way to describe the motivation of this robustness check is to relate it to the phenomenon where individuals seek out second or third opinions for high stakes medical diagnoses. We may view models in a Rashomon set like expert opinions (hypotheses) about a diagnostic problem ($X \mapsto Y$). The similarity with medicine ends there because whereas patients will be lucky to have access to 3 expert doctor opinions, data scientists can have access to myriad opinions from models and, therefore, will require a way to synthesize them in a coherent and concise manner. 

Since, for each dataset, we trained 10 highly accurate black box models, we have 10 hypotheses about each of the corresponding $X\mapsto Y$ relationships that can only be understood through post hoc explanation. As we are explaining the models using SHAP and LIME, explanations of each model can be represented as feature importance vectors. 

Our robustness check leverages one of the simplest ways to synthesize the SHAP and LIME explanations of each model in a Rashomon set: by simply averaging together each of the feature importance vectors obtained by explaining each model. This procedure produces a single feature importance vector equating the importance of each feature $X_j$ to the average of the importances assigned to that feature by explanations of the models in the Rashomon set.

We investigated whether such a routine is significant by, for each dataset, comparing the data alignment of explanations of A) the model with the highest test set accuracy with B) explanations averaged over the Rashomon set. Our results appear below in Figure \ref{apx_fig:apx_rashomon}. The two significant results we observed were that the averaged explanations had significantly \emph{lower} mean SHAP direction alignment, but significantly \emph{higher} mean LIME strength alignment. From these results we adduce that our averaging strategy cannot be used to completely remedy the misalignment of SHAP and LIME.

   \begin{figure}[htb]
  \begin{subfigure}[t]{.45\textwidth}
    \centering
    \includegraphics[width=\linewidth]{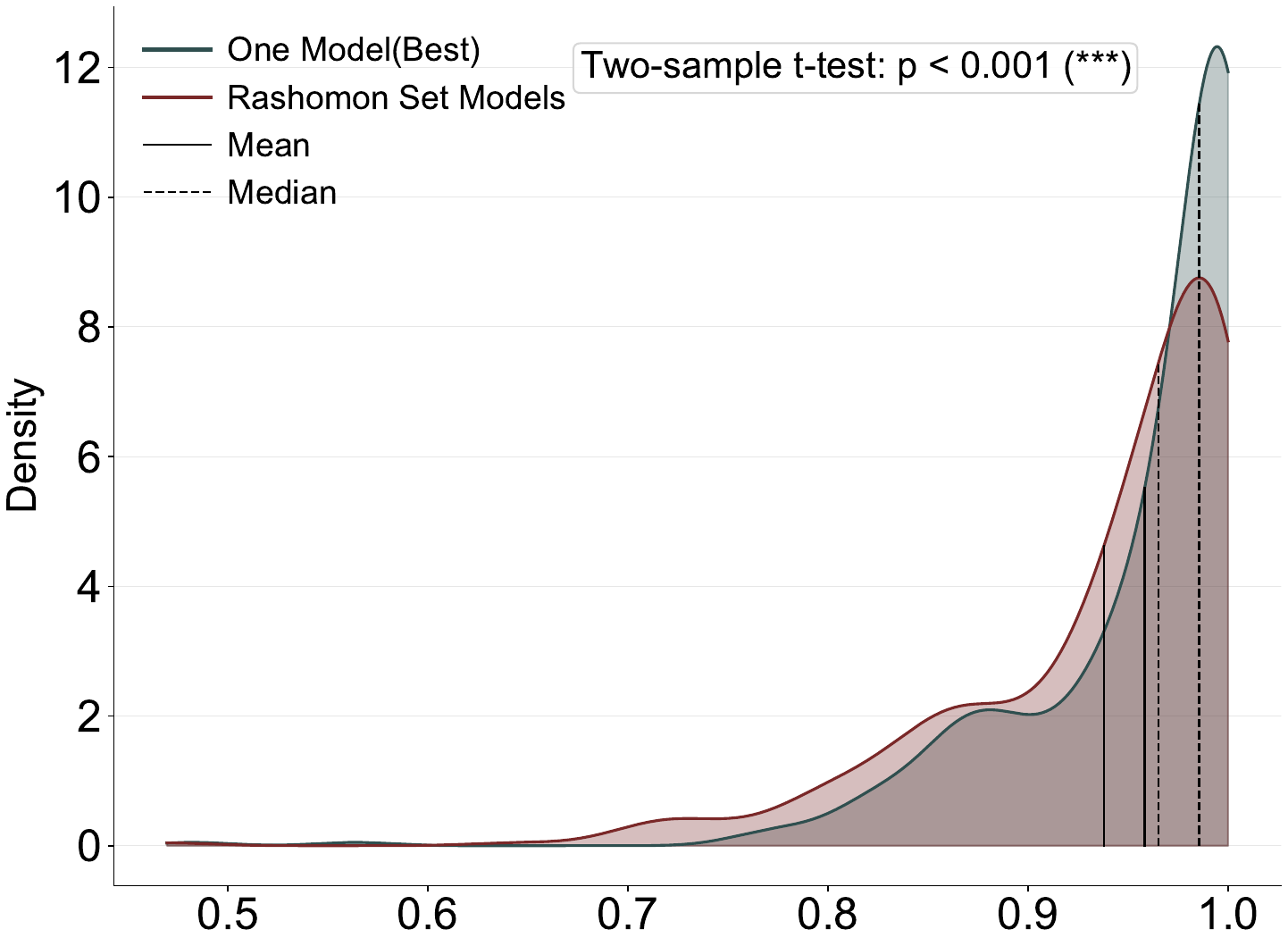}
    \caption{Directionality for SHAP}
    \label{apx_fig:shap_rashomon_3step_dir}
  \end{subfigure}
   \hfill
  \begin{subfigure}[t]{.45\textwidth}
    \centering
    \includegraphics[width=\linewidth]{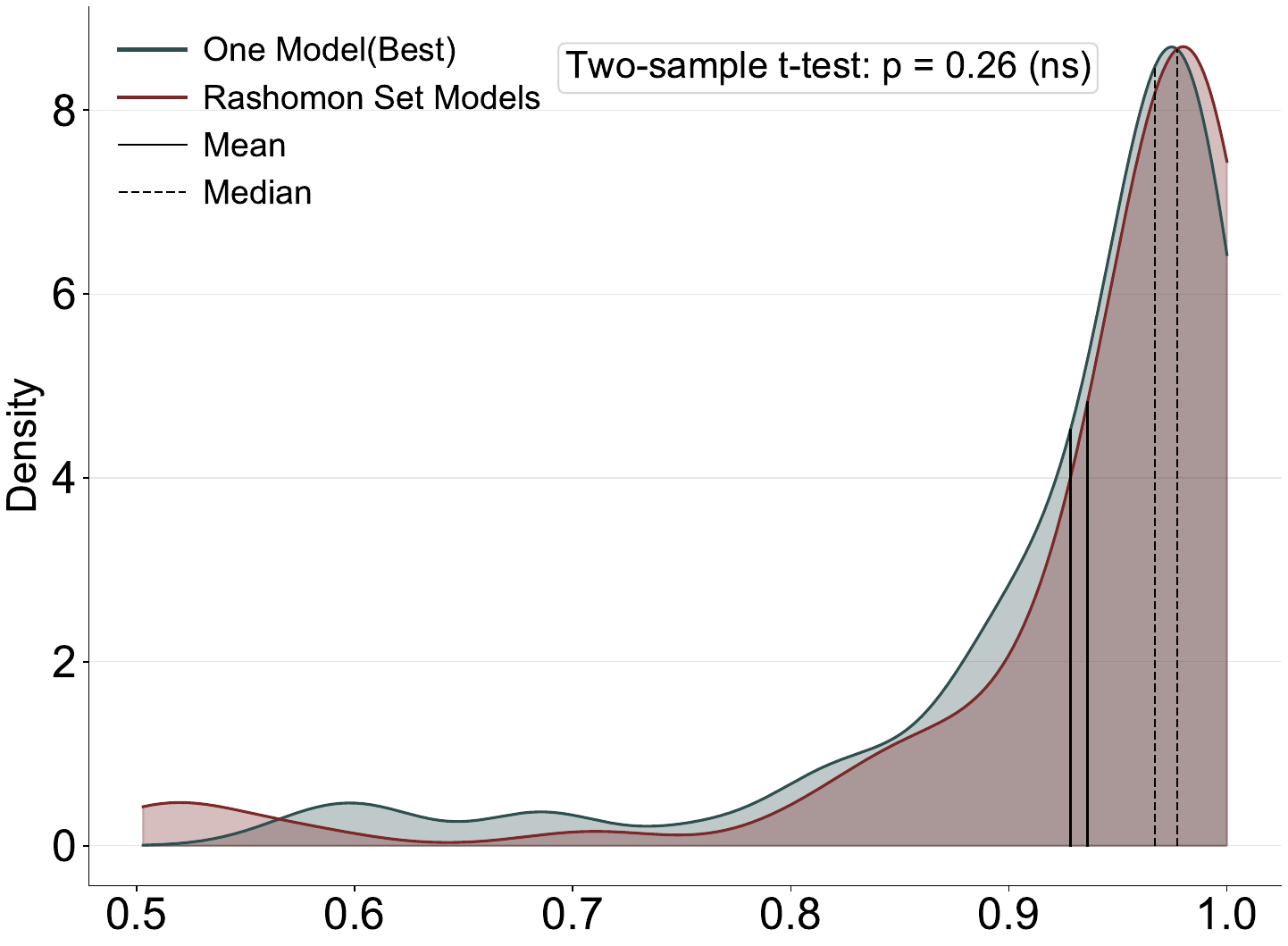}
    \caption{Strength for SHAP}
    \label{apx_fig:shap_rashomon_3step_str}
  \end{subfigure}
  \medskip

  \begin{subfigure}[t]{.45\textwidth}
    \centering
    \includegraphics[width=\linewidth]{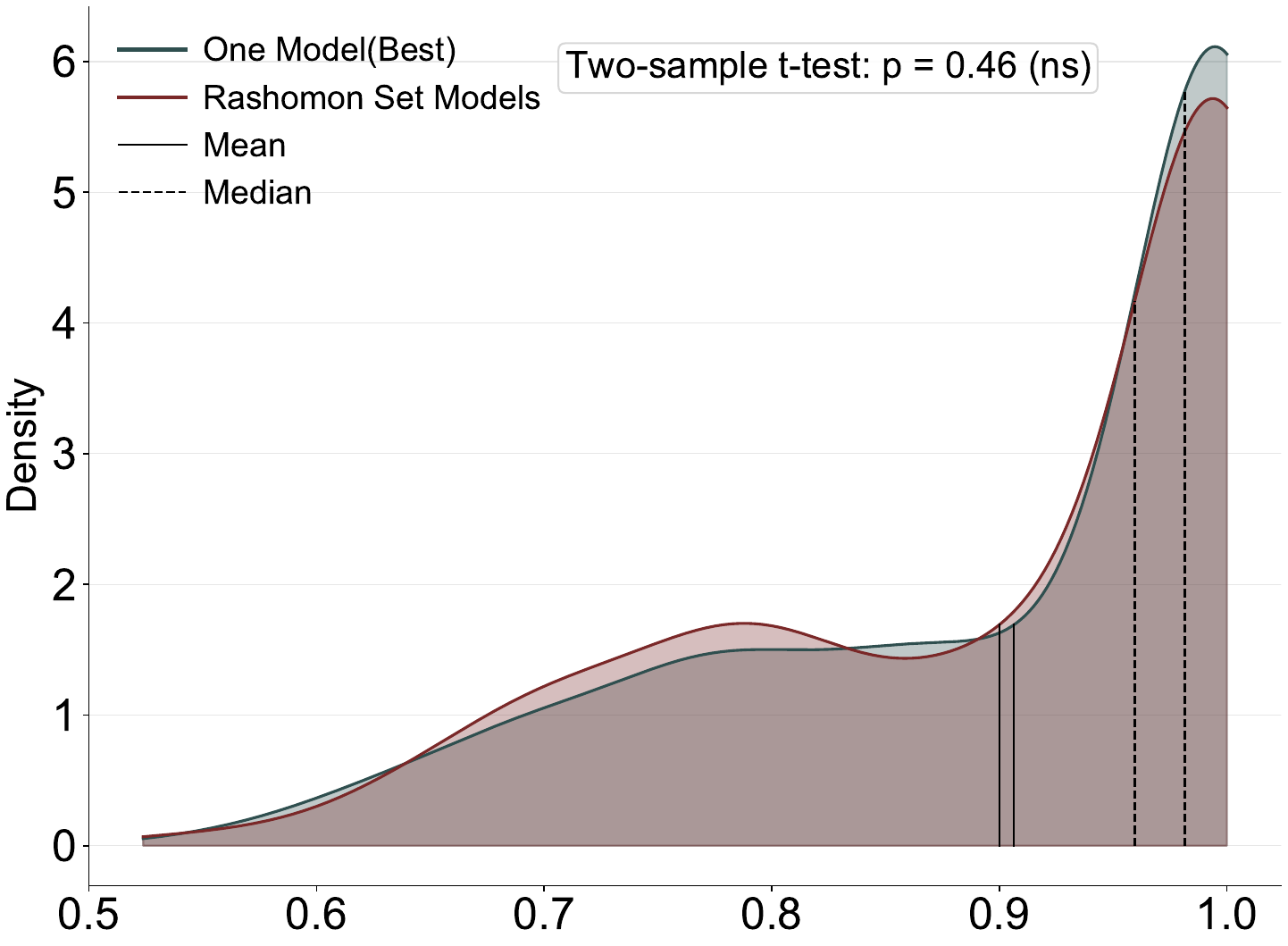}
    \caption{Directionality for LIME}
    \label{apx_fig:lime_rashomon_3step_dir}
  \end{subfigure}
   \hfill
  \begin{subfigure}[t]{.45\textwidth}
    \centering
    \includegraphics[width=\linewidth]{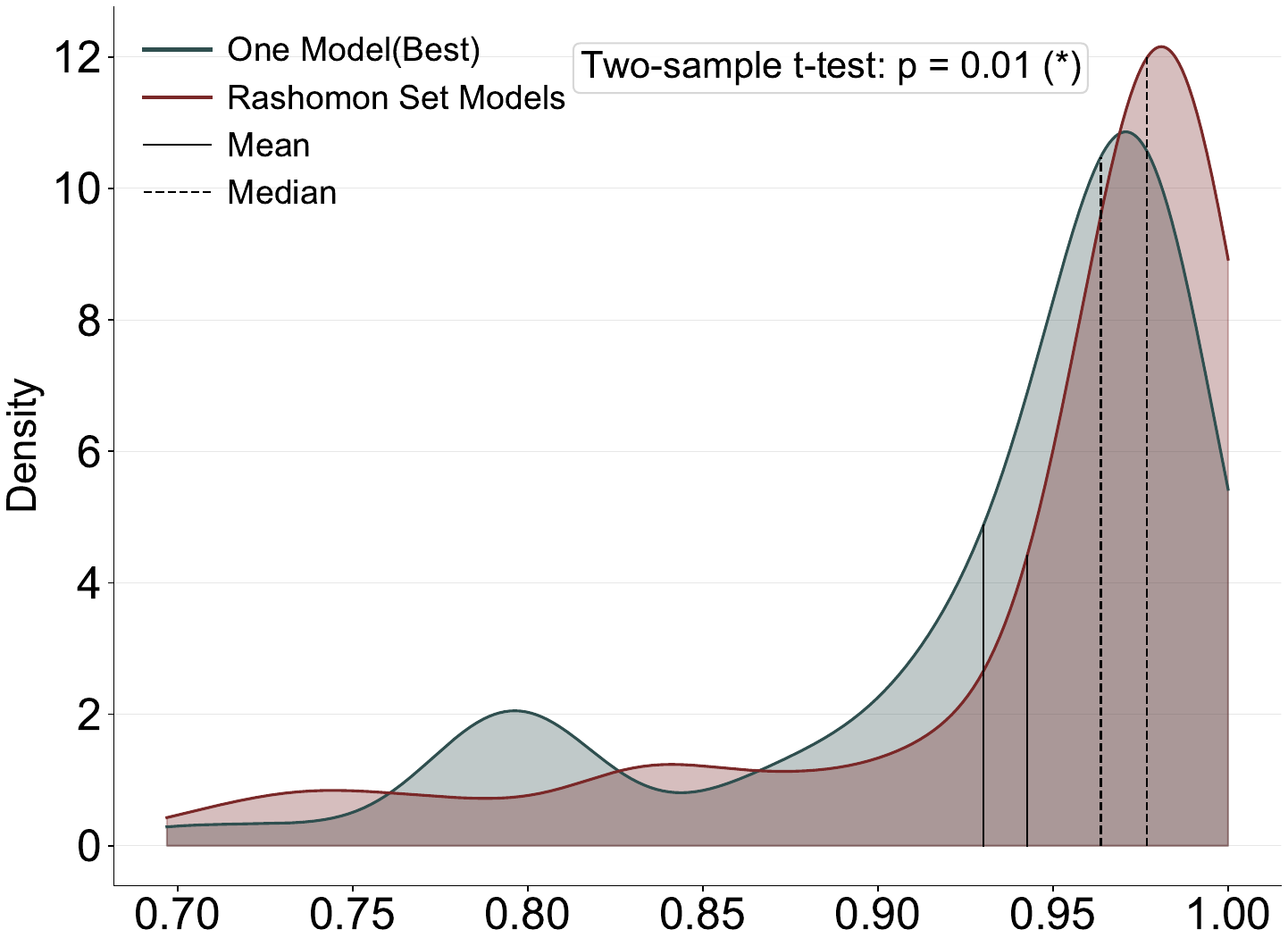}
    \caption{Strength for LIME}
    \label{apx_fig:lime_rashomon_3step_str}
  \end{subfigure}
  \caption{Robustness of SHAP and LIME data alignment to averaging explanations over a Rashomon set}\label{apx_fig:apx_rashomon}
\end{figure}

    \subsection{Sensitivity to Explainer Hyperparameters}
    
    The idea of this robustness check is to check whether leveraging non-default values of explainers' hyperparameters can remedy misalignment. LIME explainers are initialized with parameters for a random seed, sample size $N$, and kernel bandwidth $\mu$. SHAP also has parameters for random seed and sample size $N$, but is known to be essentially deterministic and so SHAP's random seed is less salient. 
    
        \paragraph{Results for LIME Hyperparameters}
        
        LIME's random seed influences the regression used to build the surrogate model, and so has some effect on misalignment. We design one robustness check where the baseline uses LIME explainers with default random seed settings and the intervention uses an average over explanations created from LIME explainers initialized with different random seeds. The averaging here is done in the same manner as was done in the context of the Rashomon set. 
        
         Theoretical justification for this robustness check is provided in Theorem \ref{thm:lime_thm} in Appendix \ref{apx_sec:lime_thms}; it implies that \emph{increases} to both $N$ and $\mu$ beyond their default values have potential to yield comparatively better data alignment. For this reason, we treat explanations produced via LIME explainers with the default hyperparameter settings of $N=1000$ and $\mu\simeq 2.37$ as set $N=\mu=10000$ as the intervention we compare the baseline against. 
        
        Figures \ref{apx_fig:apx_lime_kernel} and \ref{apx_fig:apx_lime_seed} below show that, both increasing $N,\nu$ and averaging explanations from LIME explainers initialized with different random seeds lead to the mean direction alignment was significantly \emph{lower},  but the mean strength alignment was significantly higher. 
        
           \begin{figure}[htb]
        
          \begin{subfigure}[t]{.45\textwidth}
            \centering
            \includegraphics[width=\linewidth]{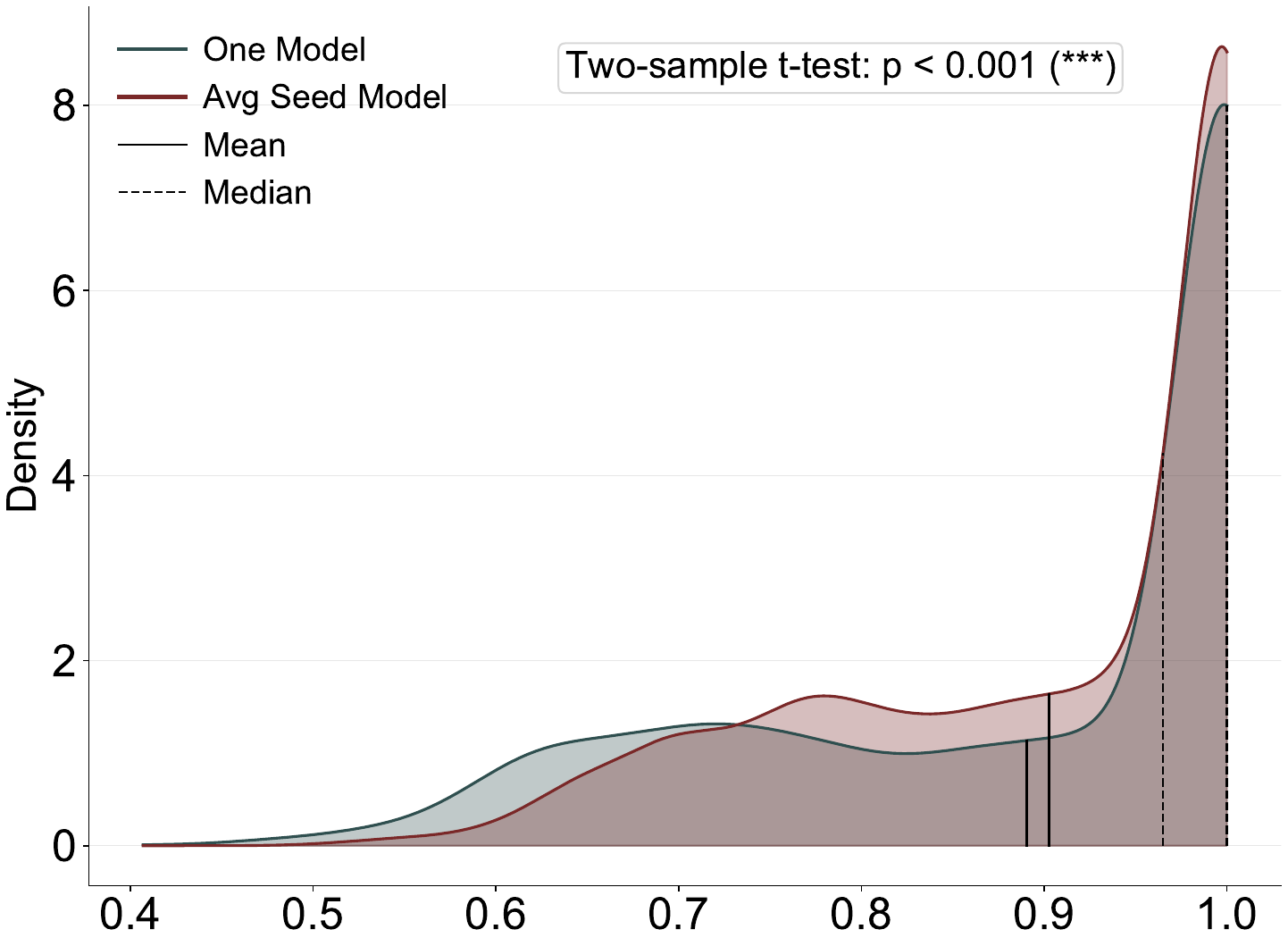}
            \caption{Directionality}
            \label{apx_fig:lime_seed_3step_dir}
          \end{subfigure}
           \hfill
          \begin{subfigure}[t]{.45\textwidth}
            \centering
            \includegraphics[width=\linewidth]{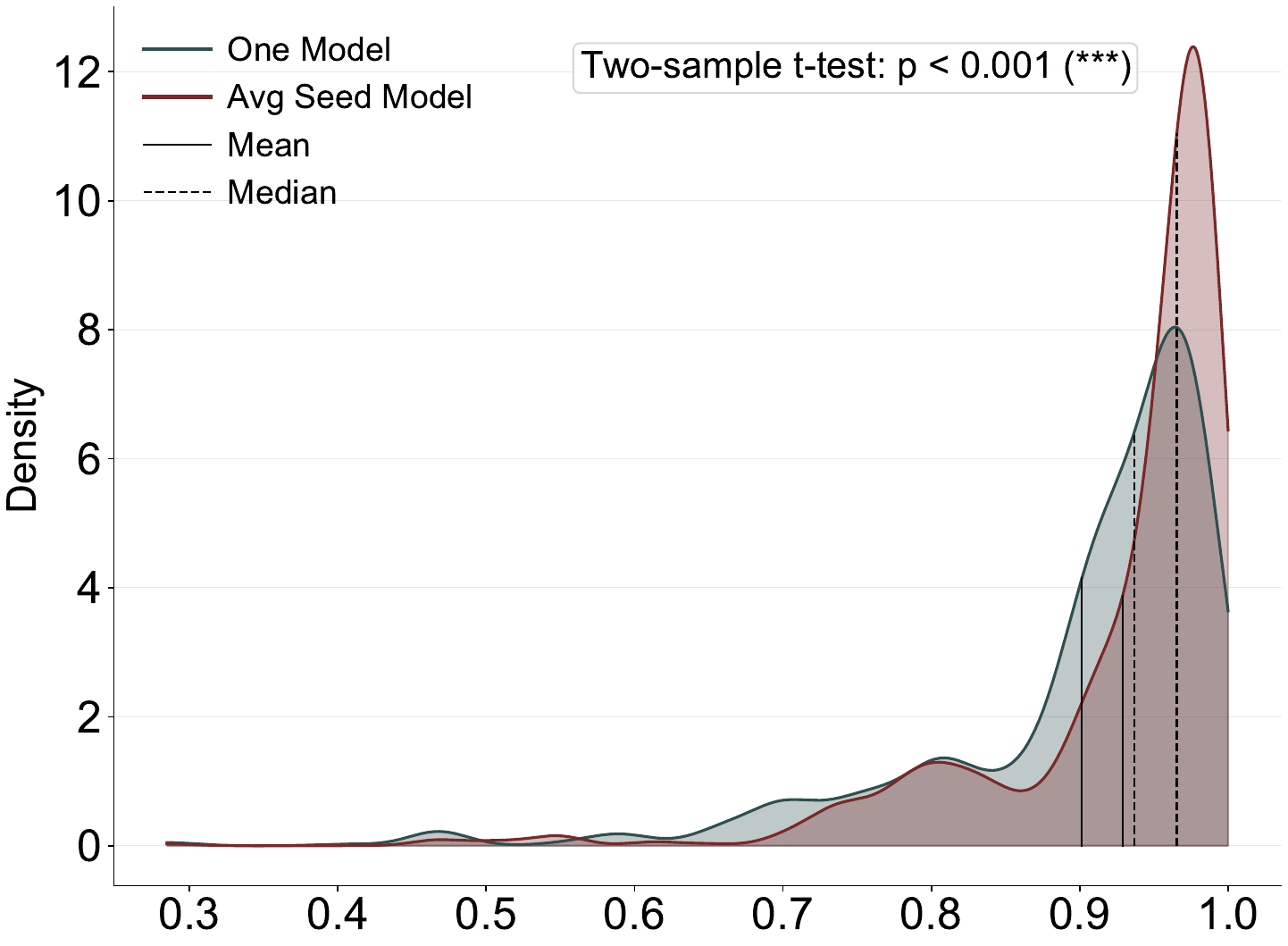}
            \caption{Strength}
            \label{apx_fig:lime_seed_3step_str}
          \end{subfigure}
          \caption{Robustness for LIME's instability, varying its random seed}\label{apx_fig:apx_lime_seed}
        \end{figure}
        
           \begin{figure}[ht!]
        
          \begin{subfigure}[t]{.45\textwidth}
            \centering
            \includegraphics[width=\linewidth]{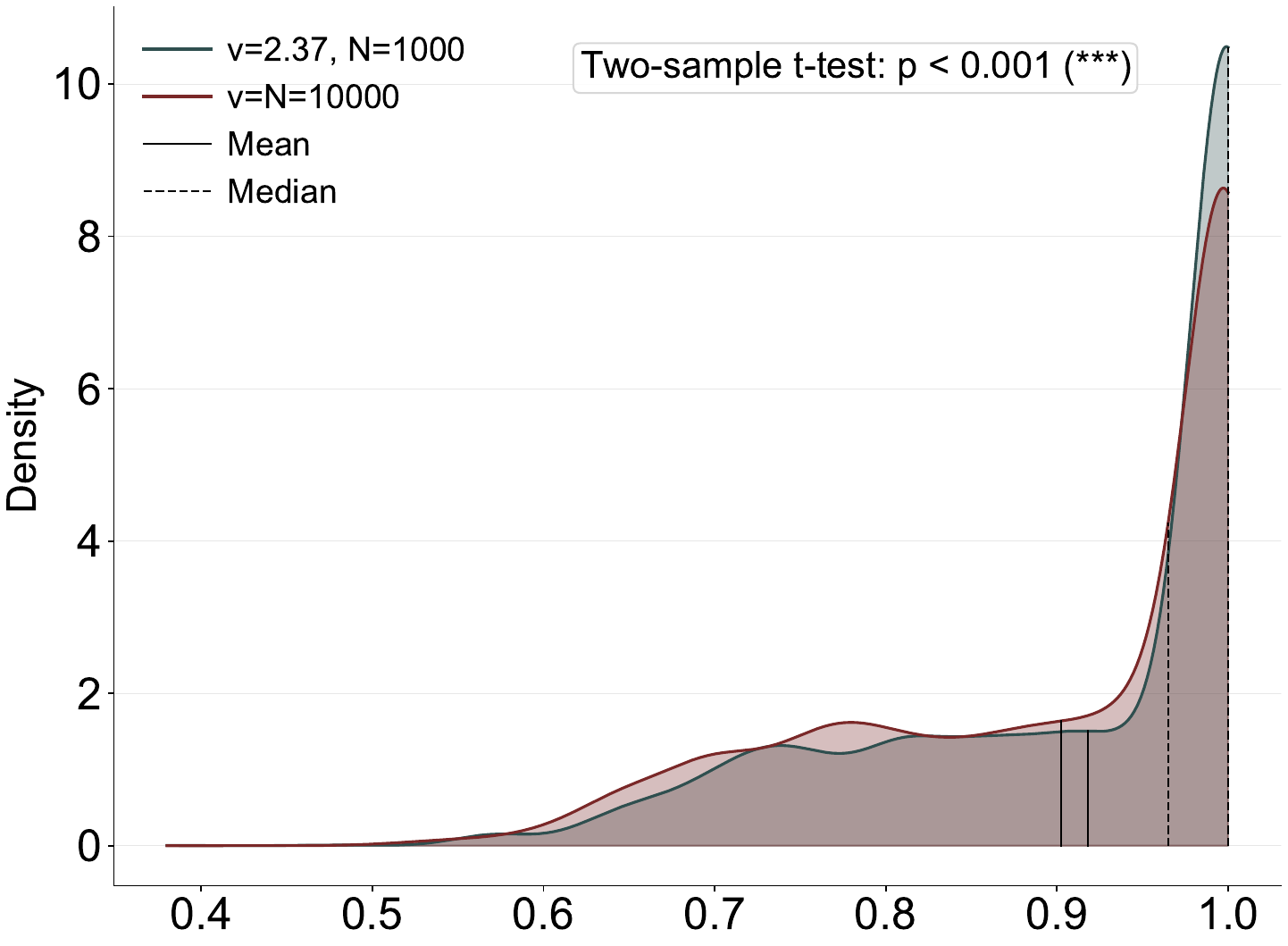}
            \caption{Directionality}
            \label{apx_fig:lime_kernel_3step_dir}
          \end{subfigure}
           \hfill
          \begin{subfigure}[t]{.45\textwidth}
            \centering
            \includegraphics[width=\linewidth]{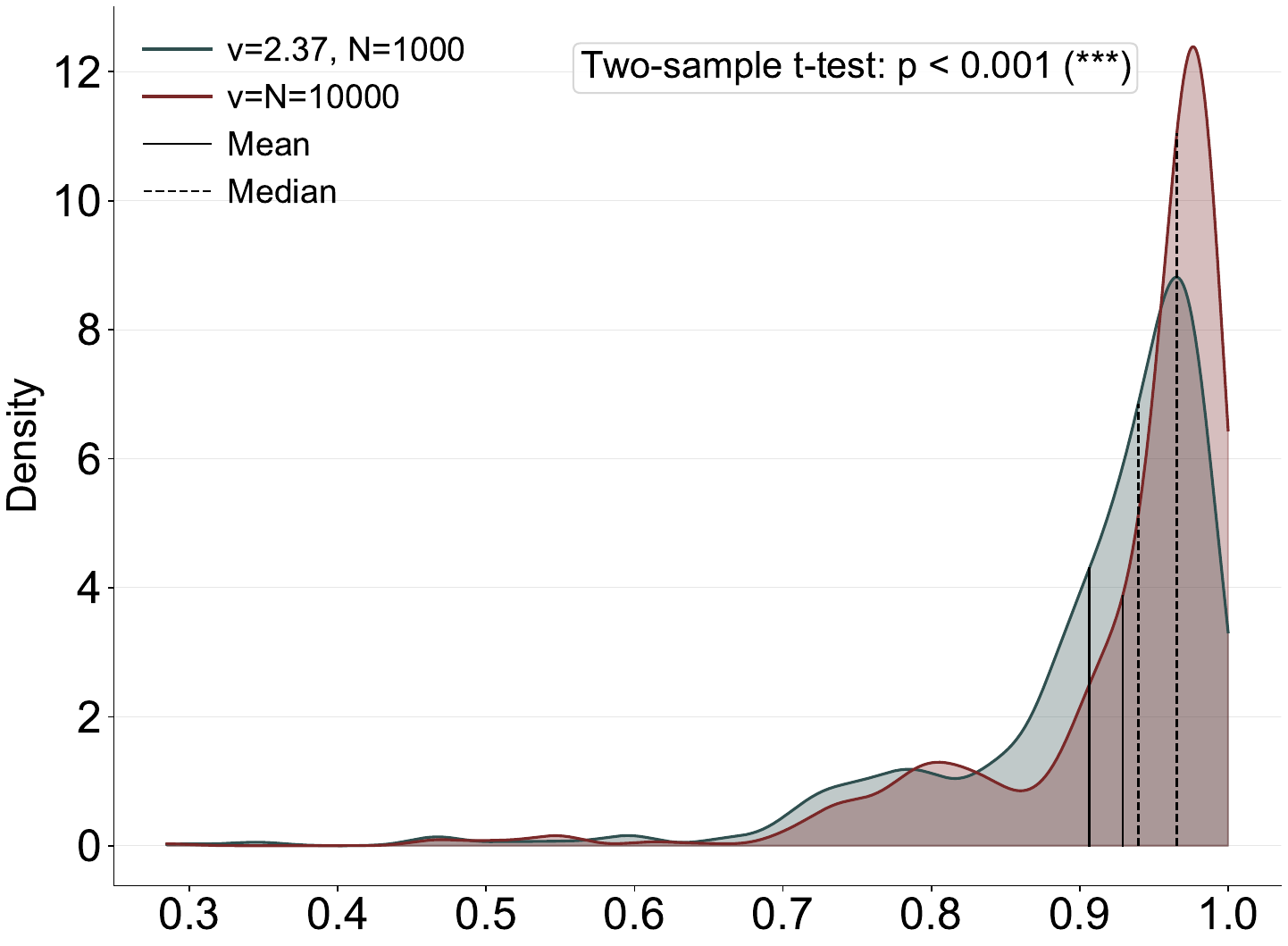}
            \caption{Strength}
            \label{apx_fig:lime_kernel_3step_str}
          \end{subfigure}
          \caption{Robustness for LIME's instability, varying its kernel bandwidth}\label{apx_fig:apx_lime_kernel}
        \end{figure}
        
        \paragraph{Results for SHAP Sample Size}

            The relevant idea behind SHAP's sampling procedure for this section is that increasing the number of samples in the background dataset for SHAP explanations should lead to a better Shapley value approximation. Before we discuss the results however, we will address a potential point of confusion in the Python API for SHAP. There is an \textbf{nsamples} parameter that is passed to explainer objects that controls how many data samples SHAP will produce explanations for. This is \emph{not} the parameter we are changing in this robustness check. Rather, we are changing a parameter in the constructor for SHAP explainer objects controlling number of samples used for the background dataset. Specifically, setting a baseline of 500 samples, our intervention is to set the number of samples to 2000 (using the \textbf{shap.sample()} function). 

\begin{figure}[htb]
                   \begin{subfigure}[t]{.45\textwidth}
                    \centering
                    \includegraphics[width=\linewidth]{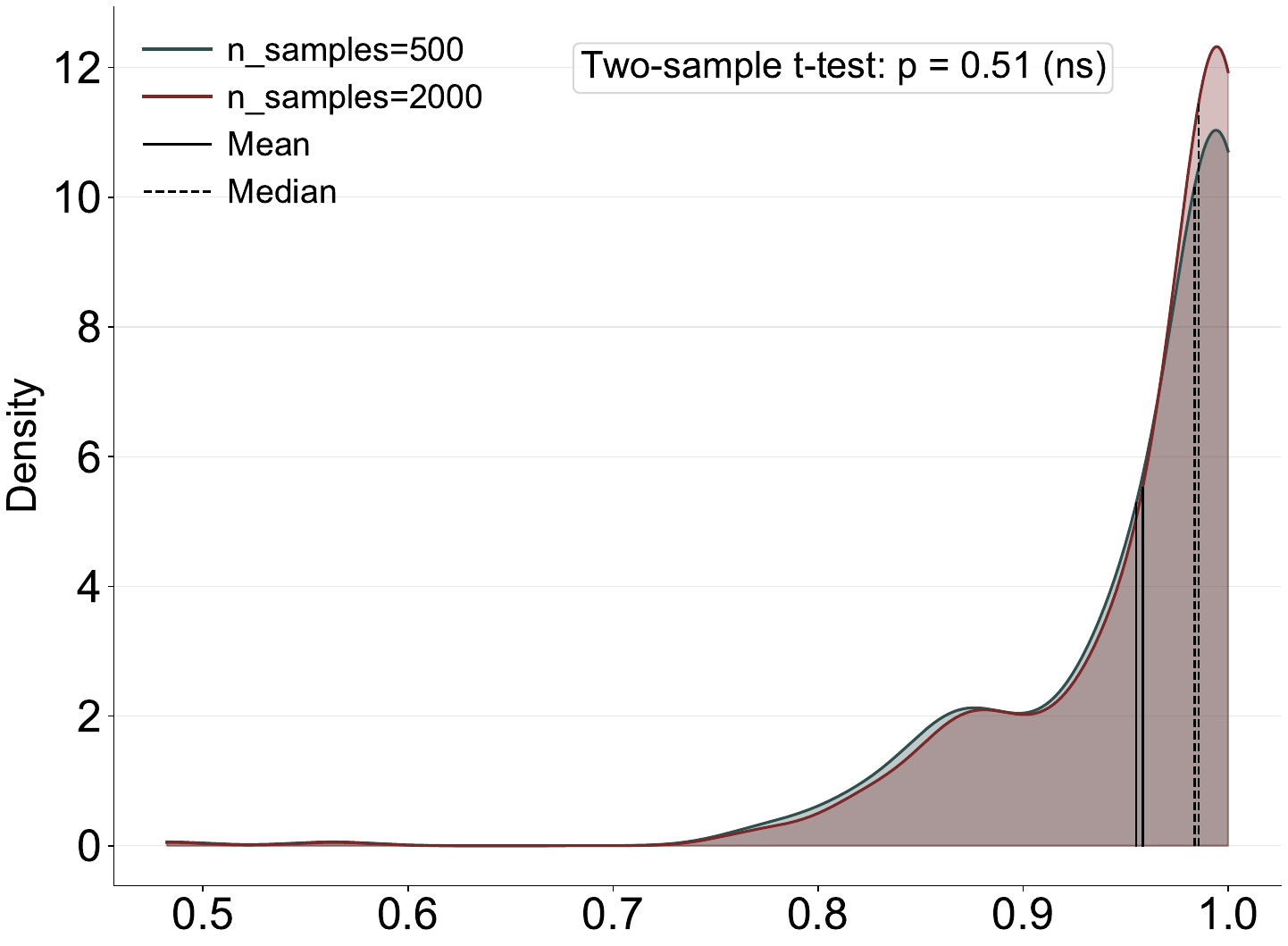}
                    \caption{Direction}
                    \label{apx_fig:shap_sample_3step_dir}
                  \end{subfigure}
                   \hfill
                  \begin{subfigure}[t]{.45\textwidth}
                    \centering
                    \includegraphics[width=\linewidth]{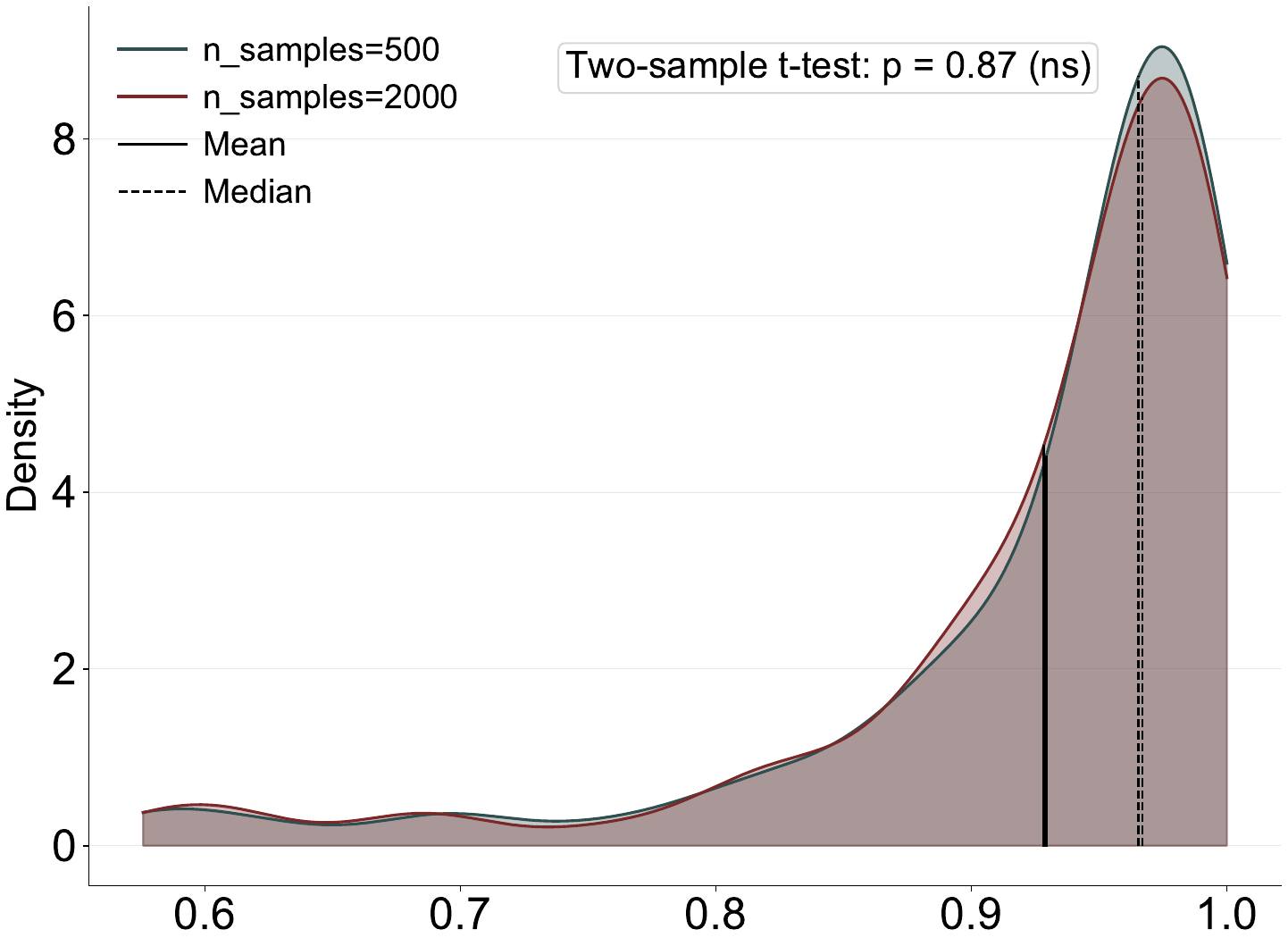}
                    \caption{Strength}
                    \label{apx_fig:shap_sample_3step_str}
                  \end{subfigure}
                  \caption{Robustness for SHAP's instability, varying its sample size}\label{apx_fig:apx_shap_sample}
                \end{figure}

            Figure \ref{apx_fig:apx_shap_sample} suggests that our quadrupling of the number of samples for SHAP's background datasets was not associated with a significant change to SHAP's data alignment. More specifically, the results of our t-tests did not indicate a significant difference between the mean alignments for both direction and strength. In fact, the PDF curves are overall quite similar and for the most part coincide with each other. 

\subsection{Sensitivity to Metric Parameters}

This third robustness check is based on our acknowledgment of the possibility that the value we assign to the free step size parameter used in the definition of our direction alignment metric influences our appraisals of SHAP LIME. As outlined in Section \ref{sec:metrics} in the main body of the paper, the direction alignment metric definition includes a perturbation grid $\Delta = \{m\cdot\texttt{Std}\mid m\in\{-M,\ldots,-1,1,\ldots,M\}\}$ depending on a number of perturbation steps $M$. By default we have $M=3$; the baseline $\Delta$ is $\{-3,-2,-1,1,2,3\}$. The interventions we consider in this robustness check are to set $\Delta$ instead to $\{-1,1\}$, $\{-2,2\}$, and $\{-,3,3\}$. 

The experimental results of this robustness check are shown below in Figures \ref{apx_fig:apx_lime_step} and \ref{apx_fig:apx_shap_step} for LIME and SHAP, respectively. Both figures have four subplots; the first subplot (a) represents the baseline while the other three are part of the intervention. Evidence for our intuition that LIME alignment could suffer as a result of larger step sizes appears in Figure \ref{apx_fig:apx_lime_step}; the mean direction alignment is at least .9 for all choices of $\Delta$ except for the most extreme: $\Delta = \{-3,3\}$. In contrast, it is not apparent from Figure \ref{apx_fig:apx_shap_step} that the choice of $\Delta$ is associated with a significant change to the mean or median direction alignment for SHAP. 

   \begin{figure}[H]
  \begin{subfigure}[t]{.24\textwidth}
    \centering
    \includegraphics[width=\linewidth]{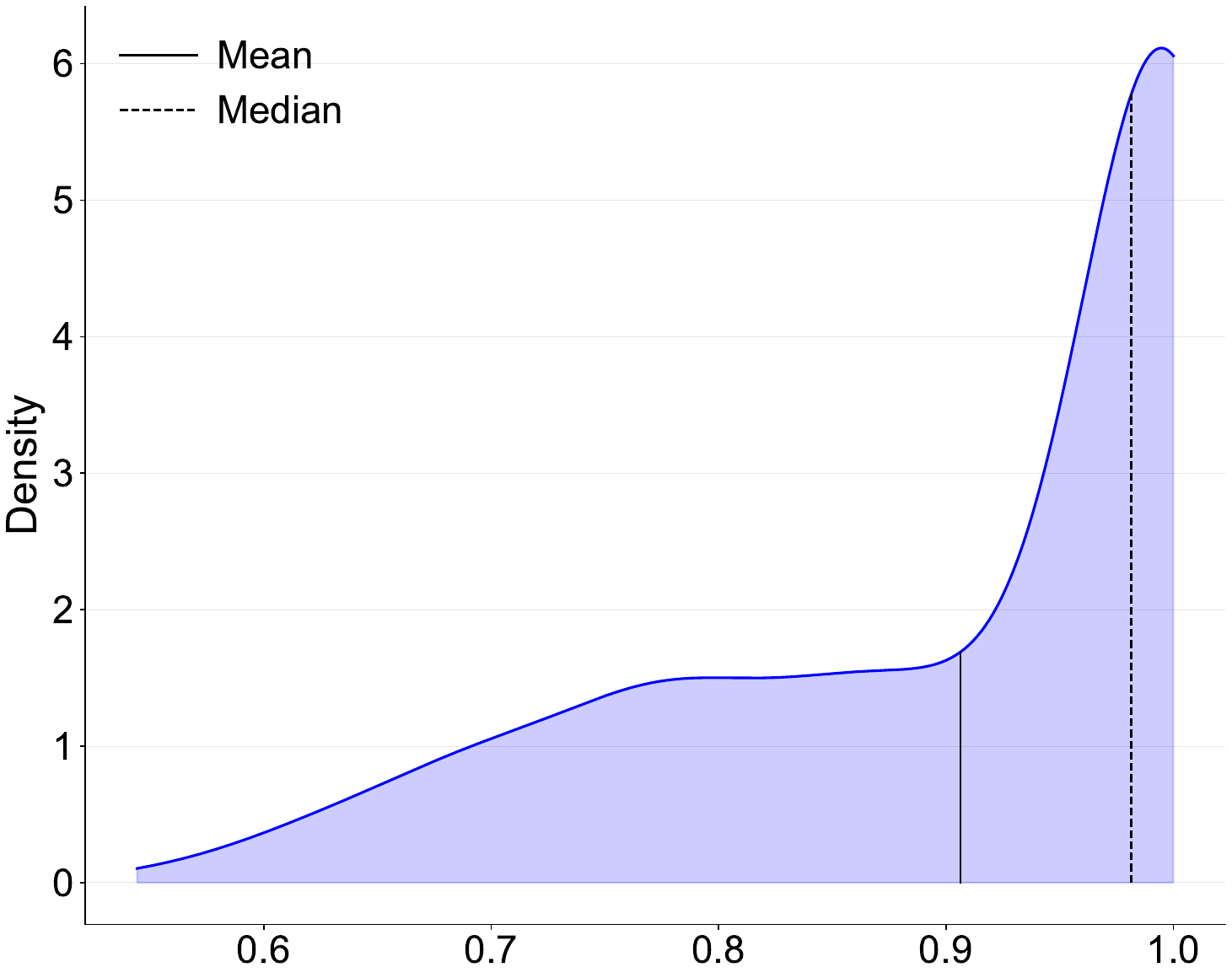}
    \caption{Step = 1,2,3 (default)}
    \label{apx_fig:lime_step_123_dir}
  \end{subfigure}
   \hfill
  \begin{subfigure}[t]{.24\textwidth}
    \centering
    \includegraphics[width=\linewidth]{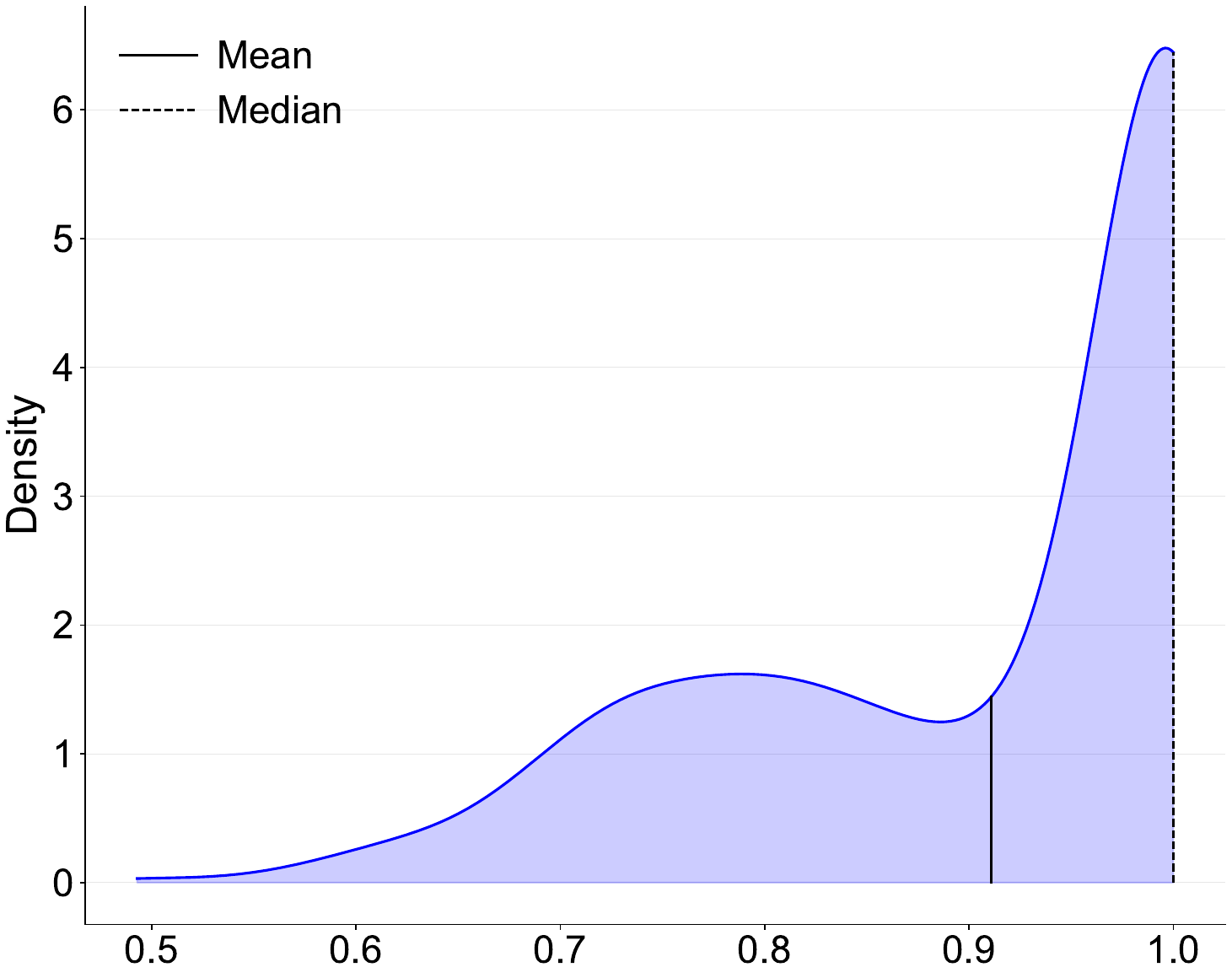}
    \caption{Step = 1}
    \label{apx_fig:lime_step_1_dir}
  \end{subfigure}
   \hfill
  \begin{subfigure}[t]{.24\textwidth}
    \centering
    \includegraphics[width=\linewidth]{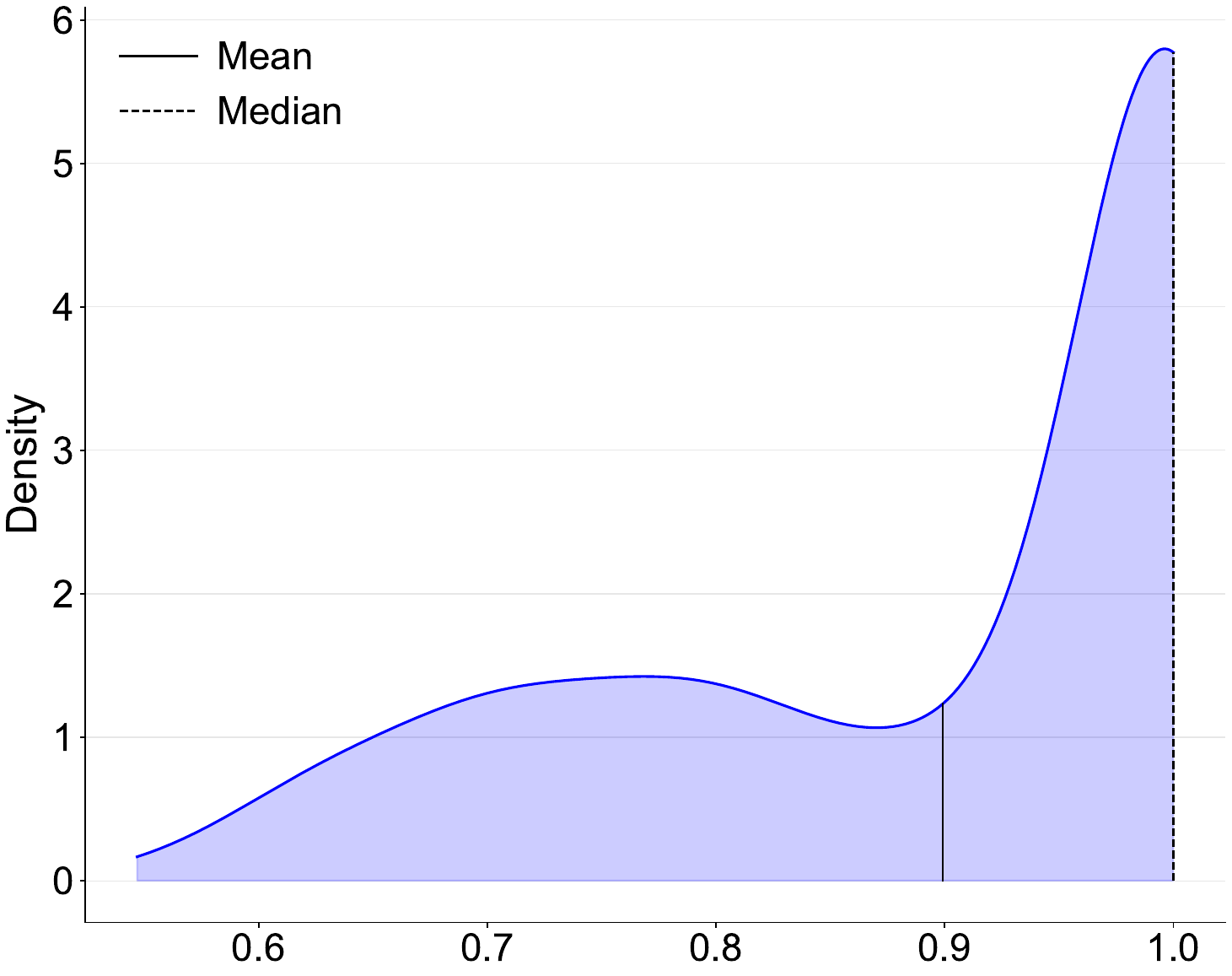}
    \caption{Step = 2}
    \label{apx_fig:lime_step_2_dir}
  \end{subfigure}
   \hfill
  \begin{subfigure}[t]{.24\textwidth}
    \centering
    \includegraphics[width=\linewidth]{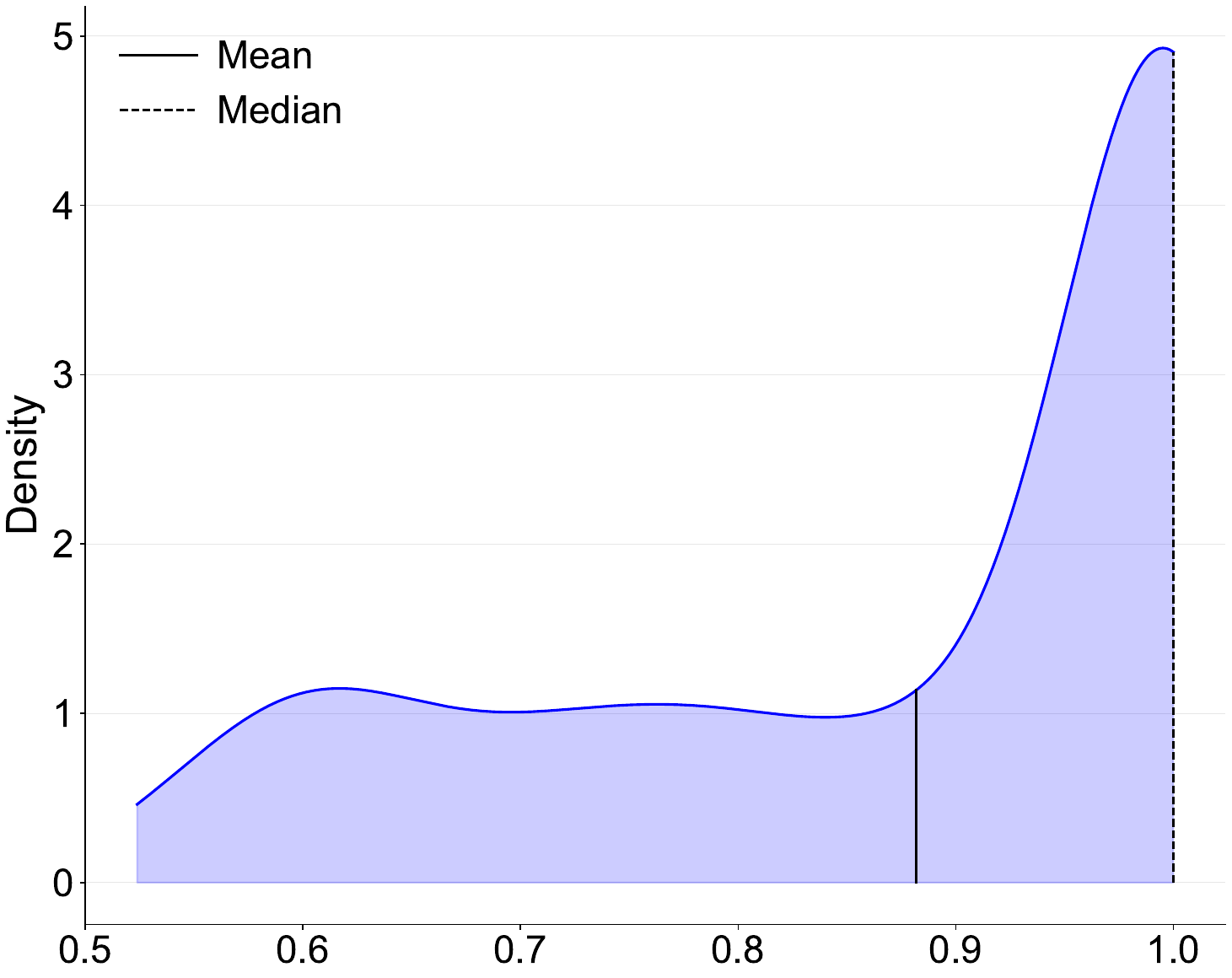}
    \caption{Step = 3}
    \label{apx_fig:lime_step_3_dir}
  \end{subfigure}
  \caption{LIME Step Perturbation}\label{apx_fig:apx_lime_step}
\end{figure}

   \begin{figure}[ht!]
  \begin{subfigure}[t]{.24\textwidth}
    \centering
    \includegraphics[width=\linewidth]{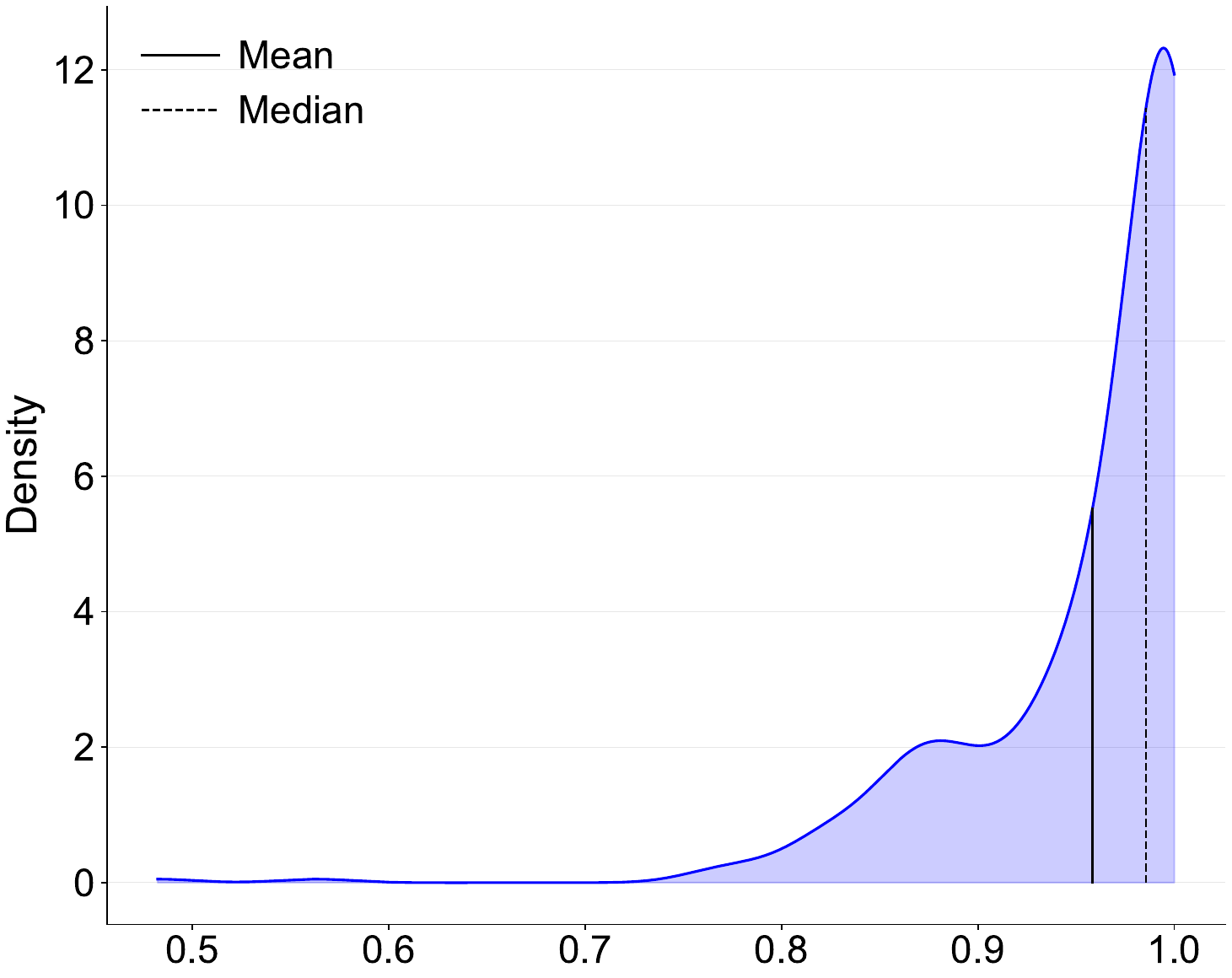}
    \caption{Step = 1,2,3 (default)}
    \label{apx_fig:shap_step_123_dir}
  \end{subfigure}
   \hfill
  \begin{subfigure}[t]{.24\textwidth}
    \centering
    \includegraphics[width=\linewidth]{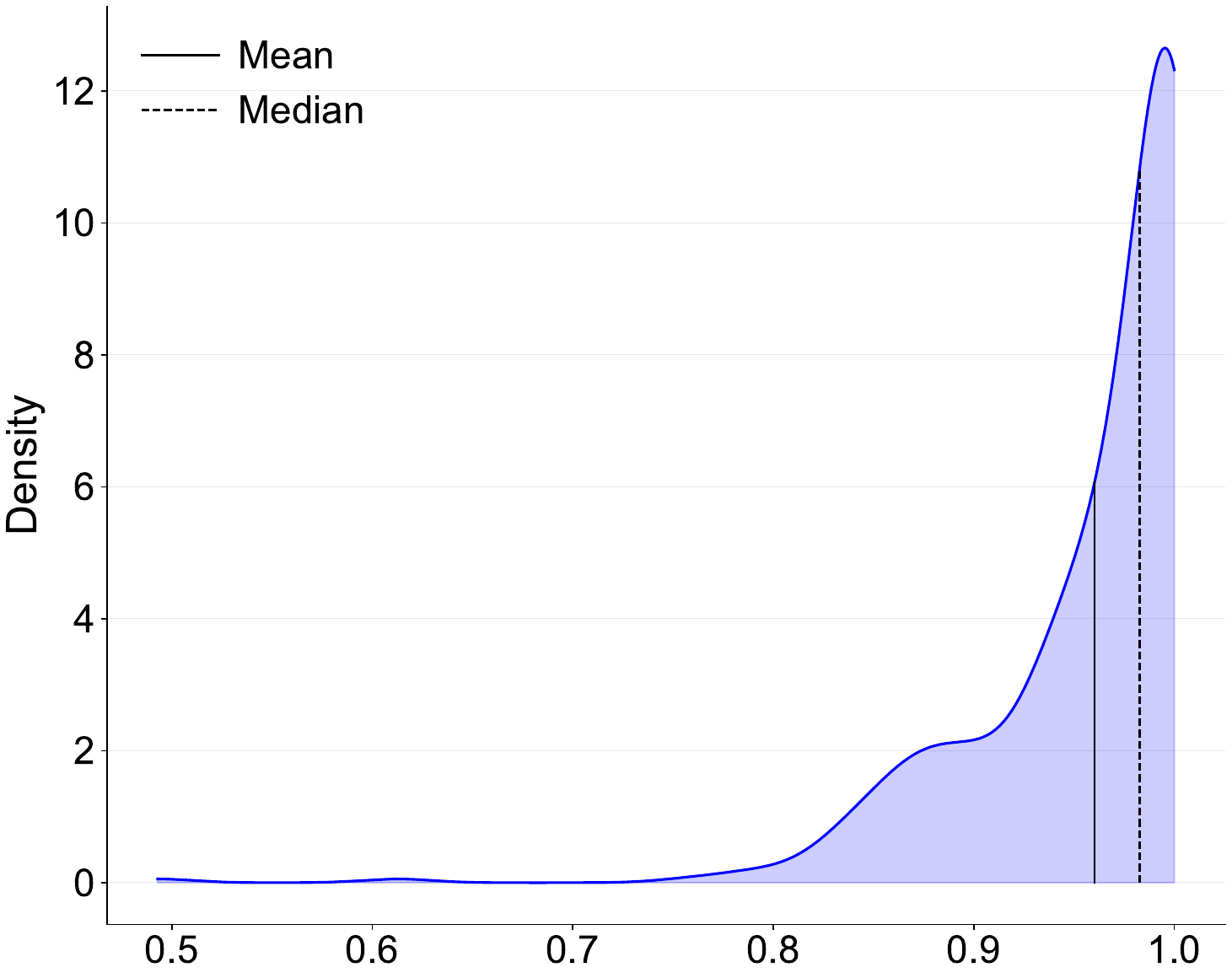}
    \caption{Step = 1}
    \label{apx_fig:shap_step_1_dir}
  \end{subfigure}
   \hfill
  \begin{subfigure}[t]{.24\textwidth}
    \centering
    \includegraphics[width=\linewidth]{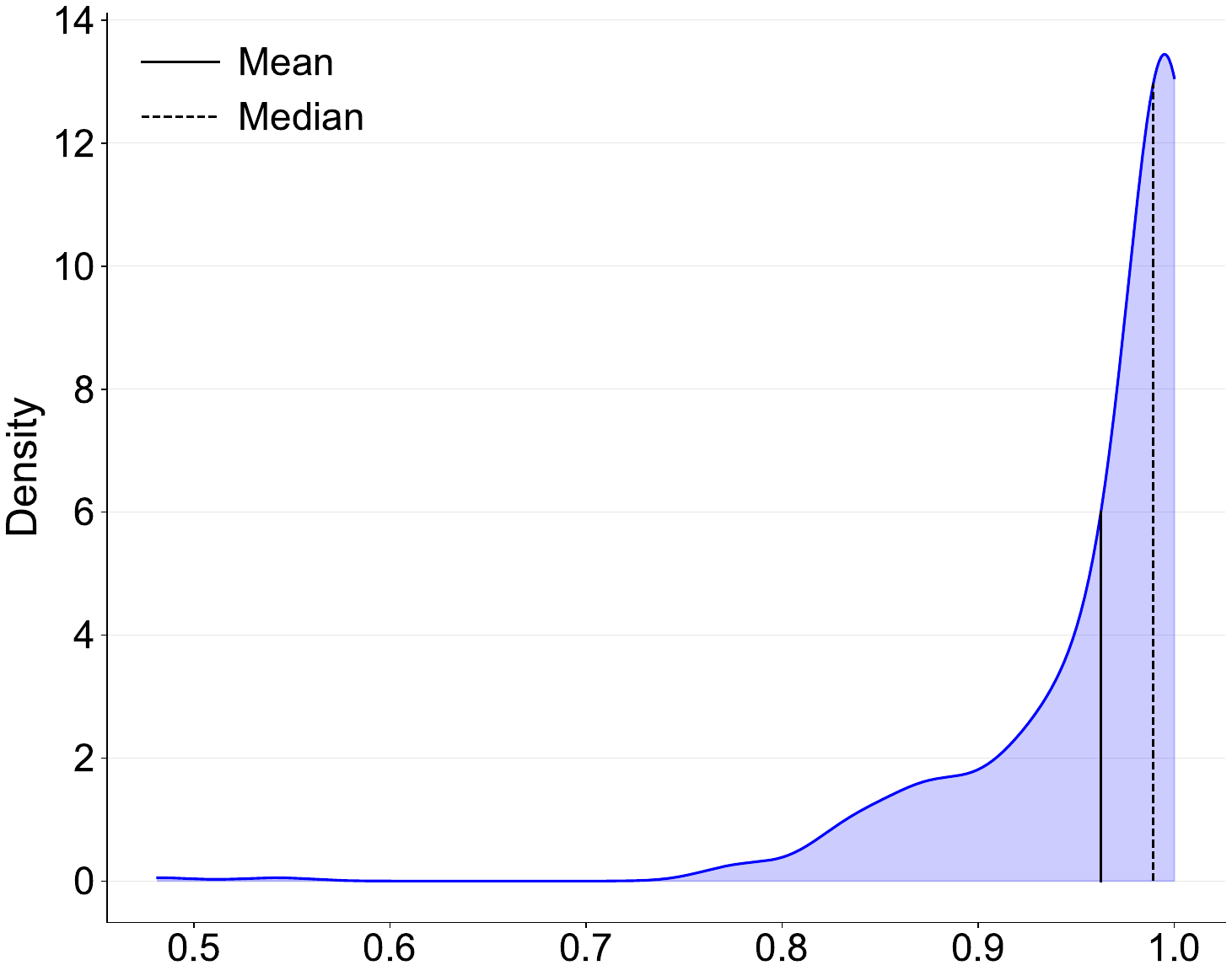}
    \caption{Step = 2}
    \label{apx_fig:shap_step_2_dir}
  \end{subfigure}
   \hfill
  \begin{subfigure}[t]{.24\textwidth}
    \centering
    \includegraphics[width=\linewidth]{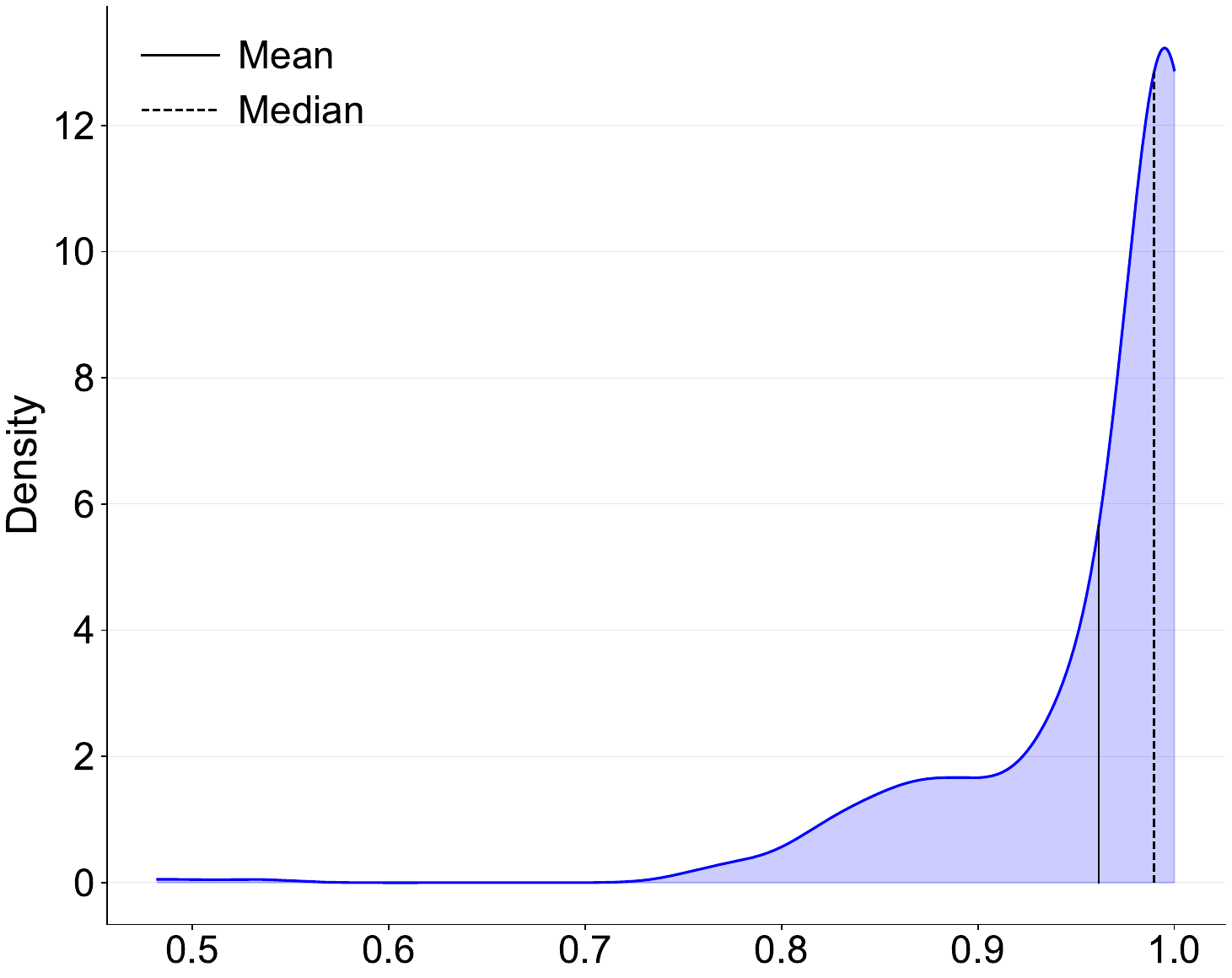}
    \caption{Step = 3}
    \label{apx_fig:shap_step_3_dir}
  \end{subfigure}
  \caption{SHAP Step Perturbation}\label{apx_fig:apx_shap_step}
\end{figure}

\section{Theoretical Justification of LIME Robustness Check 2}
\label{apx_sec:lime_thms}

In Section \ref{sec:lime_params} of the main paper we proposed robustness checks involving an increases to LIME's sample size $N$. The theorem below provides some theoretical justification for this suggestion.
 
        \paragraph{$e_{\text{LIME}}$ when Explaining a Linear Model}
            \label{sec:lime_linear_case}
             
             LIME's explanation model is a weighted ridge regressor trained on discretized data sampled around the instance being explained. By discretizing numerical features and using weighted ridge regression (rather than standard least squares), the possibility of LIME's data-alignment is compromised. The use of discretized features in LIME's ridge regression model can eliminate critical information about the model being explained. Additionally, the sparsity penalty which is used in ridge regression may zero out coefficients for features that actually have a non-zero marginal effect, distorting the picture one may obtain of the marginal effects. Also, employing weighted ridge regression introduces a weight matrix that further skews the estimation of marginal effects. This distortion can be observed in the closed-form solution of the weighted ridge regression, indicating potential limitations in LIME's ability to provide accurate estimates under certain conditions.
             
              Having identified the aforementioned properties of LIME's regression design as contributors to its lack of data-alignment, we derive conditions under which LIME would be data-aligned. We begin this effort by analyzing the large sample behavior of LIME coefficients in an idealized scenario where LIME does not use discretization and the model being explained is a linear function of i.i.d. normal features. Under such idealized assumptions, our Theorem \ref{thm:lime_thm} below shows that LIME's coefficients converge to the ground truth coefficients $\boldsymbol{\beta}$ as the number of samples $N$ tends to $\infty$. 
            \begin{theorem}
                \label{thm:lime_thm}
                Suppose $f:\mathbb{R}^D\rightarrow \mathbb{R}$ that maps $\mathbf{x} \mapsto \boldsymbol{\beta}^T \mathbf{x}$ is explained at a sample $\boldsymbol{\xi}$ drawn from $\mathcal{N}(\mathbf{0}_D, \bI_{D\times D})$ with LIME. Suppose further that LIME does not discretize features and samples i.i.d. data points $\mathbf{x}^{(1)},\ldots,\mathbf{x}^{(N)}$ from $ \mathcal{N}(\boldsymbol{\xi}, \bI_{D\times D})$ for its regression.
                 Then, in the large sample limit, LIME recovers $\boldsymbol{\beta}$.  
            \end{theorem}

             Theorem 1 uses an application of the law of large numbers to show that, in the large sample limit of random design matrices $\mathbf{X}$, LIME recovers $\boldsymbol{\beta}$. The practical implication of this theorem is that, when using LIME, $N$ should be large.  
             Its proof, along with a that of a require lemma, are given below. 


        
         \subsection{Theorem  
        \ref{thm:lime_thm} Proof}
        \paragraph{Theorem Setting}
        
            To investigate the limitations of LIME, we consider the simple case of explaining a linear function $f:\mathbb{R}^D \rightarrow \mathbb{R}$ such that $f(\mathbf{x}) = \boldsymbol{\beta}^T \mathbf{x}$ at the point $\boldsymbol{\xi}\in \mathbb{R}^D$ (which itself was sampled from $\mathcal{N}(\mathbf{0}_D, \bI_{D\times D})$) using a LIME explainer with the following properties. Suppose that LIME samples random vectors $\{\mathbf{x}^{(n)}\}_{n=1}^N$ with $\mathbf{x}^{(n)} \sim \mathcal{N}(\boldsymbol{\xi},\bI_{D\times D})$ for each $n$ and, given a bandwidth parameter $\nu > 0$, weights their distances to $\boldsymbol{\xi}$ using its default exponential kernel $\pi(\xi,\mathbf{x}) = \exp(-\norm{\boldsymbol{\xi} - \mathbf{x}}_2^2/2\nu^2)$. 
            
        We assume that LIME follows its default settings to solve a ridge regression problem, minimizing the following loss function to obtain a feature importance vector $\hat{\bu}$. 
        \begin{equation*}
            (\mathbf{y} - \mathbf{X}\mathbf{u})^T \mathbf{W}(\mathbf{y} - \mathbf{X}\mathbf{u}) + (\mathbf{u} - \mathbf{u}_0)^T \Delta(\mathbf{u} - \mathbf{u}_0)
        \end{equation*}
        In this loss function, $\mathbf{X} \in \mathbb{R}^{N\times D}$ is a data matrix  whose $n^{th}$ row is $\mathbf{x}^{(n)}$ and $\mathbf{W} \in \mathbb{R}^{N\times N}$ is a diagonal weight matrix for each of the $n=1,\ldots,N$ samples with $w_{nn} = \pi(\boldsymbol{\xi},\mathbf{x}^{(n)}) = \exp(-\norm{\boldsymbol{\xi} - \mathbf{x}^{(n)}}_2^2/2\nu^2)$. The response vector $\mathbf{y}\in \mathbb{R}^N$ is defined pointwise by $y^{(n)} = \boldsymbol{\beta}^T \mathbf{x}^{(n)}$. The ridge penalty parameter $\Delta\in \mathbb{R}^{D\times D}$ has $\Delta = \lambda \bI_{D\times D} \bI_{D\times D}$ so that $\Delta$ represents the ridge penalty with hyperparameter $\lambda=1$. As we shall see, $\lambda$ does not matter because LIME does not set it to scale with $N$. There is also another hyperparameter $\mathbf{u}_0 \in \mathbb{R}^D$ which is the target that $\mathbf{u}$ is to be shrunken toward. By default, it is $\mathbf{0}_D$. 

        In the proof, $n$ will be used to index the $N$ data samples. $i$ and $j$ will be used to index the $D$ features. Individual data samples in the data matrix $\mathbf{X}$ will be represented with superscript notation, e.g., $\mathbf{x}^{(n)}$. Individual features of a sample will be indexed with subscript notation, e.g. $x^{(n)}_j$. We need not refer to whole rows or columns of $\mathbf{W}$, so it will be indexed entry-wise with $w_{nn}$ for $n=1,\ldots,N$. Finally, we will use the Dirac delta function $\delta_{i=j}$, which is 1 when $i=j$ and 0 otherwise. 

            \paragraph{\textbf{Theorem \ref{thm:lime_thm}}}
                Suppose $f:\mathbb{R}^D\rightarrow \mathbb{R}$ that maps $\mathbf{x} \mapsto \boldsymbol{\beta}^T \mathbf{x}$ is explained at a sample $\boldsymbol{\xi}$ drawn from $\mathcal{N}(\mathbf{0}_D, \bI_{D\times D})$ with LIME. Suppose further that LIME does not discretize features and samples i.i.d. data points $\mathbf{x}^{(1)},\ldots,\mathbf{x}^{(N)}$ from $ \mathcal{N}(\boldsymbol{\xi}, \bI_{D\times D})$ for its regression.
                 Then, in the large sample limit, LIME recovers $\boldsymbol{\beta}$.   

\noindent \emph{Proof.}
    The main idea underlying the proof is to apply the law of large numbers to a closed form solution to generalized ridge regression. Using the notation previously established, we can write a closed form solution of the regression as
    \begin{equation}
        \label{eq:closed_form}
        \hat{\bu} = (\bX^T \bW \bX + \bI)^{-1}(\bX^T \bW \by)
    \end{equation}
    Here, $\hat{\bu}$ is a point estimate using a single sample of data $\bX$. By simple algebraic manipulation we have, equivalently, 
    \begin{gather*}
         \hat{\bu} =  (\bX^T \bW \bX + \bI)^{-1}(\bX^T \bW \bX) \boldsymbol{\beta}
    \end{gather*}
    Let us define $\hat{\bS} = \frac{1}{N}(\bX^T \bW \bX + \bI) $ and $\hat{\bG} = \frac{1}{N}\bX^T \bW \bX$ so that $\hat{\bu} = \hat{\bS}^{-1}\hat{\bG}\boldsymbol{\beta}$.
    
    We may write $\hat{\bS}$ and $\hat{\bG}$ entry-wise as follows.
    \begin{gather*}
        (\hat{\bS})_{ij} = \frac{1}{N}\sum_{n=1}^N( x_i^{(n)} w_{nn} x_j^{(n)} + \delta_{i=j}) \\
        (\hat{\bG})_{ij} = \frac{1}{N} \sum_{n=1}^N x_i^{(n)} w_{nn} x_j^{(n)} 
    \end{gather*}
    Above, $\delta_{i=j}$ represents the Dirac delta function which is 1 when $i=j$ and 0 otherwise. For fixed $i,j$, the sequence of random variables $x_i^{(1)} w_{11} x_j^{(1)},\ldots,x_i^{(N)} w_{NN} x_j^{(NN)}$ are mutually independent because the samples $x^{(1)},\ldots,x^{(N)}$ are i.i.d. random vectors. 
    By writing the members of the sequence in the form
    \begin{equation*}   
        x_i^{(n)} e^{-\norm{\bxi - \bx^{(n)}}_2^2/2\nu^2} x_j^{(n)}
    \end{equation*}
    we can see that are also identically distributed. 
    By Lemma \ref{finitemean}, the terms have finite means. 
    
Thus, by the strong law of large numbers, the sample averages
$\frac{1}{N}\sum_{n=1}^N x_i^{(n)} w_{nn} x_j^{(n)}+\delta_{ij}$
converge to
$\mathds{E}_{\bx \sim \mathcal{N}(\bxi,\mathbf{1}_{D})}\!\left[x_i e^{-\norm{\bxi-\bx}_2^2/2\nu^2}x_j\right]$
almost everywhere. This means that the ridge adjustment vanishes (converges to a null matrix). Likewise, the averages $\frac{1}{N}\sum_{n=1}^N x_i^{(n)} w_{nn} x_j^{(n)}$ converge to the same value $\mathds{E}_{\bx \sim \mathcal{N}(\bxi,\mathbf{1}_{D})}[x_i e^{-\norm{\bxi-\bx}_2^2/2\nu^2}x_j]$ almost everywhere. Thus, in the large sample limit, $\hat{\bS}$ and $\hat{\bG}$ converge to the same matrix. \textbf{It follows that, in the large sample limit, $\hat{\bS}^{-1}$ and $\hat{\bG}$ cancel each other, and so we expect that $\hat{\bu} \rightarrow \boldsymbol{\beta}$.} That is, $\boldsymbol{\beta}$ is recovered by the regression in spite of the ridge penalty and sample weighting. 
    

\begin{lemma}
    \label{finitemean}
   $x_i^{(n)} w_{nn} x_j^{(n)}$ has a finite mean for all $i,j,n$. 
\end{lemma}
    \noindent \emph{Proof.} Fix some $i,j,n$. 
    Since a random variable having finite variance implies that it has a finite mean, it is sufficient to show that $x_i^{(n)} w_{nn} x_j^{(n)}$ has a finite variance. We do this using the fact that, for dependent random variables $A$ and $B$
    \begin{equation}
        \label{varbd}
        \mathrm{Var}(AB) \leq  2\mathrm{Var}(A)\lVert B\rVert_\infty^2+2(\mathbb E[A])^2\mathrm{Var}(B)
    \end{equation}
    When using this inequality, we set $A:=x_i^{(n)} x_j^{(n)}$ and $B:=w_{nn}$.

    We will establish that each term on the RHS of Equation \ref{varbd} is finite in order to show that $\mathrm{Var}(AB)$ is finite. To see that $\mathrm{Var}(A)$ is finite, note that $x_i^{(n)} x_j^{(n)}$ are Chi-square random variables with 1 degree of freedom, which have finite variance. The weights $B=w_{nn}$ are bounded in the interval $(0,1]$, so $\norm{B}_\infty^2$ is finite. Chi-square random variables have finite means, so $\mathds{E}[A]$ is finite. To see that $\mathrm{Var}(B)$ is finite, note again that $B$ is bounded in $(0,1]$. Then $\mathds{E}[B] \leq 1$. As $B$ is bounded, $B^2$ is also bounded. Hence, the variance $\mathrm{Var}(B) = \mathds{E}[B^2]-\mathds{E}[B]^2$ is a subtraction of finite quantities and therefore also finite. Evidently, each term in the RHS of Equation \ref{varbd} is finite, and so $x_i^{(n)} w_{nn} x_j^{(n)}$ has a finite variance. It follows that the mean of $x_i^{(n)} w_{nn} x_j^{(n)}$ is also finite.  \qed

        \section{Computational Cost and Runtime Accounting}
\label{apx:computation}

This section documents the computational resources and runtime required to conduct the experiments reported in the paper. The scale of computation reflects both the number of simulated datasets and the intensive nature of post hoc explanation methods, particularly SHAP.

\subsection{Model Construction}

Across all experiments, we generated 81 synthetic datasets. For each dataset, we estimated predictive models for two distinct experimental settings corresponding to Sections~5 and~7 of the paper.

\textbf{Models used in Section~5.}  
For the analysis in Section~5, we trained 12 XGBoost models per dataset with systematically varying predictive accuracy. To obtain evenly spaced accuracy levels in the range 0.7--0.9, we conducted a grid search over a $60 \times 10 \times 4$ grid spanning the number of estimators, tree depth, and training-set size. Each parameter configuration required approximately 3 seconds of runtime. Aggregated across all datasets, model construction for Section~5 required approximately
\[
60 \times 10 \times 4 \times 3 \times 81 / (3600 \times 24) \approx 6.75
\]
days of continuous computation.

\textbf{Models used in Section~7.}  
For the analysis in Section~7, we constructed a Rashomon set of 10 high-accuracy models per dataset using nine machine-learning algorithms (including XGBoost, CatBoost, and LightGBM). For each algorithm, we explored an average of 81 hyperparameter configurations using five-fold cross-validation, with each configuration requiring approximately 2 seconds. Aggregated across datasets, model construction for Section~7 required approximately
\[
81 \times 5 \times 2 \times 81 / (3600 \times 24) \approx 0.76
\]
days of computation.

All model training was conducted on a workstation equipped with an Intel i7 CPU and 32~GB RAM. In total, predictive model construction required approximately \textbf{7.5 days} of continuous computation.

\subsection{Post hoc Explanation Generation}

The computational burden increases substantially in the explanation stage. For each dataset--model pair, we generated SHAP and LIME explanations for 100 test instances. To compute the directionality metric, we additionally generated six perturbed versions of each instance ($m=-3,-2,-1,1,2,3$), resulting in 700 explanation tasks per model.

All explanation experiments were executed on a dedicated server equipped with 80 CPU cores (2.5~GHz) and 256~GB RAM, with up to 30 parallel processes per task.

\textbf{SHAP.}  
SHAP proved particularly computationally intensive. For models used in Section~5, generating a single SHAP explanation required approximately 30 seconds on average. For the high-accuracy models used in Section~7, the average runtime increased to approximately 90 seconds per instance due to heterogeneity in model structure and SHAP compatibility. Aggregating across all datasets and models, SHAP explanation required approximately
\[
\frac{81 \times (12 \times 700 \times 30 + 10 \times 700 \times 90)}{3600 \times 24 \times 30}
\approx 27.6
\]
days of continuous server-level computation.

\textbf{LIME.}  
By contrast, LIME was substantially faster, requiring approximately 0.1 seconds per instance on average. Across all datasets and models, LIME explanations required approximately 0.05 days of computation and did not materially affect total runtime.

\subsection{Total Runtime}

Taken together, our experimental pipeline required approximately \textbf{35.6 days} of computation: 7.5 days for model construction and 27.6 days for SHAP-based explanation. These figures underscore why evaluating SHAP on entire test sets would be computationally infeasible in this setting and motivate our use of a fixed subsample of test instances in the main analysis.

\end{document}